\documentclass[acmsmall, screen]{acmart}

\usepackage{amsmath}
\newcommand{\sr}[1]{{\color{black}{#1}}}

\usepackage{tabularx}
\AtBeginDocument{%
  \providecommand\BibTeX{{%
    \normalfont B\kern-0.5em{\scshape i\kern-0.25em b}\kern-0.8em\TeX}}}

\usepackage[algo2e,lined,commentsnumbered,ruled]{algorithm2e}
\SetKwInput{KwInput}{Input}
\SetKwInput{KwInit}{Initialize}
\SetKwInput{KwOutput}{Output}
\SetKwInput{KwReturn}{Return}
\SetKwInput{KwSet}{Set}

\usepackage{algorithm}
\usepackage{algpseudocode}

\usepackage{multirow}
\usepackage{bm}
\usepackage{graphicx}

\setcopyright{none}

\begin{document}

\title{Towards Accurate Spatiotemporal COVID-19 Risk Scores using High Resolution Real-World Mobility Data}

\author{Sirisha Rambhatla}
\authornote{Both authors contributed equally to this research.}
\email{sirishar@usc.edu}
\author{Sepanta Zeighami}
\authornotemark[1]
\email{zeighami@usc.edu}
\author{Kameron Shahabi}
\email{kyshahab@usc.edu}
\author{Cyrus Shahabi}
\email{shahabi@usc.edu}
\author{Yan Liu}
\email{yanliu.cs@usc.edu}
\affiliation{%
  \institution{University of Southern California}
  \streetaddress{941 Bloom Walk}
  \city{Los Angeles}
  \state{California}
  \country{USA}
  \postcode{90089}}


\renewcommand{\shortauthors}{Rambhatla and Zeighami, et al.}
\newcommand{\riskModelName}{LocationRisk@T}
\newcommand{\simlName}{SpreadSim}

\begin{abstract}
As countries look towards re-opening of economic activities amidst the ongoing COVID-19 pandemic, ensuring public health has been challenging. While contact tracing only aims to track past activities of infected users, one path to safe reopening is to develop reliable spatiotemporal risk scores to indicate the propensity of the disease. Existing works which aim to develop risk scores either rely on compartmental model-based reproduction numbers (which assume uniform population mixing) or develop coarse-grain spatial scores based on reproduction number (R0) and macro-level density-based mobility statistics. Instead, in this paper, we develop a Hawkes process-based technique to assign relatively fine-grain spatial and temporal risk scores by leveraging high-resolution mobility data based on cell-phone originated location signals. While COVID-19 risk scores also depend on a number of factors specific to an individual, including demography and existing medical conditions, the primary mode of disease transmission is via physical proximity and contact. Therefore, we focus on developing risk scores based on location density and mobility behaviour. We demonstrate the efficacy of the developed risk scores via simulation based on real-world mobility data. Our results show that fine-grain spatiotemporal risk scores based on high-resolution mobility data can provide useful insights and facilitate safe re-opening. 
 \end{abstract}


\maketitle

\section{Introduction}

As the Coronavirus Disease, COVID-19, becomes a long-term challenge in our day-to-day lives, planning and learning effective ways to navigate the disease has become critical. In the earlier phases of the disease when there were relatively small number of cases, contact tracing -- identifying people who may have come into contact with an infected individual -- served as an effective tool to mitigate disease spread \cite{Keeling861}. However, as the number of cases reach record levels, and a general sense of \emph{pandemic fatigue} sets in, it has become important to develop practical tools for navigating the disease safely to resume normal activities \cite{pandemicfatigue1, pandemicfatigue2, ferguson2020report}. 

COVID-19 transmits through contacts between individuals. It spreads in a population due to infected people moving and co-locating with people susceptible to the disease \cite{lotfi2020covid}. For such susceptible individuals, infection risk boils down to going to locations with high probability of co-locating with infected individuals \cite{cdccolocation}. Thus, one approach towards allowing safe resumption of normal activities is assigning risk scores for different regions to show the danger in each area. This can be used both for policy making at the government level as well as individual decision making (e.g., to avoid high-risk areas). To this end, the risk-scores provided need to be (1) spatiotemporally fine-grained, (2) reliable, and (3) accurately evaluated.

\textbf{Fine-Grained Risk-Scores using mobility patterns}. Coarse-grain risk scores and reproduction number estimates at county or state level are not readily usable for public policy decisions at finer spatial and temporal scales \cite{Chande2020, globalepidemic}. Analogous to traffic congestion prediction for transportation applications (e.g., car navigation) where high-resolution information such as average speed at a specific freeway segment at a particular time is critical to avoid potential congested routes, risk scores need to be local to an area to influence decision making \cite{kiamari2020}. In other words, coarse-grain risk scores are akin to reporting average speed of cars for an entire city, which provides no useful information for both individual and policy-making plans. Nevertheless, such information becomes increasingly useful with finer granularity, both spatially and temporally. This improvement in spatiotemporal resolution has also led to recent advances in traffic prediction \cite{li2018diffusion}.  

Motivated from this insight, we focus on developing finer-grain spatiotemporal risk scores leveraging the real-world mobility patterns from one area in a city to another. Recent works which primarily focus on disease forecasting, consider coarse-grain mobility densities (at county-level) \cite{Chiang2020}, or densities at certain point-of-interests within a city \cite{chang2020mobility}, potentially due to lack of access to fine-grained data. In this work, we leverage actual mobility -- as opposed to relying on popularity of an area (density) -- using Origin-Destination (OD) mobility graphs \cite{abrahamsson1998estimation}.

In contrast to traffic flow forecasting where the aim is to predict future traffic flow from the current flow, predicting risk score has an additional challenge of forecasting infections from the current spatial density and mobility behavior. That is, rather than predicting future mobility as is done for traffic forecasting, model needs to predict the future chance of infection that relies on both mobility behavior and rate of infection. To this end, we leverage a Hawkes process-based modelling procedure to predict future infections based on mobility patterns. Here, the Hawkes process-based modelling, which is extensively used to model event-based infection spread, provides a flexible way to incorporate the mobility data while allowing for explicit modelling of disease transmission \cite{Chiang2020}. In addition to the mobility-based model which leverages high-resolution location densities, we also develop a variant of our model which utilizes the mobility of infections from one region to another, i.e. \emph{infection mobility} to aid infection and risk predictions. We show that use of high-resolution mobility patterns along with infection mobility leads to improved infection and risk prediction.

\textbf{Reliable Risk Scores}. 
Popular disease prediction models employ compartmental modelling based on the classical Susceptible-Infected-Removed (SIR) model \cite{kermack1927} to explicitly model disease transmission. Although popular in practice due to their simplicity, these models rely on the \emph{homogeneous population mixing} (i.e., each person has equal probability of coming into contact with another individual) to learn the model parameters, and thus leading to coarse-grain reproduction number (R0) estimates \cite{Bertozzi16732, mohler2020analyzing, pei2020initial, zou2020epidemic, kamra2020polsird}. Nevertheless these are useful when finer-grain data is not available, and recent works leverage time-varying models to develop dynamic reproduction number for relatively finer scales, assuming homogeneous mixing \cite{kiamari2020}.  

One way to relax the homogeneous mixing assumption is to employ self-excitation-based Hawkes point process models \cite{meyer2015spatio, farrington2003branching} which are mathematically related to the compartmental models \cite{rizoiu2018sir}. 
To this end, a recent work \cite{Chiang2020} leverages the flexibility of the Hawkes process models to incorporate the demographic and mobility density indices for COVID-19 prediction.  As a result, existing models either a) do not incorporate high-resolution mobility data \cite{Chiang2020}, or b) use compartmental models which assume homogeneous mixing \cite{kiamari2020}. In this work, we show that incorporating high-resolution mobility data along with Hawkes process-based modeling leads to more reliable fine-grained risk scores.

\textbf{Accurately Evaluating the Risk Scores}. Another challenge in research on accurately modeling the spread of the disease is not having access to reliable fine-grain infection statistics. This is due to inaccurately reporting the number of infection in the real-world as not everyone who gets infected is tested for the virus \cite{infection_count}. To address this, existing work either a) ignore this issue and use the reported number of infections as ground truth \cite{kiamari2020, Chiang2020, Bertozzi16732, mohler2020analyzing} or try to circumvent it by b) using the number of deaths to study the spread of the disease \cite{Chiang2020}. In the former case, judging the accuracy of the model becomes difficult as it is not tested on the ground truth. In the latter case, the model will not be able to model how infections occur, but rather it only models number of deaths. Furthermore, fine-grained statistic for infection location is not publicly available. That is, at best we can know number of infections per county using public sources, which makes it difficult to assign risk scores to different locations (e.g., a zipcode or shopping center) within it.

To summarize, there are three main limitations with existing methods a) they assume homogeneous mixing of population i.e., do not incorporate mobility information, and/or b) assign coarse-grain scores (at county level at best) and c) they are difficult to evaluate. As a result, there is a need to build a technique to assign COVID-19 risk scores which is a) informative (fine-grain and time-varying), b) reliable (considers in-homogeneous mixing) and c) accurately evaluated. 


\subsection{Our Approach}

To address these challenges, we develop a Hawkes process-based spatiotemporal risk measure, dubbed \riskModelName{}, which utilizes the high-resolution mobility patterns in a city available via cell-phone location signals \cite{veraset} (a location signal is a record containing the location of the device at a particular point in time). We specifically show that such fine-grained mobility information can be used to assign risk scores that closely track the future infections in a region. We corroborate the efficacy of \riskModelName{} by showing its ability to track infections on simulated disease spread on real-world mobility for months of December 2019, January and March 2020. 

In particular, to evaluate \riskModelName{}, we use an agent-based simulation to compute the ground-truth number of infections. Our simulation, called \simlName{}, uses location signals from  large number of cell-phones across the US. \simlName{} simulates how the disease spreads across the population by utilizing this real-world high-resolution mobility patterns. Since \simlName{} utilizes the real-world data of people's movement, it can generate infection patterns for the population that are closer to the real-world. Furthermore, since the infection are generated by the simulation, we have access to the ground truth number of infections which allows for accurate evaluation of \riskModelName{}. 

Overall, our disease transmission simulation uses real-world mobility patterns and co-locations to emulate the spread of disease in real-world. Our point-process-based prediction model then leverages the mobility patterns to forecast future infections, the resulting intensity function of the Hawkes process-based model yields the risk score. Our main contributions are as follows. 

\begin{itemize}
    \item \textbf{\simlName{}: A Disease Spread Simulation Using Real-World Location Signals}. We build a co-location-based disease spread model using real-world location signals. \simlName{} emulates the disease transmission process in the real-world to generate infection patterns, and can be of independent interest for analysis of disease spread and intervention policies.
    \item \textbf{\riskModelName{}: High-resolution Spatiotemporal Risk Scores}. As opposed to previous works which utilize coarse-grain location density indices, we first develop a mobility-aware Hawkes process-based model -- \riskModelName{}$_{Mob}$ -- which leverages the high-resolution mobility patterns between different regions of a city to predict infections and assign spatiotemporal risk scores. Subsequently, we develop \riskModelName{}$_{Mob^+}$, which also accounts for the movement of infected population to improve the performance of the model.
    \item \textbf{Analyze the disease spread patterns}. We leverage \simlName{} and \riskModelName{} to  analyze the differences in disease spread at different conditions of real-world mobility rates, namely, before and after the March 2020 lockdown. Our results on cities across United States demonstrate that high-resolution mobility data can be used as a reliable public health tool to assess potential risk associated with parts of a city over time.
\end{itemize}

One possible application of \riskModelName{} risk score is to use it to reduce foot traffic to areas of high risk during the time the predicted risk score is high. Furthermore, since \riskModelName{} risk score leverages region-specific \textbf{aggregate} mobility patterns, it preserves privacy of device owners. In other words, although our \simlName{}, for evaluation purposes, utilizes individual-level co-locations, our \riskModelName{} risk prediction model can be used to assign risk scores while keeping the user data private.

\subsection{Overview and Organization}
Our approach can be summarized as follows. 

First, our agent-base simulation, \simlName{}, uses real-world location signals to generate infection patterns. Specifically, \simlName{} takes as input location trajectories of a number of individuals, as well as parameters that control how the disease spreads between the individuals to simulate how the infection progresses in the population over time. Finally, it outputs a list of infected individuals during the simulation together with the time and location they became infected. The details of \simlName{} are discussed in Sec. \ref{sec:simulator}.

Second, our \riskModelName{} takes as input the infection statistics (aggregate data, not individuals' locations or co-locations) that were generated from the simulation. It uses the statistics for the first $n$ days of the simulation, for a parameter $n$, to learn a model that is able to accurately predict number of infections for different locations, as well as provide a risk score for each location. Infection statistics from day $n$ until the end of the simulation are used to evaluate the learned model. The details of \riskModelName{} are discussed in Sec. \ref{sec:hawkes}.

 The rest of the paper is organized as follows. In Sec.~\ref{sec:exp}, we present the results of the analysis over different months for different cities across the United States. We also discuss related works in context of the proposed technique in Sec. \ref{sec:related}, and conclude our discussion in Sec.~\ref{sec:discuss}.

\section{\simlName{}}\label{sec:simulator}
Our agent-based simulation, \simlName{}, uses real-world mobility data and parameters from the existing literature on how COVID-19 spreads in a population to generate realistic infection patterns in the population. \simlName{} consists of a set of agents, collectively referred to as the population. Some agents are initially infected. Each agent moves and co-locates with other agents based on real-world fine-grain mobility data provided by Veraset \cite{veraset}. As the agents move based on the location signal data (described in Sec. \ref{sec:sim:mobility}), according to their co-locations, they get infected and spread the disease following a variation of the SIR model and using incubation and generation periods of COVID-19 reported in the literature. The output of \simlName{} is the information on which agents get infected, when and where. Thus, \simlName{} is comprised of three components. (1) The location of each agent at each point in time is determined by a predefined \textit{mobility pattern}. (2) The spread of the disease amongst the agents is determined by a \textit{transmission model}. (3) Who is initially infected is determined by the initialization conditions. We next discuss each of the components and the relevant implementation details.

\begin{figure}
\begin{minipage}{0.45\textwidth}
    \centering
    \includegraphics[width=\textwidth]{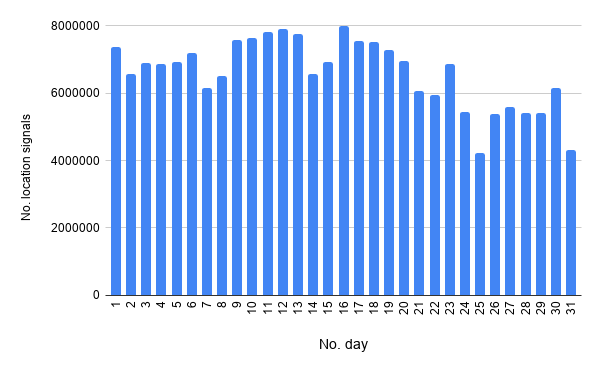}
    \caption{No. of location signals per Day in Manhattan in Dec 2019.}
    \label{fig:no_checkins_manhattan}
\end{minipage}\hfill
\begin{minipage}{0.54\textwidth}
    \centering
    \includegraphics[width=\textwidth]{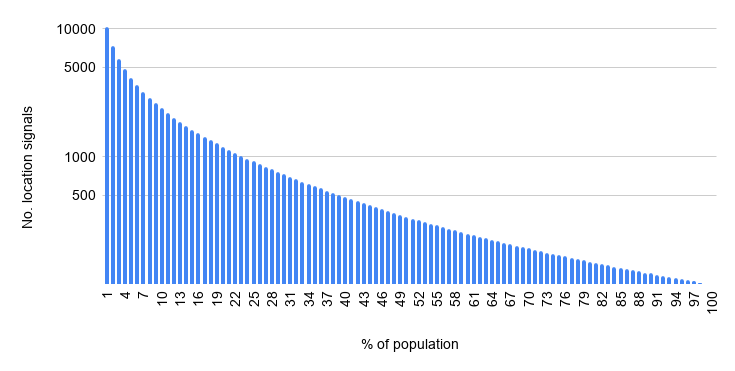}
    \caption{Distribution of location signals in Manhattan in Dec 2019.}
    \label{fig:distribution_checkins_manhattan}
\end{minipage}
\end{figure}
\subsection{Mobility Pattern}\label{sec:sim:mobility}
The mobility pattern determines where each agent is at each point in time during the simulation. We use real-world location signals provided by Veraset \cite{veraset}. Veraset \cite{veraset} is a data-as-a-service company that provides anonymized population movement data collected
through GPS signals of cell-phones across the US. We were provided access to this dataset for the months of December, January and (from 5th to 26th of) March. For a single day in December, there are 2,630,669,304 location signals across the US. Each location signal corresponds to a \texttt{device\_id} and there are 28,264,106 \texttt{device\_id}s across the US in that day. We assume each \texttt{device\_id} corresponds to a unique individual . Fig. \ref{fig:no_checkins_manhattan} shows the number of daily location signals recorded in the month of Dec. 2019 in the area of Manhattan, New York. Furthermore, Fig. \ref{fig:distribution_checkins_manhattan} shows the distribution of location signals across individuals in Manhattan in Dec 2019. A point $(x, y)$ in Fig. \ref{fig:distribution_checkins_manhattan} means that $x$ percent of the individuals have at least $y$ location signals in the month of Dec. in Manhattan. Using this real-world location data, we are able to create infection patterns for different cities and at different periods of time. This allows us to study the hypotheses that concern the spread of the disease for different cities and at different times. 

Detailed statistics about the subset of the Veraset data we used for our experiments are discussed in Sec. \ref{sec:exp}. Here, we discuss our general approach of using real-world location signals for \simlName{}, independently of the actual dataset used. We consider a dataset such that each record in the dataset consists of an anonymized \texttt{user\_id}, \texttt{latitude}, \texttt{longitude} and \texttt{timestamp} and is a location signal of the user with id \texttt{user\_id} at the specified time and location. We consider each individual to be an agent in the simulation and for each agent, we have access to their latitude and longitude for various timestamps. Sorting this by time, we obtain a trajectory of location signals of an agent over time, denoted by the sequence $\langle c_1, c_2, ..., c_k\rangle$. For two consecutive location signals, $c_i$ and $c_{i+1}$ of an agent at times $t_{i}$ and $t_{i+1}$, we assume the agent is at the location specified by $c_i$ from time $t_i$ to $t_{i+1}$. Using this interpolation, together with our location data, we have access to the location of every agent at every point in time during the simulation. We note that, although more sophisticated interpolations are possible, this falls beyond the scope of this paper, since our focus is on providing risk scores for different locations.

\subsection{Transmission Model}
\begin{figure}
\begin{minipage}{0.49\textwidth}
    \centering
    \includegraphics[width=\textwidth]{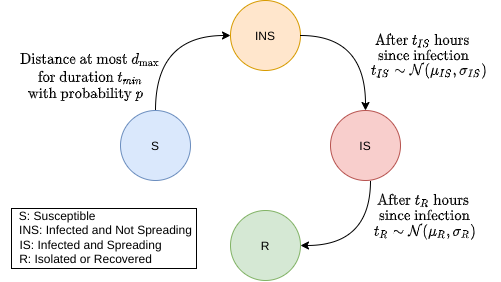}
    \caption{Transmission Model}
    \label{fig:transmission}
\end{minipage}
\begin{minipage}{0.49\textwidth}
    \centering
    \vspace{10pt}
    \includegraphics[width=\textwidth]{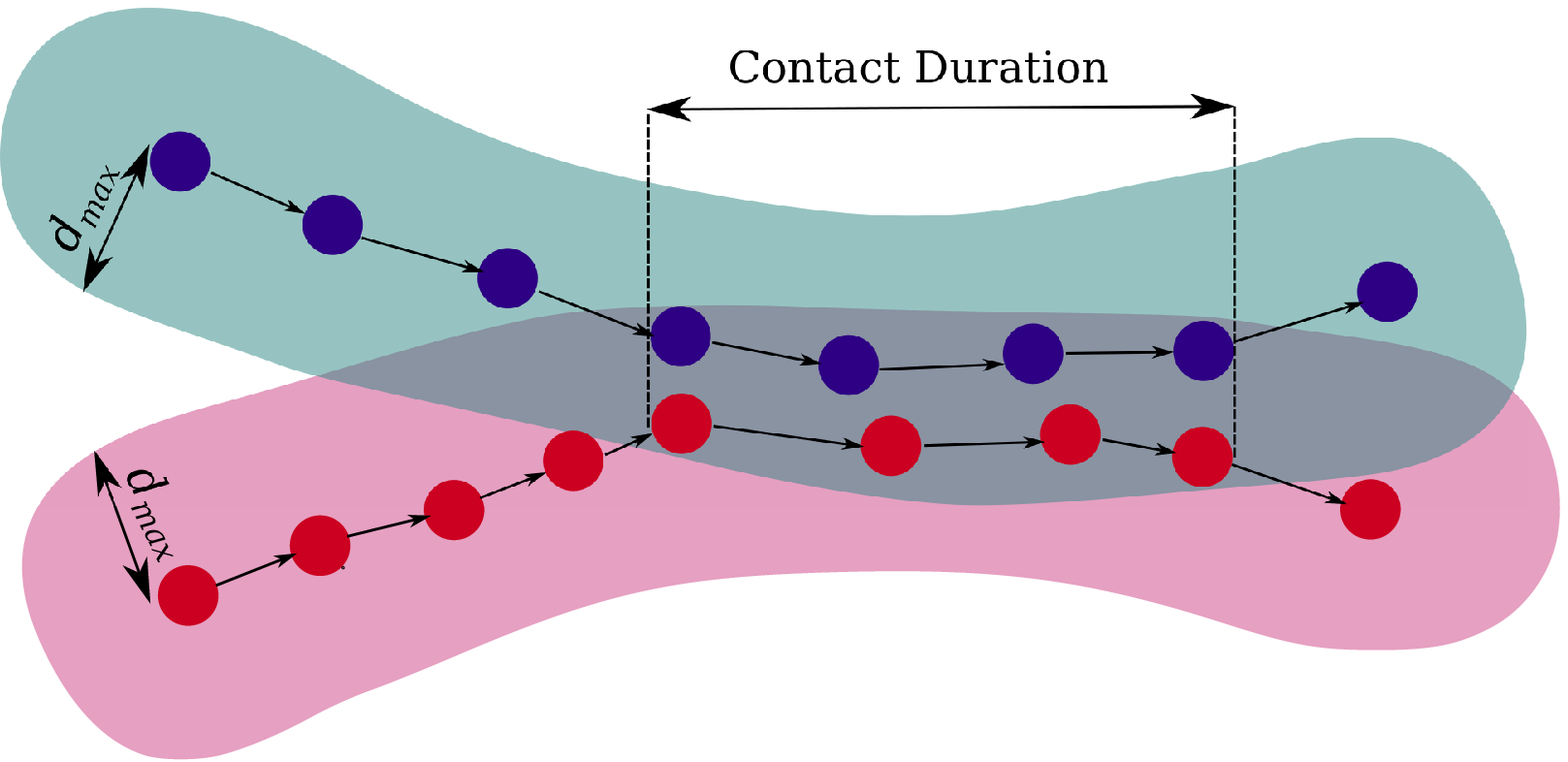}\\
    \vspace{16pt}
    \caption{Transmission of disease through a co-location. Arrows show movement in time.}
    \label{fig:contact}
\end{minipage}
\end{figure}

Agents belong to either one of the following compartments: Susceptible (S), Infected and Not Spreading (INS), Infected and Spreading (IS) and Isolated or Recovered (R). An agent is said to be infected if they are in either INS or IS compartments. Intuitively, a susceptible agent can contract the disease, an INS agent is infected but cannot spread the disease, an IS agent is infected and can spread the disease and an Isolated or Recovered agent cannot spread the disease or contract it anymore. Figs. \ref{fig:transmission} and \ref{fig:contact} illustrate the transmission model. 

\subsubsection{Compartments}\label{sec:compartments} All agents are initially Susceptible, with the exception of the agents that are infected initially as discussed in Sec. \ref{sec:initialization}. We call the compartment an agent is in the agent's status. Assume an agent gets infected at time $t$ based on the dynamics discussed in Sec. \ref{sec:transmision}. Thus, at time $t$, the agent changes its status from Susceptible to Infected and Not Spreading. Consequently, at time $t+t_{IS}$, the agent becomes Infected and Spreading. Furthermore, at time $t+t_{R}$, the agent becomes Isolated or Recovered, where $t_{IS}\sim \mathcal{N}(\mu_{IS}, \sigma_{IS})$ and $t_{R}\sim\mathcal{N}(\mu_{R}, \sigma_{R})$, where $\mu_{IS},~ \sigma_{IS}, \mu_{R}$ and $\sigma_{R}$ are the parameters of the model. These parameters can be set based on existing research on COVID-19 \cite{ferretti2020quantifying, he2020temporal, li2020early}. For instance, \cite{he2020temporal} mentions that `` 1\% of transmission would occur before 5 days and 9\% of transmission would occur before 3 days prior to symptom onset'', while \cite{li2020early} finds that incubation period (time from infection until onset of symptoms) has mean 5.2 days. Using this, we can find the parameter for the model that match the empirical evidence the best. We discuss the specific parameter setting used in our experiments in Sec. \ref{sec:exp}. We note that, first, $\mathcal{N}(\cdot)$ is a truncated normal distribution, such that if the sampled $t_{IS}$ or $t_{R}$ are less than zero, they are discarded and a new sample is obtained. Furthermore, if $t_{R} < t_{IS}$, the agent never spreads the disease, and moves directly from INS to the R compartment (the probability of either of these happening is very low based on the parameter setting chosen).

\subsubsection{Transmission Dynamics}\label{sec:transmision} Only IS agents infect other agents, and only S agents get infected. Simply put, if an IS agent is within distance $d_{max}$ of an S agent for duration at least $t_{min}$ then, with probability $p$, the IS agent infects the S agent. $d_{max}$, $t_{min}$ and $p$ are the parameters of the transmission model. The model is designed to mimic how the disease spreads in the physical world, following the works of \cite{bromage2020risks, wiersinga2020pathophysiology, chu2020physical}, where prolonged close-range contacts is the main source of transmission of the disease. Specifically, consider an IS agent, $u$, and an S agent $v$. At any time, $t$, during the simulation, consider the situation when the distance between $u$ and $v$ becomes at most $d_{max}$ (i.e., distance between $u$ and $v$ was larger than $d_{max}$ right before time $t$) from time $t$ until time at least $t+t_{min}$. Then, a number, uniformly at random, is generated in the range $[0, 1]$. If the number is at most $p$, then $u$ infects $v$. When $v$ becomes infected, as explained in Sec. \ref{sec:compartments}, its status changes from S to INS, and eventually to IS and R.

\subsection{Initialization and Implementation}
\subsubsection{Initialization}\label{sec:initialization}. For a parameter $n_{init}$, we infect $n_{init}$ number of agents at the beginning of the simulation uniformly at random. The initial infections are treated as if the agent was infected by another agent. That is, an initially infected agent has initially the INS status and becomes IS after time $t_{IS}$ and recovered after $t_R$, where $t_{IS}$ and $t_R$ are sampled from the normal distributions discussed before. 

\subsubsection{Implementation} Our implementation \simlName{} first sorts all the location signals based on time. Subsequently, location signals are processed one by one in the order of time. Every location signal corresponds to an agent moving from an old location to a new location. As such, if the agent is susceptible, we check if there are any IS agents within $d_{max}$ of the new location. If the agent is IS, we check if there are any S agents within its $d_{max}$. We call it a co-location if an two agents are within $d_{max}$ distance of each other. Next, for any co-location between an S agent $u$ and an IS agent $v$, we check if it lasts for at least $t_{min}$. If it does, then, with probability $p$, we infect $u$. We use a grid to index the location of the agents. Given a new location signal at a new location, $(x, y)$, we perform a range search on the grid to find agents within $d_{max}$ of $(x, y)$. For each agent that matches the range search predicate, we then check if their co-location lasts for at least $t_{min}$. This can be optimized further by building a multidimensional index on both time and location.

\section{Learning Spatiotemporal Risk Scores with \riskModelName{}}
\label{sec:hawkes}
We now describe our Hawkes process model, \riskModelName{}, which leverages location data along with mobility patterns in a particular region to assign spatiotemporal risk scores. For a given city, we first form clusters  $c\in \mathcal{C}$ using $k$-means clustering algorithm based on location signals in the city. The number of such clusters can be chosen according to the desired spatial resolution \footnote{Provided each cluster contains adequate number of location signals and infections for reliable learning.}. 

\subsection{Mobility-aware Modelling}
\label{sec:form_features}
\riskModelName{} leverages the Origin-Destination (OD) matrix to represent the flow between two clusters $c$ and $c'$ in a city. OD matrices are a popular way to encode spatial traffic flow information between two nodes in a transportation graph \cite{abrahamsson1998estimation}. Specifically, let $\mathcal{G}^t(V,E)$ denote a directed graph, where the edge weight $w_{i\rightarrow j}(t)$ encodes the traffic volume from cluster $i$ to $j$ between time $(t-1)$ and $t$. The OD matrix $W (t) \in \mathbb{R}^{|\mathcal{C}| \times |\mathcal{C}|}$ represents the traffic flows between different clusters $c \in \mathcal{C}$ in a city at time $t$. Here, the diagonal elements $w_{i\rightarrow i}(t)$ denote the traffic generated in a cluster $i$. We use this traffic flow information to inform \riskModelName{}. 

We form the the mobility feature vector at time $t$,  $\boldsymbol{m}_{\mathcal{G}_c}^t \in \mathbb{R}^{f}$, elements of which encode the different mobility features derived from the OD matrix $W$. Here, we develop two Hawkes process-based models a) \riskModelName{}$_{Mob}$, and b) \riskModelName{}$_{Mob^+}$, depending upon the traffic flow features. For \riskModelName{}$_{Mob}$ we set
\begin{align}
\tag{\riskModelName{}$_{Mob}$ Features}
    \boldsymbol{m}_{c}^t  = 
    \begin{bmatrix}
    w_{c\rightarrow c}(t),  ~w_{to-c}(t),  ~w_{from-c}(t)
    \end{bmatrix}^\top, 
\end{align}
where 
\begin{align}
    w_{to-c}(t) := \sum\limits_{c'\in \mathcal{C}\backslash c} w_{c'\rightarrow c}(t) \hspace{1cm}\text{and} \hspace{1cm} w_{from-c}(t) := \sum\limits_{c'\in \mathcal{C}\backslash c} w_{c\rightarrow c'}(t) \notag,
\end{align}
and $``\backslash''$ denotes the set difference operator. \sr{In effect, \riskModelName{}$_{Mob}$ considers the net traffic to and from a clusters along with the location signals' density, while being agnostic to the infections in each cluster. }

\sr{To make the model infection-aware,} we design \riskModelName{}$_{Mob^+}$ which has additional mobility-based features to account for the infections at the origin cluster. Let $I_c(t)$ denote the infections in cluster $c$ at time $t$, and further let $I_{-c}(t)$ denote the total infections (over all clusters) except those in cluster $c$, i.e., 
\begin{align}
    I_{-c}(t) = \sum\limits_{c'\in \mathcal{C} \backslash c} I_{c'}(t). \notag
\end{align}
We formulate the \emph{infection mobility} to a cluster $c$ as 
\begin{align}
     Im_{to-c}(t) :=  \dfrac{w_{to-c}(t)}{\sum\limits_{c'\in \mathcal{C}\backslash c} \left( w_{to-c'}(t) + w_{from-c'}(t) \right)} \cdot I_{-c}(t) . \notag
\end{align}
Here, $Im_{to-c}(t)$ captures the proportion of infections travelling to a cluster $c$. Similarly, we also form $Im_{of-c}(t)$ to weigh the total number of infections in a cluster $c$ by the ratio of self-mobility and traffic from other clusters.
\begin{align}
     Im_{of-c}(t) :=  \dfrac{w_{c\rightarrow c}(t)}{ w_{to-c}(t)} \cdot I_{c}(t). \notag
\end{align}
With these features, we form the feature set for \riskModelName{}$_{Mob^+}$ as follows.
\begin{align}
\tag{\riskModelName{}$_{Mob^+}$ Features}
    \boldsymbol{m}_{c}^t  = 
    \begin{bmatrix}
    w_{c\rightarrow c}(t),  ~w_{to-c}(t),  ~w_{from-c}(t), ~Im_{to-c}(t), ~Im_{of-c}(t)
    \end{bmatrix}^\top
\end{align}

\subsection{Incorporating Mobility Features into Hawkes Process}
Now, given the daily infections at each cluster $c\in \mathcal{C}$, i.e. a realization of a point process $N_c(t)$ on $[0, T]$ for $T< \infty$, at timestamps $\mathcal{T}_c = \{t_1^c, t_2^c, \dots t_n^c\}$, we model the rate of new cases at time $t$, $\lambda_c(t) $ associated with a cluster $c$ by incorporating the corresponding traffic mobility features $\boldsymbol{m}_c^{t}$ developed in Sec.~\ref{sec:form_features} as
    \begin{align}
    \label{eq:intensity}
        \lambda_c(t) = \mu_c + \sum_{t>t_j, t_j \in \mathcal{T}} R_c^{t_j}(\boldsymbol{m}_c^{t_j - \Delta}, \theta)~wbl(t_i-t_j),
    \end{align}
    where, $\mu_c$ is known as the \emph{background rate}, and captures the inherent proclivity of a cluster to produce infections. We use Weibull distribution, $wbl(\cdot)$ with shape $\alpha$ and scale $\beta$ to model the inter-event time, which specifies the influence of past events. The time-delay parameter $\Delta$ is used to account for the delay between the mobility and the infections.
    
    The mobility features are used to model the time-varying cluster-dependent reproduction number $R_c^{t}(\boldsymbol{m}_c^{t-\Delta}, \theta)$. In particular, we model the dynamic reproduction number $R_c^{t}$ as a Poisson regression task \citep{Chiang2020}, including the traffic-flow based mobility features for each cluster as 
    \begin{align}
    \label{eq:exp_R}
        \boldsymbol{E}_R[R_c^{t_j}| \boldsymbol{m}_c^{t_j - \Delta}, \theta] = \exp(\theta^\top\boldsymbol{m}_c^{t_j - \Delta} ),
    \end{align}
    where $\boldsymbol{E}[\cdot]$ denotes the expectation operator.
    
    The mobility and cluster-dependent reproduction number $R_c^{t}(\boldsymbol{m}_c^{t-\Delta}, \theta)$ can be viewed as the average number of secondary infections \emph{caused} by a primary infection (in the Granger sense \cite{Granger1969, eichler2017graphical,yuan2019multivariate}). In addition, the first term $\mu_c$ helps in modelling the effect of factors not captured by the mobility-dependent second term.
    
    Overall, \riskModelName{} incorporates the actual high-resolution mobility patterns within a city, which enables us to learn informative spatiotemporal risk scores (developed in Sec.~\ref{sec:risk}). As compared to related mobility-based methods (such as \citep{Chiang2020}) which rely only on mobility density, \riskModelName{} leverage fine-grained flow information. \sr{The main task now is to infer the model parameters $\theta$ of \riskModelName{} using the mobility features and the infections. We now describe the Expectation Maximization-based inference procedure to learn $\theta$.} 
    
    \subsection{Expectation Maximization-based Inference Procedure}
    We adopt an Expectation Maximization-based approach to infer the parameters $\theta$. Our algorithm (outlined in Algorithm~\ref{alg:our_alg}) provides an iterative way to evaluate the maximum-likelihood estimates of \riskModelName{}. We begin by introducing latent variables $Y$ in order to model the unobserved variables of our model. To this end, we begin by writing the likelihood $\mathcal{L}(\Theta; X)$ for Hawkes process given data $X = \left( \{t_j^c\}_{j=1, c=1}^{|\mathcal{T}_c|, |\mathcal{C}|},  \{\boldsymbol{m}_c^t\}_{c=1, t=1}^{|\mathcal{C}|, |\mathcal{T}|}\right)$ and model parameters $\Theta = \left(\{\mu_c\}_{c=1}^{\mathcal{C}}, \theta\right)$ as
    \begin{align}
        \mathcal{L}(\Theta; X) =  \prod_{c=1}^{|\mathcal{C}|}\prod_{i=1}^{n} \lambda_c(t_i) \exp^{-\int_0^T \lambda_c(t)dt}.
    \end{align}

 For the EM procedure, we introduce latent variables $Y^{c}_{ij}$ to indicate that an event $j$ is an off-spring event $i$ in cluster $c$, and $Y_{ii}^{c}$ to denote that it was generated by a background event (in $c$), to formulate the \emph{complete} data log-likelihood as 
    \begin{align}
    \label{eq:log_likeli}
       \log(\mathcal{L}(\Theta; X, Y)) =  \sum_{c=1}^{|\mathcal{C}|} \sum_{i=1}^{n} Y^{c}_{ii} \log(\mu_c) + Y^{c}_{ij} \log \left( \sum_{t_i>t_j, t_j \in \mathcal{T}} R_c^{t_j}(\boldsymbol{m}_c^{t_j - \Delta}, \theta)~wbl(t-t_j)\right) - \int_0^T \lambda_c(t) dt.
    \end{align}
        
\begin{algorithm}[t]
\caption{\textbf{\textsc{\riskModelName{}}}: Mobility-based Hawkes process Model using Poisson Regression} \label{alg:our_alg}
\SetAlgoLined
\begin{flushleft}
\begin{algorithmic}
\noindent\KwInput{
The timestamps set $\mathcal{T}_c = \{t_j^c\}_{j=1}^{|\mathcal{T}|}$ for each cluster $c\in \mathcal{C}$ (e.g. daily infections in each cluster within a city).
Daily Origin-Destination matrix $\{W(t)\}\}_{t=0}^{T}$ consisting of mobility patterns within a city to form mobility features $\{\boldsymbol{m}_c^{t}\}_{c=1, t=0}^{|\mathcal{C}|, |\mathcal{T}|}$. Weibull shape and scale parameters $\alpha$ and $\beta$,  time delay parameter $\Delta$, and the tolerance parameter $\delta$.}\\ \vspace{3pt}
\noindent\KwOutput{Estimates $\theta$ and $\{\mu_c\}_{c=1}^{\mathcal{C}}$. } \\ \vspace{3pt}
\noindent\KwInit{$\mu_c \leftarrow 0.5$ for all $c\in\mathcal{C}$,  $R_c^t \leftarrow 1$ for all $c\in\mathcal{C}$ and $t\in\mathcal{T}$, $T \leftarrow \max \mathcal{T}$, and $k =1$.} \\ \vspace{3pt}
\While{$\|\Delta\theta\| \geq \delta$ and $\|\Delta \mu_c\| \geq \delta$ for all $c\in\mathcal{C}$}{
\vspace{3pt}
\textbf{Expectation Step}:\\
\indent \indent \For{$\forall~i\geq j$ and $0 < i$, $j <T$ and $\forall~c \in \mathcal{C}$}{
\uIf{i>j}{
$p_c(i,j) = \dfrac{R_c^{t_j}(\boldsymbol{m}_c^{t_j - \Delta}, \theta)~wbl(t-t_j)}{\lambda_c(t_i)}$
}\uElseIf{ i = j }{ 
$p_c(i,i)= \dfrac{\mu_c}{\lambda_c(t_i)}$}
}
\vspace{3pt}
\textbf{Maximization Step}:\\

\indent \indent \indent\emph{Update $\theta$}: \indent$\hat{\theta} \leftarrow \underset{\theta}{\arg\max} \sum_{c=1}^{|\mathcal{C}|} \left(\sum_{j=1}^{n} P_c(j) \theta^\top\boldsymbol{m}_c^{t_j - \Delta} - \exp(\theta^\top\boldsymbol{m}_c^{t_j - \Delta})\right)$ \\
\indent \indent \indent\emph{Update $\mu_c$s}: 
\indent $\hat{\mu_c} \leftarrow \sum_{i=1}^{n} \tfrac{p_c(i,i)}{T}$ for all $c \in \mathcal{C}$.\\ \vspace{3pt}
\indent \indent $k \leftarrow k + 1$ \\ \vspace{3pt}
}
\end{algorithmic}
\end{flushleft}
\end{algorithm}
    \subsubsection{Expectation Step}
    Since both $\log(\mathcal{L}(\Theta; X, Y))$ and $Y$ are random variables, at the $k$-th iteration in the E-step we evaluate the expectation function $Q(\Theta, \Theta^{k-1})$ as
    \begin{align}
    \label{eq:Q}
        Q(\Theta, \Theta^{k-1}) &= \boldsymbol{E}_Y[\log(\mathcal{L}(\Theta; X, Y))|X,\Theta^{k-1}], \\
        &= \int \log(\mathcal{L}(\Theta; X, Y)) f(Y|X, \Theta^{k-1})dY \notag,
    \end{align}
    where $\Theta^{k-1}$ are the estimated parameters at the $(k-1)$-th iteration, and $f(Y|X, \Theta^{k-1})$ is the conditional distribution of $Y$ given $X$ and $\Theta^{k-1}$. Specifically, as in \cite{Chiang2020}, we estimate the probability $p_c(i,j) $ as follows
    \begin{align}
    p_c(i,j) := \boldsymbol{E}_Y[Y^{c}_{ij}|X,\Theta^{k-1}] = \dfrac{R_c^{t_j}(\boldsymbol{m}_c^{t_j - \Delta}, \theta)~wbl(t-t_j|\alpha,\beta)}{\lambda_c(t_i)} \notag
    \end{align}
    and $p_c(i,i)$ as
     \begin{align}
    p_c(i,i):= \boldsymbol{E}_Y[Y^{c}_{ii}|X,\Theta^{k-1}] = \dfrac{\mu_c}{\lambda_c(t_i)}. \notag
    \end{align}
    \subsubsection{Maximization Step}
    Based on the probabilities $p_c(i,j)$ and $p_c(i,i)$, we estimate the parameters by maximizing $Q(\Theta, \Theta^{k-1})$ in \eqref{eq:Q} w.r.t. to each parameter\footnote{The Weibull shape and scale parameters $\alpha$ and $\beta$ can also be learned by adding additional EM-based update steps when adequate data samples are available, i.e. $|\mathcal{T}|$ is large enough; see \cite{Chiang2020}. Since we run our simulation over relatively shorter time scales, in our formulation we set $\alpha$ and $\beta$ based on grid-search.}. With this, we arrive at the following update steps for each parameter \cite{Chiang2020}. Specifically, we learn the parameters $\theta$ via Poisson regression as 
    \begin{align}
        \hat{\theta} := \underset{\theta}{\arg\max} \sum_{c=1}^{|\mathcal{C}|} \left(\sum_{j=1}^{n} P_c(j) \theta^\top\boldsymbol{m}_c^{t_j - \Delta} - \exp(\theta^\top\boldsymbol{m}_c^{t_j - \Delta})\right), \notag
    \end{align}
    where $P_c(j) = \sum_{i=j+1}^{n} p_c(i,j)$, and the following closed form update for the background rate parameters:
    \begin{align}
        \hat{\mu_c} := \underset{\mu_c}{\arg\max} \sum_{i=1}^{n} p_c(i,i) \log(\mu_c) - \int_0^T \mu_c dt = \sum_{i=1}^{n} \tfrac{p_c(i,i)}{T}. \notag
    \end{align}


    \subsection{Characterizing Risk}
    \label{sec:risk}
    The dynamic reproduction number (commonly referred to as $R0$) has been used to assign risk scores to communities \cite{kiamari2020}. Indeed as explored by \cite{Chiang2020}, the reproduction number is dependent on the mobility density over time. \sr{However, since the reproduction number is highly sensitive to parameter choices and models \cite{delamater2019, li2011failure}, \riskModelName{} leverages both the dynamic reproduction number $R_c^t$ and the background rate $\mu_c$ associated with a cluster $c$ to assign risk scores.}  
    Specifically, we propose a spatiotemporal risk score based on the intensity function $\lambda_c(t)$ of \riskModelName{}. Let $\Lambda \in \mathbb{R}^{|\mathcal{C}| \times T}$ be a matrix where $\Lambda(c,t) = \lambda_c(t)$. Then, we define our risk score $\rho \in [0,1]$ for a cluster $c$ at time $t$ as
            \begin{align}
            \label{eq:risk}
                Risk ~Score ~\rho_c(t) = \dfrac{\lambda_c(t) - \underset{c' \in \mathcal{C}, t'\in T}{\min}~ \Lambda (c', t')}{\underset{ c' \in \mathcal{C}, t'\in T}{\max}~ \Lambda(c', t') - \underset{c' \in \mathcal{C}, t'\in T}{\min}~ \Lambda(c', t')}.
            \end{align}
    This essentially scales the intensity function of the disease relative to the intensities in other clusters over time, and alleviates the issues associated with calculating accurate reproduction number.

\begin{table}[t]
    \centering
    \begin{minipage}{0.47\textwidth}
    \centering
    \resizebox{0.9\textwidth}{!}{
    \begin{tabular}{l|c|c|c}
        \textbf{City} & \textbf{December} & \textbf{January} & \textbf{March}  \\\hline
        San Francisco & 116$\times 10^6$ &96$\times 10^6$ & 61$\times 10^6$\\\hline
        Miami & 75$\times 10^6$ & 105$\times 10^6$& 64$\times 10^6$\\\hline
        Chicago &135$\times 10^6$ & 175$\times 10^6$ & 107$\times 10^6$\\\hline
        Houston & 135$\times 10^6$ &182$\times 10^6$ & 105$\times 10^6$
    \end{tabular}}
    \caption{Total No. location signals per Month}
    \label{tab:checkin_data}
    \end{minipage}
    \hfill
    \begin{minipage}{0.47\textwidth}
    \centering
    \resizebox{0.87\textwidth}{!}{
    \begin{tabular}{l|c|c|c}
        \textbf{City} & \textbf{December} & \textbf{January} & \textbf{March}  \\\hline
        San Francisco & 62$\times10^3$ &  72$\times10^3$&  55$\times10^3$\\\hline
        Miami & 42$\times10^3$ & 55$\times10^3$ & 46$\times10^3$\\\hline
        Chicago & 76$\times10^3$ & 106$\times10^3$ & 90$\times10^3$\\\hline
        Houston & 69$\times10^3$ & 97$\times10^3$& 78$\times10^3$
    \end{tabular}}
    \caption{No. Agents per Month}
    \label{tab:agent}
    \end{minipage}
\end{table}

\begin{table}[t]
    \centering
    \resizebox{0.9\textwidth}{!}{
    \begin{tabular}{c|c|c}
        \textbf{Parameter} & \textbf{Description} & \textbf{Value} \\\hline
        $d_{max}$ & Maximum distance for a co-location & $\sim11m$\\\hline
        $t_{min}$ & Minimum duration for infection & $1h$\\\hline
        $p$ & Probability of infection & 1\\\hline
        $n_{init}$ & Number of agents initially infected & 1,000\\\hline
        $\mu_{IS}$ & Average number of days for an agent to become IS from infection & 5\\\hline
        $\sigma_{IS}$ & Std. deviation of number of days for an agent to become IS from infection & 1\\\hline
        $\mu_{R}$ & Average number of days for an agent to become R from infection & 12 \\\hline
        $\sigma_{R}$ & Std. deviation of number of days for an agent to become R from infection & 2.4
    \end{tabular}}
    \caption{\simlName{} Parameter Setting}
    \label{tab:sim_setting}
    \vspace{-0.5cm}
\end{table}

\section{Experimental Results}\label{sec:exp}
\label{sec:exps} 
We now evaluate and compare the performance of our proposed method with competing techniques based on the accuracy of their infection and risk prediction over different months and cities in the United States (US).

\subsection{Data} The statistics of our dataset is summarised in Tables \ref{tab:checkin_data} and \ref{tab:agent}. We obtained the data from Veraset \cite{veraset} a data-as-a-service company that provides anonymized population movement data collected through GPS signals of cell-phones across the US. The obtained dataset consists of location signals across the US for the time-periods December 1, 2019 to January 31, 2020, as well as March 5, 2020 to March 26, 2020. For each of the cities or counties considered, we first use a range defined by a rectangle that roughly covers the area of the city or county to select the location signals that fall within the area. Each record in the dataset consists of \texttt{anonymized\_device\_id}, \texttt{latitude}, \texttt{longitude}, \texttt{timestamp} and \texttt{horizontal accuracy}.  We assume each \texttt{anonymized\_device\_id} corresponds to a unique individual. We discard any location signal with \texttt{horizontal accuracy} of worse than 25 meters. Furthermore, we filter out individuals with less than 100 location signals for every one-month period considered. 
\begin{figure}
    \centering
    \resizebox{1\textwidth}{!}{
    \begin{tabular}{cccc}
    & \hspace{0pt} Day $10$ & \hspace{-5pt} Day $15$ & \hspace{-5pt} Day $20$ \\ 
      \rotatebox{90}{\hspace{0.9cm} Dec. `19} \hspace{-15pt}& \hspace{-8pt} \includegraphics[width=0.3\textwidth]{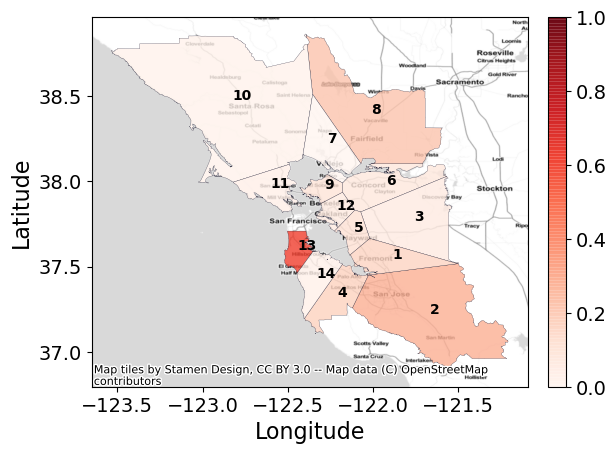}  & \hspace{-10pt}
       \includegraphics[width=0.3\textwidth]{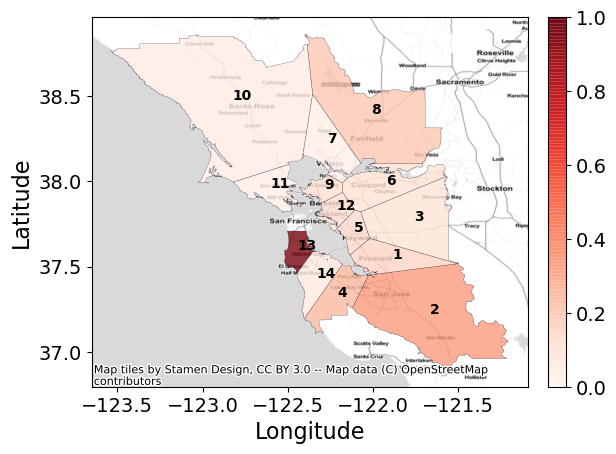} & \hspace{-10pt}
       \includegraphics[width=0.3\textwidth]{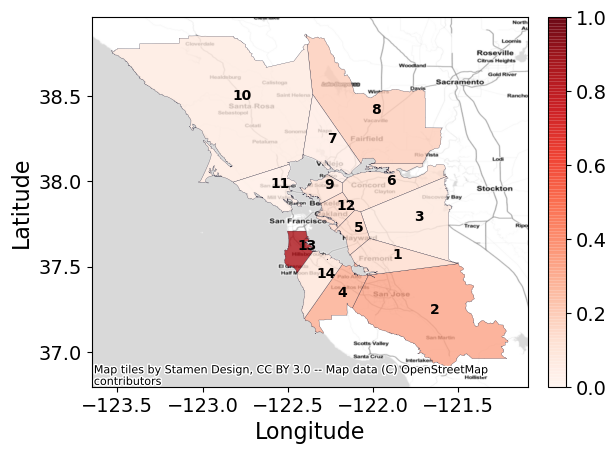} \hspace{-5pt}\\
       & \hspace{0pt} {\footnotesize (a-i)} & \hspace{-5pt} {\footnotesize (a-ii)} & \hspace{-5pt} {\footnotesize (a-iii)}\\
       \rotatebox{90}{\hspace{0.9cm} Jan. `20} \hspace{-15pt}& \hspace{-8pt} \includegraphics[width=0.3\textwidth]{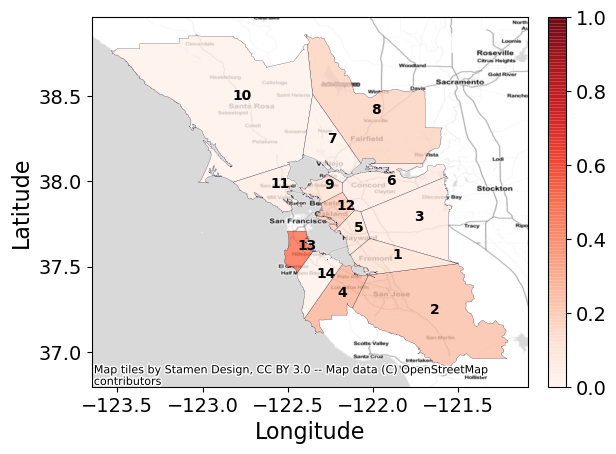} & \hspace{-10pt}
       \includegraphics[width=0.3\textwidth]{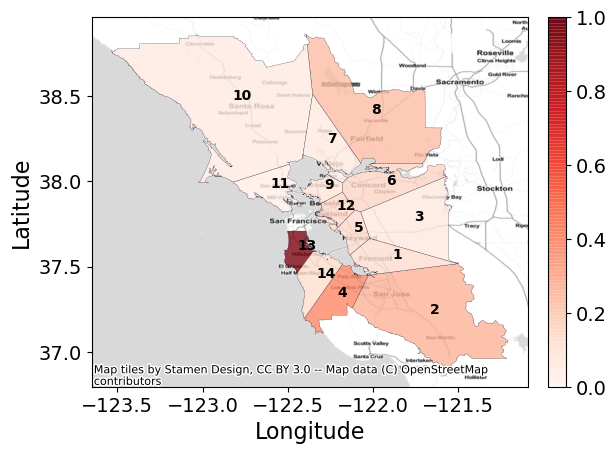} & \hspace{-15pt}
       \includegraphics[width=0.3\textwidth]{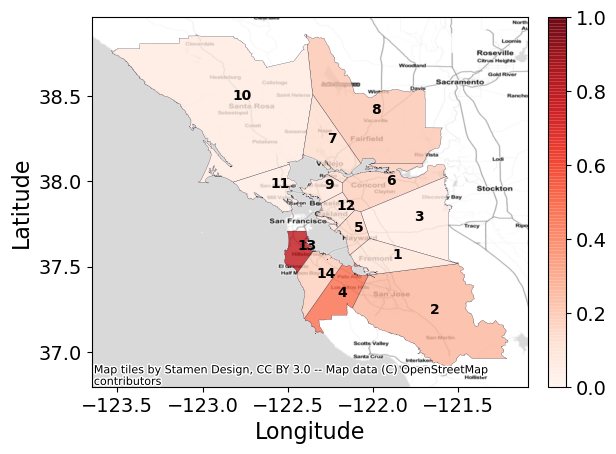}\hspace{-5pt} \\
          & \hspace{0pt} {\footnotesize(b-i)} & \hspace{-5pt} {\footnotesize (b-ii)} & \hspace{-5pt} {\footnotesize (b-iii)}\\
     \rotatebox{90}{ \hspace{0.9cm} Mar. `20} \hspace{-15pt}& \hspace{-8pt} \includegraphics[width=0.3\textwidth]{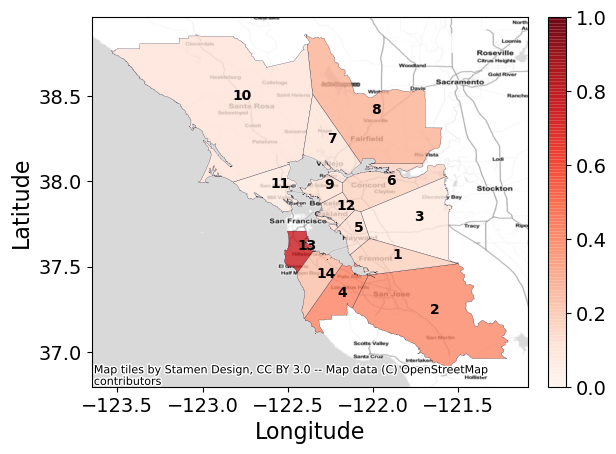} &
      \hspace{-10pt} \includegraphics[width=0.3\textwidth]{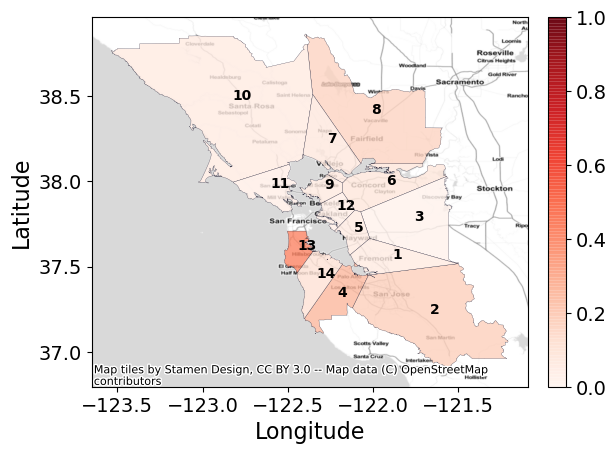} & \hspace{-10pt}
       \includegraphics[width=0.3\textwidth]{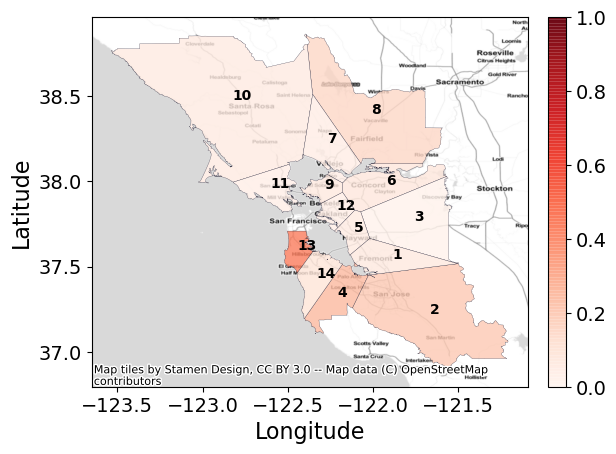} \\
        &  \hspace{0pt} {\footnotesize (c-i)} & \hspace{-5pt} {\footnotesize(c-ii)} & \hspace{-5pt} {\footnotesize(c-iii)}\\
        &\hspace{-17pt} \includegraphics[width=0.29\textwidth]{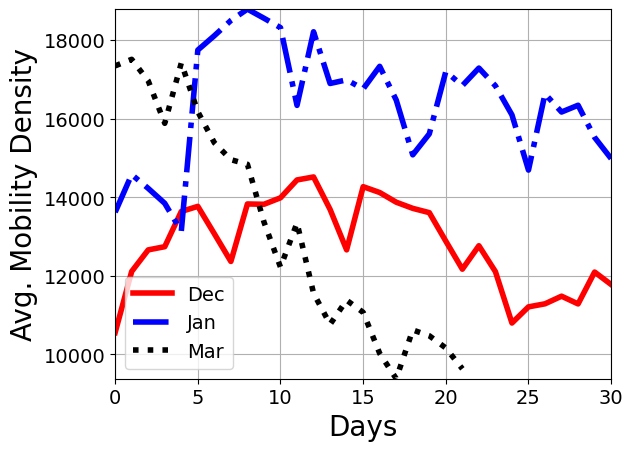}  & \hspace{-17pt}
       \includegraphics[width=0.27\textwidth]{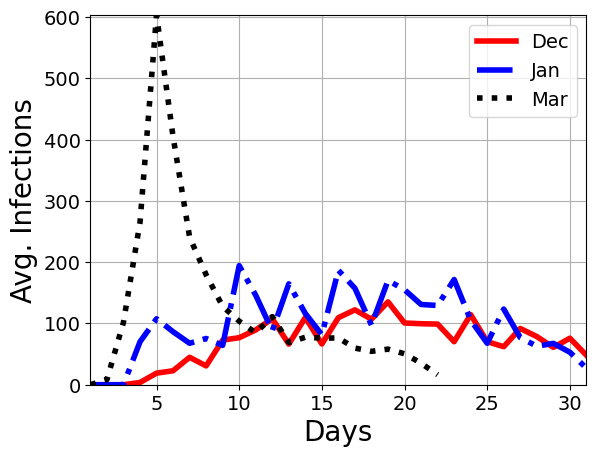} & \hspace{-17pt}
       \includegraphics[width=0.27\textwidth]{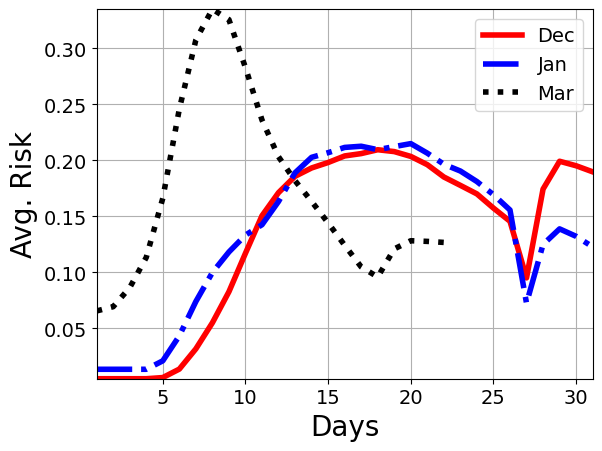} \vspace{-5pt} \\
        & {\footnotesize (d-i)} & \hspace{-5pt} {\footnotesize(d-ii)} & \hspace{-5pt} {\footnotesize(d-iii)}\\
    \end{tabular}}
    \vspace{-10pt}
    \caption{Comparing the Risk across months in different regions of San Francisco--Bay Area, CA. Panels (a), (b), and (c) show the evaluated risk ($\rho$) for each cluster (marked via numbers) for the months of Dec `19, Jan `20, and Mar `20, respectively. The panels (i), (ii), and (iii) show the risk varying over day 10, 15, and 20, respectively. In addition, panel (d-i), (d-ii), and (d-iii) show the comparison between the average number of location signals, infections and risk, across clusters over months, respectively. We observe that while the risk for months of Dec and Jan show similar trends for different days, the month of Mar has lower risk later in the month, which can be attributed to the drop in mobility; see cluster $12$ (best viewed digitally). }
    \label{fig:sf}
\end{figure}

\begin{table}[!b]
    \centering
    \resizebox{0.98\textwidth}{!}{
    \begin{tabular}{c c c c c |c c c c |c c c c}
    \multirow{4}{*}{\textbf{\Large Model}} &  \multicolumn{12}{c}{ \textbf{\Large Infection and Risk Prediction for San Francisco--Bay Area, CA \vspace{2pt}}}\\\cline{2-13}
    & \multicolumn{4}{c}{December `19 } & \multicolumn{4}{|c}{January `20} & \multicolumn{4}{|c}{March `20}\\
    & \multicolumn{4}{c}{$(\alpha, \beta, \Delta) = (2, 2, 3)$ } & \multicolumn{4}{|c}{$(\alpha, \beta, \Delta) = (2, 2, 12)$} & \multicolumn{4}{|c}{$(\alpha, \beta, \Delta) = (2, 2, 12)$}\\
    & \texttt{R-MAE(I)} & $\sigma$\texttt{(I)} &
    \texttt{MAE}($\rho_{\mathrm{test}}$) & \texttt{MAE}($\rho_{\mathrm{all}}$) & \texttt{R-MAE(I)} & $\sigma$\texttt{(I)} &
    \texttt{MAE}($\rho_{\mathrm{test}})$ & \texttt{MAE}($\rho_{\mathrm{all}}$) & \texttt{R-MAE(I)} & $\sigma$\texttt{(I)} &
    \texttt{MAE}($\rho_{\mathrm{test}}$) & \texttt{MAE}($\rho_{\mathrm{all}}$)\\ \hline
    Hawkes$_{Den}$ & 0.272 & 0.138 & \textbf{0.073} & \textbf{0.037} & 0.615 & 0.162 & 0.117 & 0.048 & 1.284 & 0.197 & 0.168 & 0.139 \\
    \riskModelName{}$_{Mob}$ & 0.224 & 0.131 & 0.313& 0.257  &0.340  &0.139 & 0.067 & 0.038 & 0.758 & 0.157 & 0.169 & 0.142 \\
    \riskModelName{}$_{Mob^+}$ & \textbf{0.136} & \textbf{0.115} & 0.091  & 0.046  & \textbf{0.140} & \textbf{0.123} & \textbf{0.045} & \textbf{0.040} & \textbf{0.281} & \textbf{0.126} & \textbf{0.101} & \textbf{0.112} \\\hline
    \end{tabular}}\vspace{2pt}
    \caption{Predicting (5-day) Infections and Risk for San Francisco--Bay Area, CA. The table shows the error in predicted infections (\texttt{I}), the corresponding standard deviation, risk ($\rho$) for the test set, and over all days for Dec `19, Jan `20, and Mar `20 for the top-$5$ clusters.}
    \label{tab:SF_res}
    \vspace{-0.5cm}
\end{table}  

\subsection{Experimental Setup} 
We run \simlName{} described in Sec.~\ref{sec:simulator} for the areas around the cities of San Francisco (and the bay area), Miami (Miami-Dade county), Chicago (Cook county) and Houston (Harris county) over the available days in December 2019, January 2020, and March 2020. The results for more cities across the US can be found in Appendix \ref{sec:exp_additional}. For each city and month, \simlName{} starts on day 1 and ends on the last day of the month, generating a disease spread pattern based on activities of the agents. For each experiment, we use the last $5$ days as our test set to evaluate \riskModelName{}'s infection and risk prediction performance. 

\subsubsection{\simlName{} Parameters} 
For \simlName{}, the parameter setting is shown in Table~\ref{tab:sim_setting}. We set $\mu_{IS}$ to 5 similar to the mean incubation period reported in \cite{li2020early}, and $\sigma_{IS}=1$ since \cite{he2020temporal} observed that most infections occur within 3 days of symptom onset. The work in \cite{he2020temporal} reports ``Infectiousness was estimated to decline quickly within 7 days'', so we set $\mu_{R}$ to $\mu_{IS}+7$. We use a larger value for $\sigma_{R}$ to account for the fact that some people may self-isolate after the onset of the symptoms. We use a value for $d_{max}$ larger than the common 2 meter (m) recommendation for social distancing as the discussion in \cite{bromage2020risks} shows how infections can happen through co-locations with a larger separation between individuals if the co-location lasts for a long enough duration. Thus, we also set $t_{min}$ to 1 hour, larger than the 15 minute minimum threshold mentioned by \cite{wiersinga2020pathophysiology}. Moreover, because we consider co-locations that are at least 1h, we set $p=1$, as the probability of infection becomes higher when co-locations last for long periods.

Furthermore, note that the accuracy of our location data is to about 25m, which is larger than the value we used for $d_{max}$. Thus, the inaccuracy in our location data can affect \textit{who} gets infected in the simulation. However, this does not impact the quality of the infection patterns generated, because the 25m accuracy still ensures that an agent in the \textit{same area} as the infected agent will get infected. For instance, if an infected person is in a shopping mall, the other infected agents will still be in the shopping mall.

Finally, we note that although we have made every effort to design \simlName{} such that it follows the transmission dynamics of COVID-19, existing inaccuracies in the transmission model do not affect the observations made in this paper. This is because, firstly, \simlName{} is used to generate the ground-truth. The models predicting risk-scores are both trained and evaluated against on this ground-truth data. Secondly, \simlName{} is used consistently across all the baselines and our proposed models, and thus our evaluation is fair.

\subsubsection{Baselines}
We analyze the performance of \riskModelName{}$_{Mob}$ and \riskModelName{}$_{Mob^+}$ -- the two variants which utilize the fine-grain mobility-flow information (as described in Sec.~\ref{sec:form_features}) along with Hawkes$_{Den}$, which relies only on the mobility density in the clusters and is the state-of-the-art approach for spatiotemporal modeling of COVID-19 from mobility data \cite{Chiang2020}. Specifically, Hawkes$_{Den}$ uses the location signals at the clusters, with the mobility density feature defined as
\begin{align}
\tag{Hawkes$_{Den}$ Features}
    \boldsymbol{m}_{c}^t  = 
    \begin{bmatrix}
    w_{c\rightarrow c}(t)
    \end{bmatrix}.
\end{align}

\begin{table}[t]
    \centering
    \resizebox{0.98\textwidth}{!}{
    \begin{tabular}{c c c c c| c c c c| c c c c}
    \multirow{4}{*}{\textbf{\Large Model}} &  \multicolumn{12}{c}{\textbf{\Large Infection and Risk Prediction for Miami, FL}\vspace{2pt}}\\\cline{2-13}
    & \multicolumn{4}{c|}{December `19 } & \multicolumn{4}{c|}{January `20} & \multicolumn{4}{c}{March `20}\\
    & \multicolumn{4}{c|}{$(\alpha, \beta, \Delta) = (2, 2, 12)$ } & \multicolumn{4}{c|}{$(\alpha, \beta, \Delta) = (2, 2, 18)$} & \multicolumn{4}{c}{$(\alpha, \beta, \Delta) = (2, 2, 12)$}\\
    & \texttt{R-MAE(I)} & $\sigma$\texttt{(I)} &
    \texttt{MAE}($\rho_{\mathrm{test}}$) & \texttt{MAE}($\rho_{\mathrm{all}}$) & \texttt{R-MAE(I)} & $\sigma$\texttt{(I)} &
    \texttt{MAE}($\rho_{\mathrm{test}})$ & \texttt{MAE}($\rho_{\mathrm{all}}$) & \texttt{R-MAE(I)} & $\sigma$\texttt{(I)} &
    \texttt{MAE}($\rho_{\mathrm{test}}$) & \texttt{MAE}($\rho_{\mathrm{all}}$) \\ \hline
    Hawkes$_{Den}$ &0.383 & 0.189 & 0.121 & 0.058 & 1.216 & 0.285 & 0.102 & 0.075 & 1.365 & 0.227 & 0.244 & 0.154 \\
    \riskModelName{}$_{Mob}$ & 0.162 & 0.164 & 0.087 & 0.043 & 0.196 & 0.178 & 0.064 & 0.076 & 1.229 & 0.219 & 0.235 & 0.147 \\
    \riskModelName{}$_{Mob^+}$&\textbf{0.092} & \textbf{0.146} & \textbf{0.064} & \textbf{0.040} & \textbf{0.152} & \textbf{0.163} & \textbf{0.031} & \textbf{0.054} & \textbf{0.308} & \textbf{0.128} & \textbf{0.117} & \textbf{0.110} \\\hline
    \end{tabular}}\vspace{2pt}
    \caption{Predicting (5-day) Infections and Risk for Miami, FL. The table shows the error in predicted infections (I), the corresponding standard deviation, risk ($\rho$) for the test set, and over all days for Dec `19, Jan `20, and Mar `20 for the top-$5$ clusters.}
    \label{tab:Miami_res}
    \vspace{-0.5cm}
\end{table}  
    


\subsubsection{Metrics}
We evaluate the techniques on two main criteria -- a) infection prediction, to judge the model fit, and b) risk prediction to analyze the efficacy of using the proposed risk metric in Eq. \eqref{eq:risk} as a reliable indicator for predicting potential risk associated with a region over time. For infection prediction we use mean relative absolute error (relative-MAE) between the predicted infection trajectory and the true infections (\texttt{R-MAE(I)}) as the performance metric. This relative measure accounts for the scale differences (in number of infections) across different clusters. In addition, we also report the standard deviation of the predicted infection trajectory ($\sigma$\texttt{(I)}) corresponding to our relative-MAE metric. For risk prediction, we report the mean absolute error (MAE) between the the predicted risk and the infections (scaled between 0 and 1) on the test set (\texttt{MAE}($\rho_{\mathrm{test}}$)) and overall (\texttt{MAE}($\rho_{\mathrm{all}}$)). 
\begin{figure}
    \centering
    \begin{tabular}{cccc}
    & \hspace{0pt} Day $10$ & \hspace{-5pt} Day $15$ & \hspace{-5pt} Day $20$ \\ 
      \rotatebox{90}{\hspace{1.1cm} Dec. `19} \hspace{-8pt}& \hspace{-8pt} \includegraphics[width=0.3\textwidth]{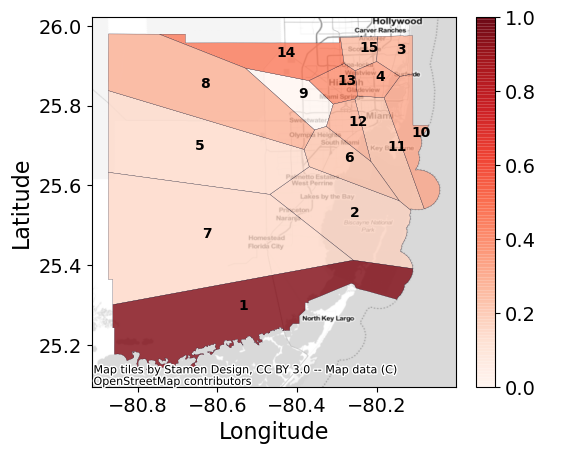}  & \hspace{-15pt}
       \includegraphics[width=0.3\textwidth]{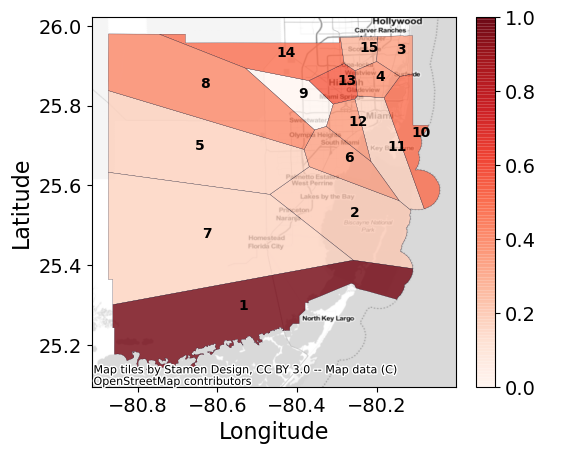} & \hspace{-15pt}
       \includegraphics[width=0.3\textwidth]{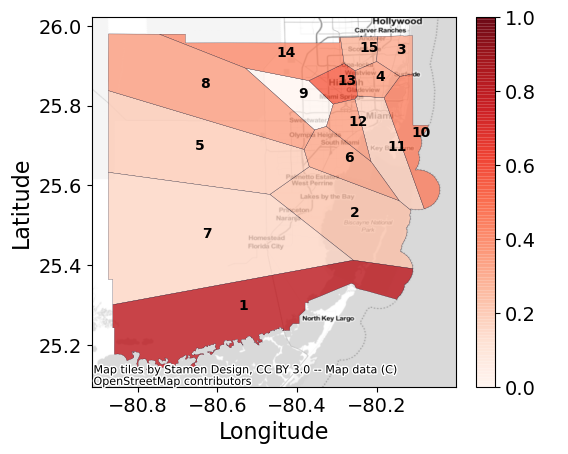} \hspace{-5pt}\\
       & \hspace{0pt} {\footnotesize (a-i)} & \hspace{-5pt} {\footnotesize (a-ii)} & \hspace{-5pt} {\footnotesize (a-iii)}\\
       \rotatebox{90}{\hspace{1.1cm} Jan. `20} \hspace{-8pt}& \hspace{-8pt} \includegraphics[width=0.3\textwidth]{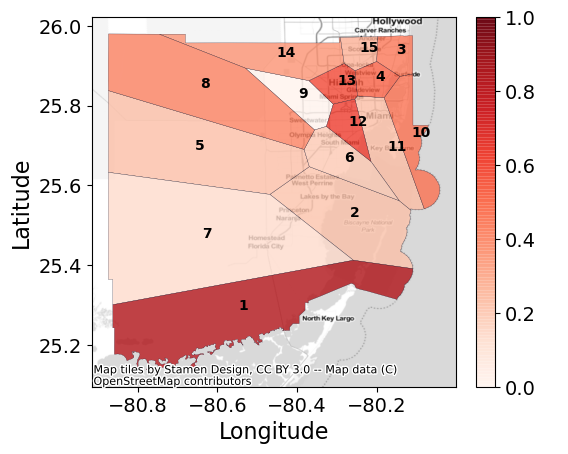} & \hspace{-15pt}
       \includegraphics[width=0.3\textwidth]{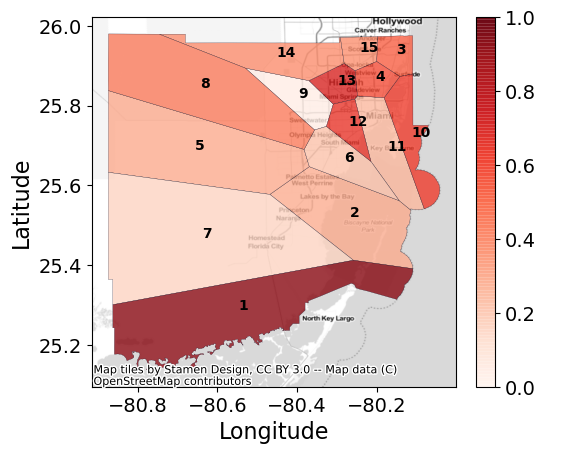} & \hspace{-15pt}
       \includegraphics[width=0.3\textwidth]{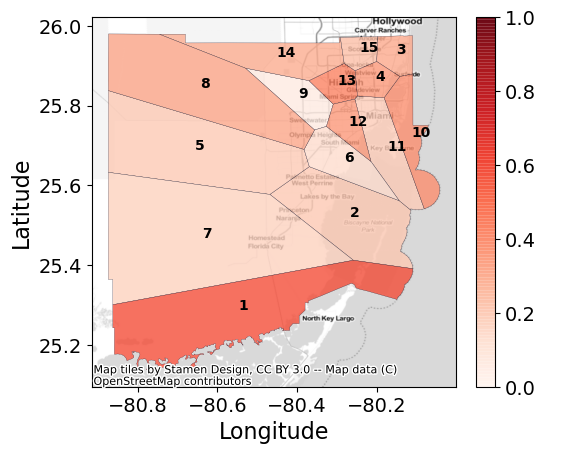}\hspace{-5pt} \\
          & \hspace{0pt} {\footnotesize(b-i)} & \hspace{-5pt} {\footnotesize (b-ii)} & \hspace{-5pt} {\footnotesize (b-iii)}\\
     \rotatebox{90}{ \hspace{1.1cm} Mar. `20} \hspace{-8pt}&\hspace{-8pt} \includegraphics[width=0.3\textwidth]{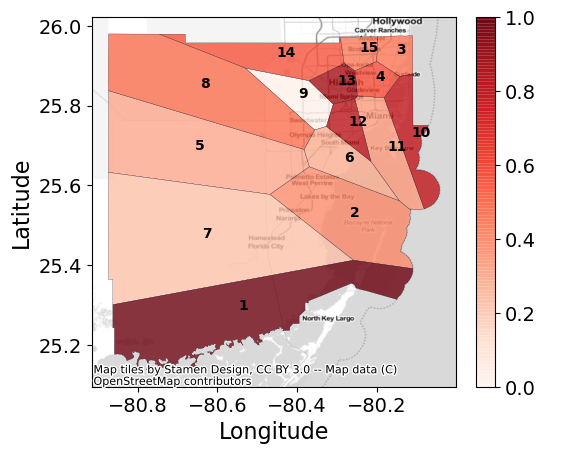} &
       \includegraphics[width=0.3\textwidth]{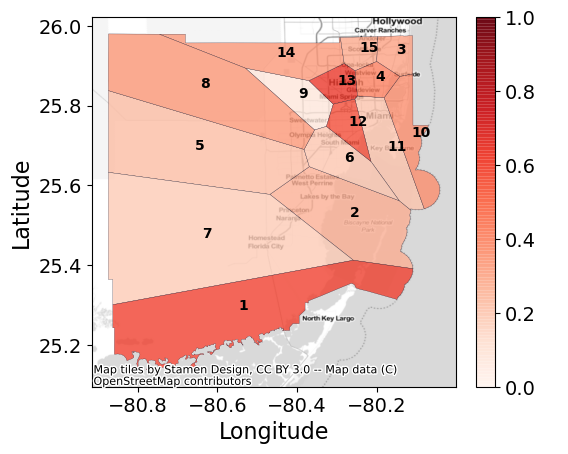} & 
       \includegraphics[width=0.3\textwidth]{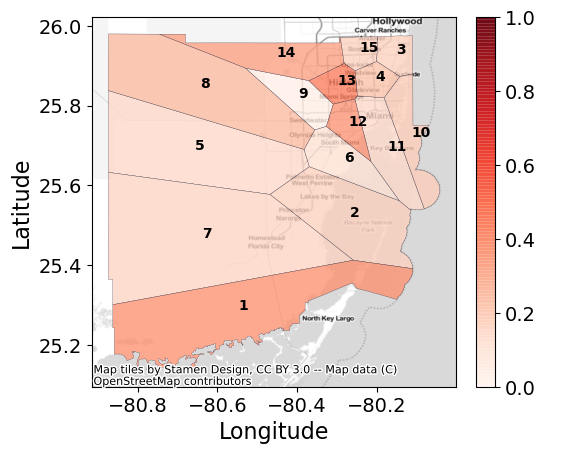} \\
        &  \hspace{0pt} {\footnotesize (c-i)} & \hspace{-5pt} {\footnotesize(c-ii)} & \hspace{-5pt} {\footnotesize(c-iii)}\\
        &\hspace{-17pt} \includegraphics[width=0.29\textwidth]{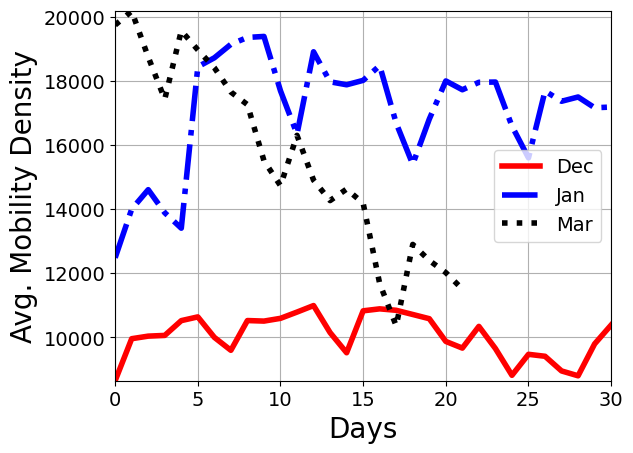}  & \hspace{-17pt}
       \includegraphics[width=0.27\textwidth]{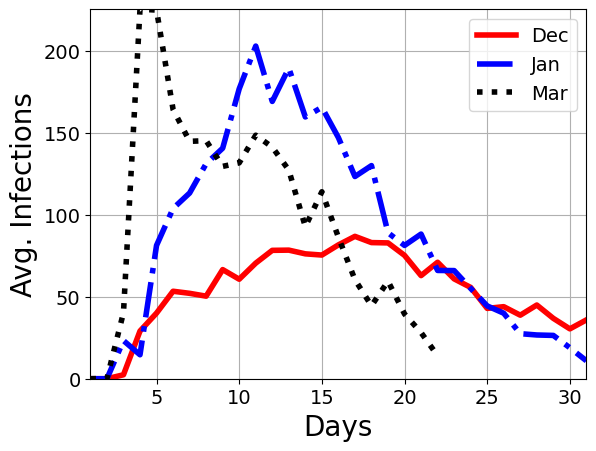} & \hspace{-17pt}
       \includegraphics[width=0.27\textwidth]{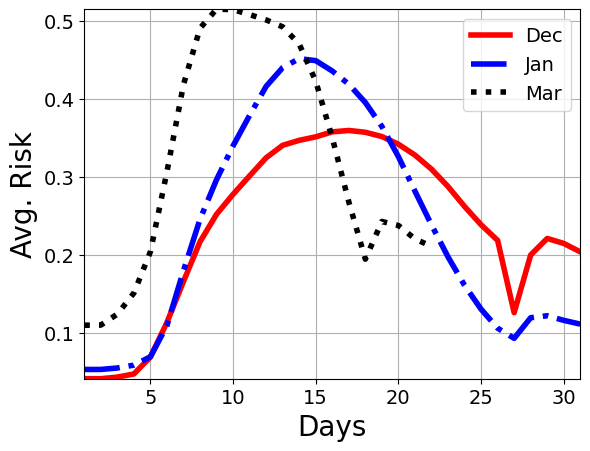} \vspace{-5pt} \\
        & {\footnotesize (d-i)} & \hspace{-5pt} {\footnotesize(d-ii)} & \hspace{-5pt} {\footnotesize(d-iii)}\\
    \end{tabular}
    \vspace{-10pt}
    \caption{Comparing the Risk across months in different regions of Miami, FL. Panels (a), (b), and (c) show the evaluated risk ($\rho$) for each cluster (marked via numbers) for the months of Dec `19, Jan `20, and Mar `20, respectively. The panels (i), (ii), and (iii) show the risk varying over day 10, 15, and 20, respectively. In addition, panel (d-i), (d-ii), and (d-iii) show the comparison between the average number of location signals, infections and risk, across clusters over months, respectively. We observe that while the risk for months of Dec and Jan show similar trends for different days, the month of Mar has lower risk later in the month, which can be attributed to the drop in mobility; see cluster $1$ (best viewed digitally). }
    \label{fig:miami}
\end{figure}
\subsubsection{Pre-processing steps}
We use Origin-Destination (OD) matrix $W(t)$ as a means to gauge the traffic-flow characteristics on day $t$. The OD matrix is a $|\mathcal{C}|\times |\mathcal{C}|$ matrix where $|\mathcal{C}|$ is the number of clusters. In our experiments we set $|\mathcal{C}| = 15$. The entry $w_{i\rightarrow j}(t)$ on the $i$-th row and the $j$-th column of $W(t)$ is calculated as the total count of consecutive location signals, $\mathscr{c}$ and $\mathscr{c}'$, over all individuals, such that $\mathscr{c}$ is in cluster $i$  and $\mathscr{c}'$ is in cluster $j$; see also Sec.~\ref{sec:form_features}.




Next, we use a 6-day moving median filter to smooth the infection and mobility traces. Empirically, we find that such a smoothing helps to counter the dominance of a cluster with sudden rise in cases on the learned model. All mobility features are standardized, meaning that we subtract the mean and divide by the standard deviation. 

\subsubsection{Parameter Choices}
The main parameters of the model are the Weibull parameters ($\alpha$, $\beta$) and the time delay parameter $\Delta$. For each city and month, we choose set of parameters ($\alpha$, $\beta$, $\Delta$) which yields the best relative-MAE on infection prediction. For a fair comparison, all techniques are provided with the same set of parameters for a city and month.

\subsection{Results}
We show the infection prediction performance measured in terms of \texttt{R-MAE} and the risk prediction performance in terms of the \texttt{MAE} between the evaluated risk and scaled infections for top-$5$ clusters (in terms of location signals) of the metropolitan areas of San Francisco, Miami, Chicago and Houston in
Tables~\ref{tab:SF_res}, \ref{tab:Miami_res}, \ref{tab:Cook_res}, and \ref{tab:Harris_res}, respectively. We provide additional results for the metro areas of Los Angeles, New York, Seattle, and Salt lake County in Appendix~\ref{sec:exp_additional}. 
For each of these cities, \riskModelName{}$_{Mob^+}$ yields the best infection prediction performance across all months considered. Furthermore, our models \riskModelName{}$_{Mob}$ and \riskModelName{}$_{Mob^+}$ are also more robust ($\rho_{\mathrm{test}}$ performance) and yield superior performance overall for risk prediction as compared to the density-only baseline of Hawkes$_{Den}$. Note that like any learning model, the prediction capabilities are sensitive to the amount of available infections. As a result, the prediction accuracy for clusters with small number of infections is not high. Arguably, the risk scores for clusters with large number of infections matter more, and for low-infection levels contact tracing may be a more effective tool.

\begin{table}[!t]
    \centering
    \resizebox{0.98\textwidth}{!}{
    \begin{tabular}{c c c c c| c c c c| c c c c}
    \multirow{4}{*}{\textbf{\Large Model}} &  \multicolumn{12}{c}{\textbf{\Large Infection and Risk Prediction for Chicago (Cook County), IL} \vspace{2pt}}\\\cline{2-13}
    & \multicolumn{4}{c|}{December `19 } & \multicolumn{4}{c|}{January `20} & \multicolumn{4}{c}{March `20}\\
    & \multicolumn{4}{c|}{$(\alpha, \beta, \Delta) = (2, 2, 7)$ } & \multicolumn{4}{c|}{$(\alpha, \beta, \Delta) = (2, 2, 9)$} & \multicolumn{4}{c}{$(\alpha, \beta, \Delta) = (2, 2, 9)$}\\
    & \texttt{R-MAE(I)} & $\sigma$\texttt{(I)} &
    \texttt{MAE}($\rho_{\mathrm{test}}$) & \texttt{MAE}($\rho_{\mathrm{all}}$) & \texttt{R-MAE(I)} & $\sigma$\texttt{(I)} &
    \texttt{MAE}($\rho_{\mathrm{test}})$ & \texttt{MAE}($\rho_{\mathrm{all}}$) & \texttt{R-MAE(I)} & $\sigma$\texttt{(I)} &
    \texttt{MAE}($\rho_{\mathrm{test}}$) & \texttt{MAE}($\rho_{\mathrm{all}}$) \\ \hline
    Hawkes$_{Den}$ &0.536 & 0.165 & 0.080 & 0.041 & 0.941 & 0.177 & 0.075 & \textbf{0.048} & 1.005 & 0.142 & 0.215 & 0.143 \\
    \riskModelName{}$_{Mob}$ & 0.337 & 0.151 & 0.071 & 0.038 & 0.238 & 0.129 & 0.037 & 0.053 & 0.342 & 0.098 & 0.173 & 0.144 \\
    \riskModelName{}$_{Mob^+}$& \textbf{0.162} & \textbf{0.126} & \textbf{0.053} & \textbf{0.035} & \textbf{0.174} & \textbf{0.118} & \textbf{0.045} & 0.071 & \textbf{0.209} & \textbf{0.083} & \textbf{0.071} & \textbf{0.078} \\\hline
    \end{tabular}} \vspace{2pt}
    \caption{Predicting (5-day) Infections and Risk for Chicago (Cook County), IL. The table shows the error in predicted infections (\texttt{I}), the corresponding standard deviation, risk ($\rho$) for the test set, and over all days for Dec `19, Jan `20, and Mar `20 for the top-$5$ clusters.} 
    \label{tab:Cook_res}
    \vspace{-0.5cm}
\end{table}  


\begin{figure}
    \centering
    \begin{tabular}{cccc}
    & \hspace{0pt} Day $10$ & \hspace{-5pt} Day $15$ & \hspace{-5pt} Day $20$ \\ 
      \rotatebox{90}{\hspace{1.1cm} Dec. `19} \hspace{-8pt}& \hspace{-8pt} \includegraphics[width=0.3\textwidth]{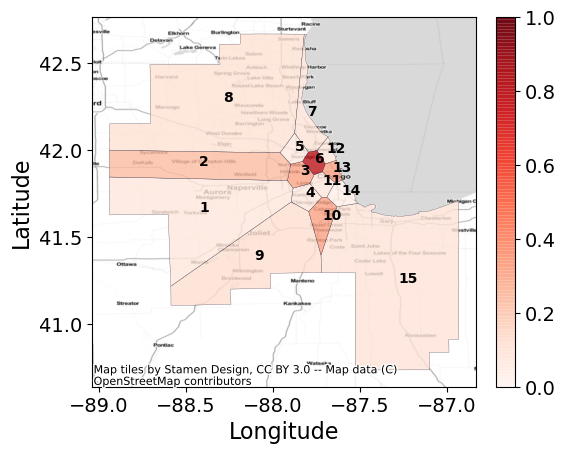}  & \hspace{-8pt}
       \includegraphics[width=0.3\textwidth]{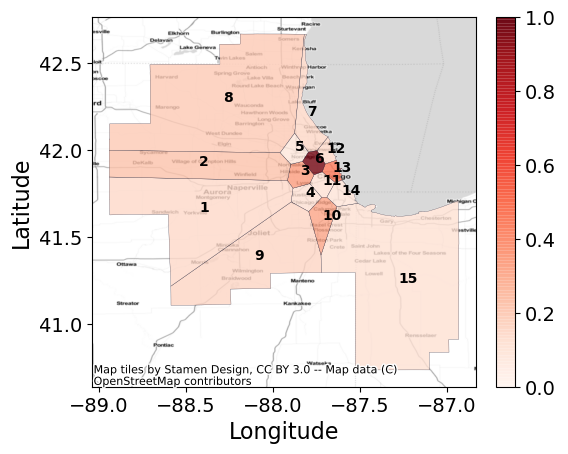} & \hspace{-8pt}
       \includegraphics[width=0.3\textwidth]{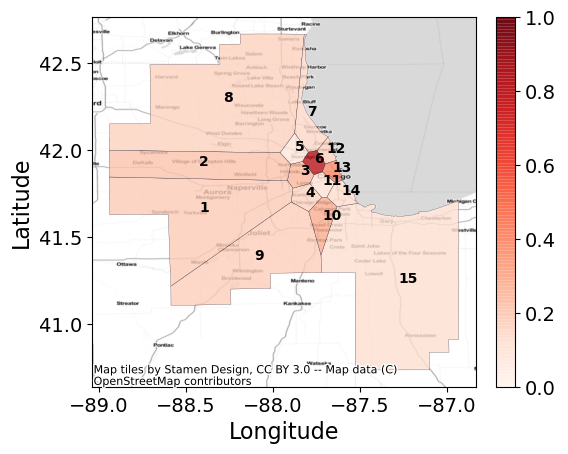} \hspace{-5pt}\\
       & \hspace{0pt} {\footnotesize (a-i)} & \hspace{-5pt} {\footnotesize (a-ii)} & \hspace{-5pt} {\footnotesize (a-iii)}\\
       \rotatebox{90}{\hspace{1.1cm} Jan. `20} \hspace{-8pt}& \hspace{-8pt} \includegraphics[width=0.3\textwidth]{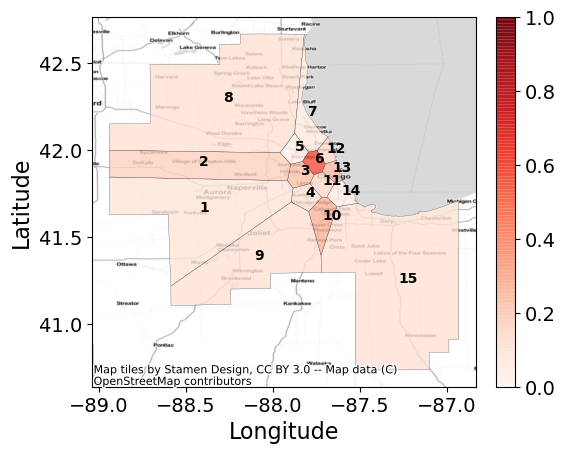} & \hspace{-8pt}
       \includegraphics[width=0.3\textwidth]{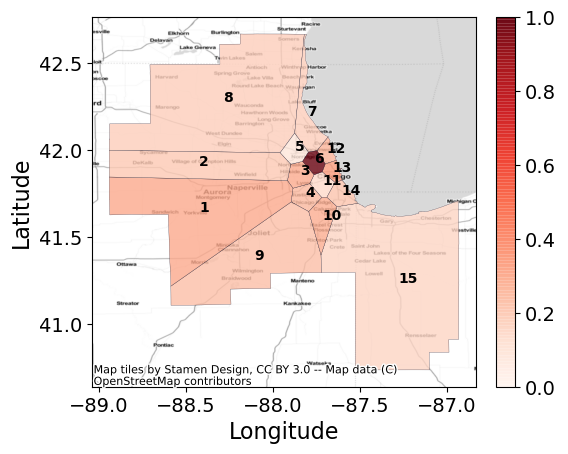} & \hspace{-8pt}
       \includegraphics[width=0.3\textwidth]{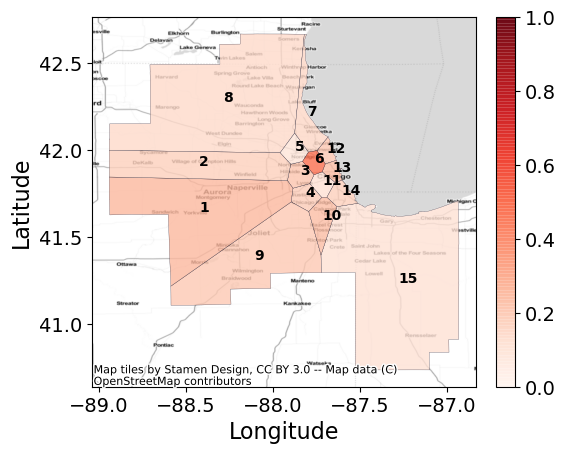}\hspace{-5pt} \\
          & \hspace{0pt} {\footnotesize(b-i)} & \hspace{-5pt} {\footnotesize (b-ii)} & \hspace{-5pt} {\footnotesize (b-iii)}\\
     \rotatebox{90}{ \hspace{1.1cm} Mar. `20} \hspace{-8pt}& \hspace{-8pt} \includegraphics[width=0.3\textwidth]{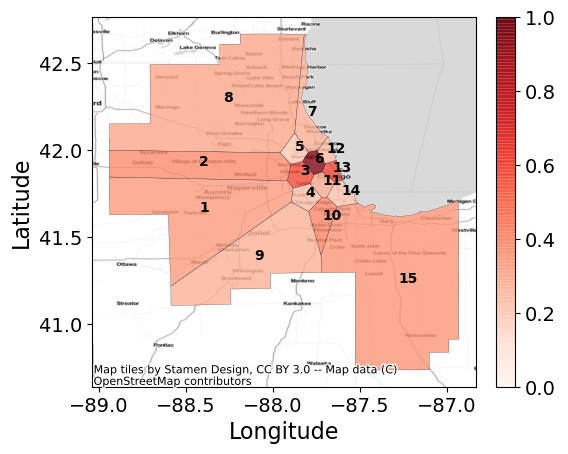} & \hspace{-8pt}
       \includegraphics[width=0.3\textwidth]{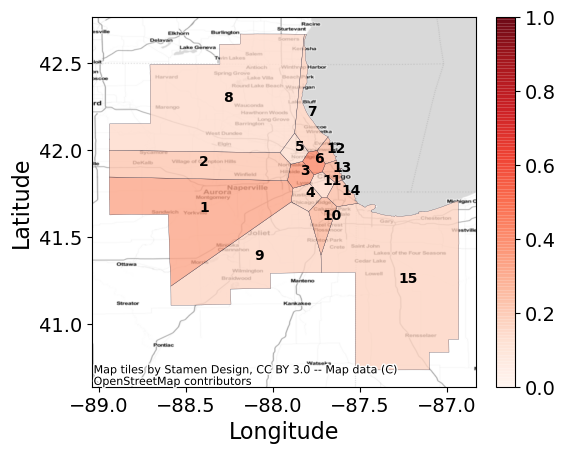} & \hspace{-8pt}
       \includegraphics[width=0.3\textwidth]{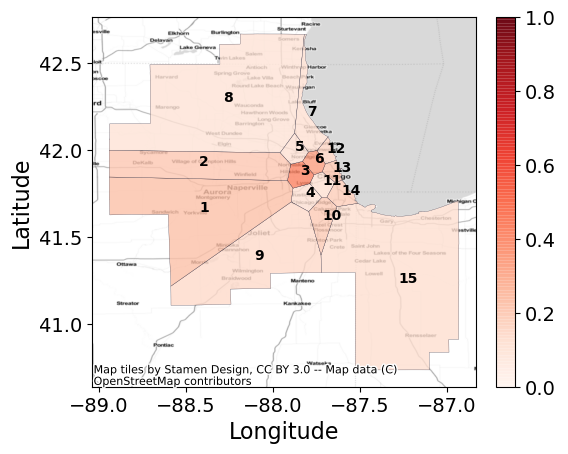} \\
        &  \hspace{0pt} {\footnotesize (c-i)} & \hspace{-5pt} {\footnotesize(c-ii)} & \hspace{-5pt} {\footnotesize(c-iii)}\\
        &\hspace{-17pt} \includegraphics[width=0.29\textwidth]{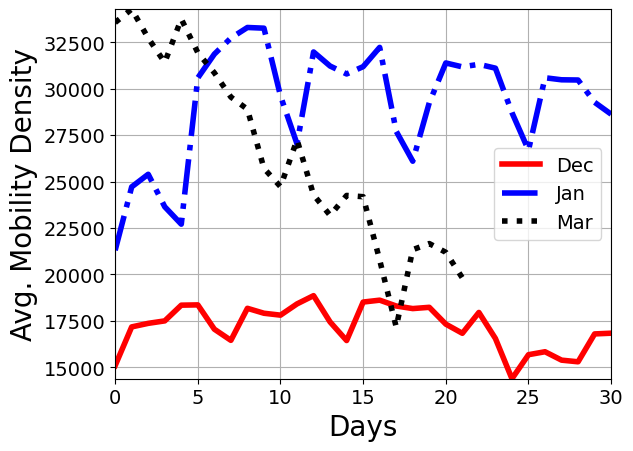}  & \hspace{-17pt}
       \includegraphics[width=0.27\textwidth]{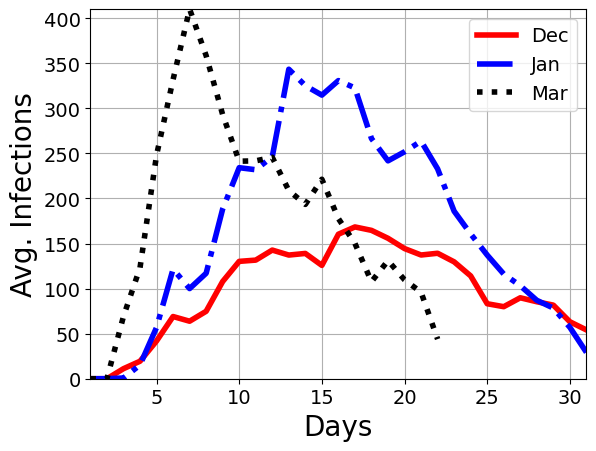} & \hspace{-17pt}
       \includegraphics[width=0.27\textwidth]{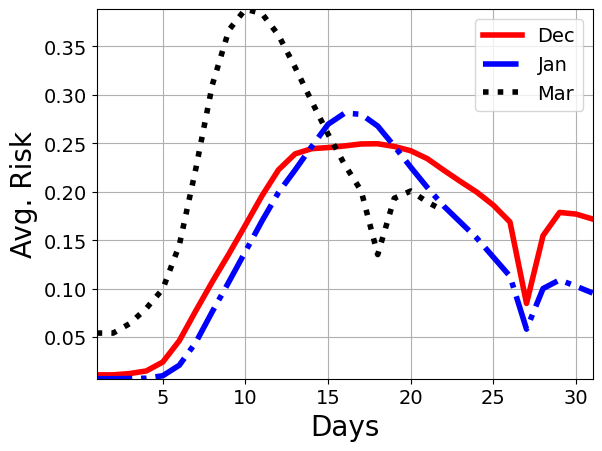} \vspace{-5pt} \\
        & {\footnotesize (d-i)} & \hspace{-5pt} {\footnotesize(d-ii)} & \hspace{-5pt} {\footnotesize(d-iii)}\\

    \end{tabular}
    \vspace{-10pt}
    \caption{Comparing the Risk across months in different regions of Chicago (Cook county), IL. Panels (a), (b), and (c) show the evaluated risk ($\rho$) for each cluster (marked via numbers) for the months of Dec `19, Jan `20, and Mar `20, respectively. The panels (i), (ii), and (iii) show the risk varying over day 10, 15, and 20, respectively. In addition, panel (d-i), (d-ii), and (d-iii) show the comparison between the average number of location signals, infections and risk, across clusters over months, respectively. We observe that while the risk for months of Dec and Jan show similar trends for different days, the month of Mar has lower risk later on, which can be attributed to the drop in mobility; see cluster $6$ (best viewed digitally). }
    \label{fig:cook}
\end{figure}

\begin{figure}
    \centering
    \begin{tabular}{cccc}
    & \hspace{0pt} Day $10$ & \hspace{-5pt} Day $15$ & \hspace{-5pt} Day $20$ \\ 
      \rotatebox{90}{\hspace{1.1cm} Dec. `19} \hspace{-8pt}& \hspace{-8pt} \includegraphics[width=0.3\textwidth]{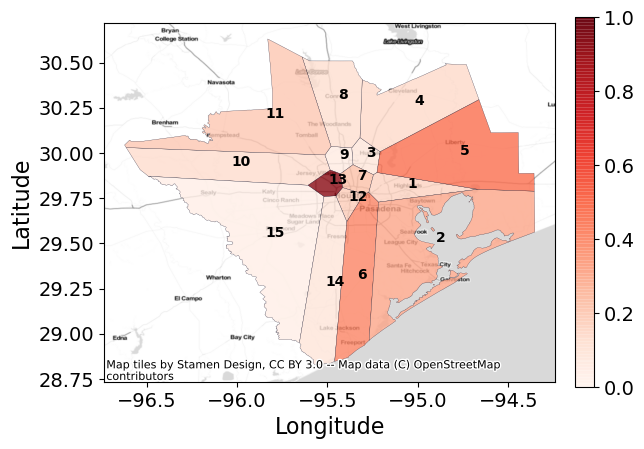}  & \hspace{-8pt}
       \includegraphics[width=0.3\textwidth]{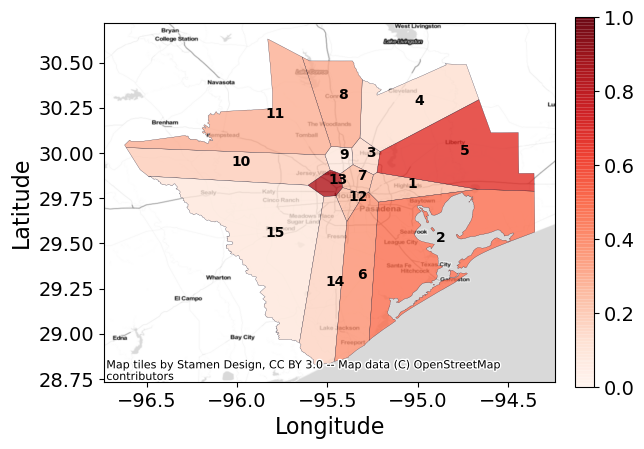} & \hspace{-8pt}
       \includegraphics[width=0.3\textwidth]{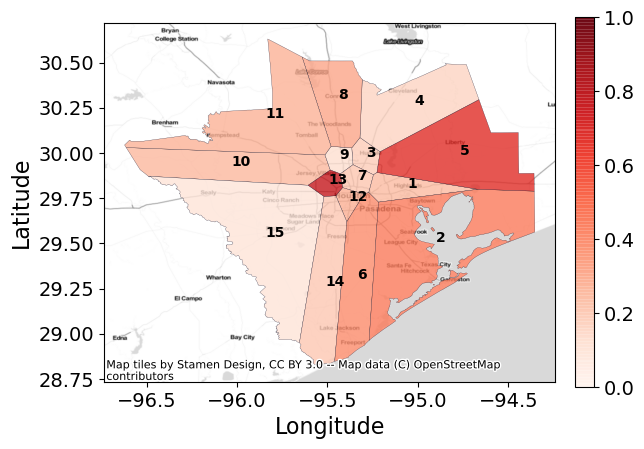} \hspace{-5pt}\\
       & \hspace{0pt} {\footnotesize (a-i)} & \hspace{-5pt} {\footnotesize (a-ii)} & \hspace{-5pt} {\footnotesize (a-iii)}\\
       \rotatebox{90}{\hspace{1.1cm} Jan. `20} \hspace{-8pt}& \hspace{-8pt} \includegraphics[width=0.3\textwidth]{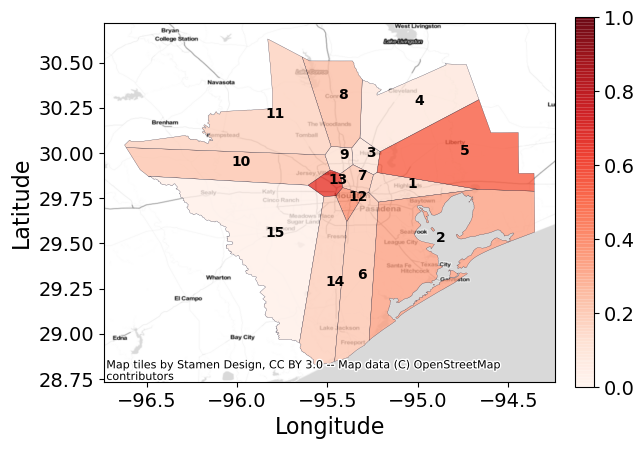} & \hspace{-8pt}
       \includegraphics[width=0.3\textwidth]{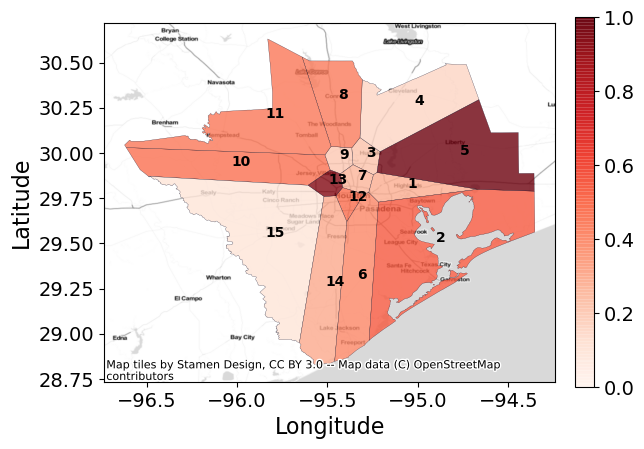} & \hspace{-8pt}
       \includegraphics[width=0.3\textwidth]{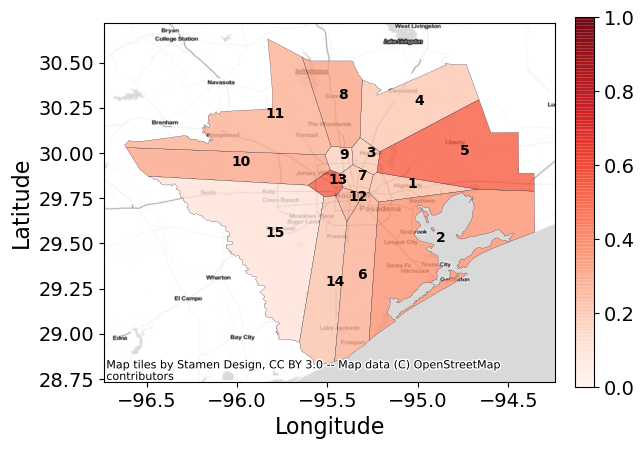}\hspace{-5pt} \\
          & \hspace{0pt} {\footnotesize(b-i)} & \hspace{-5pt} {\footnotesize (b-ii)} & \hspace{-5pt} {\footnotesize (b-iii)}\\
     \rotatebox{90}{ \hspace{1.1cm} Mar. `20} \hspace{-8pt}& \hspace{-8pt} \includegraphics[width=0.3\textwidth]{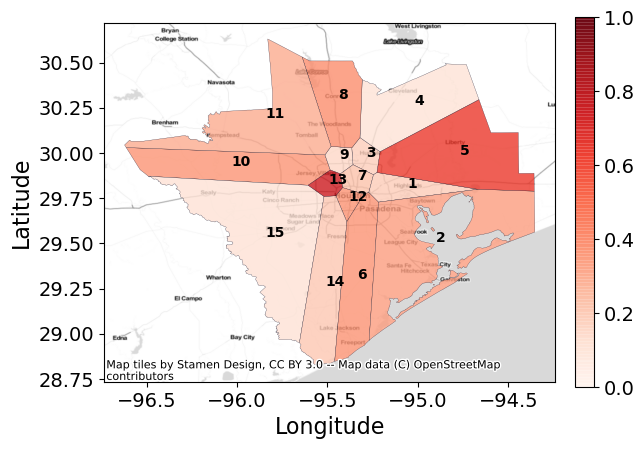} & \hspace{-8pt}
       \includegraphics[width=0.3\textwidth]{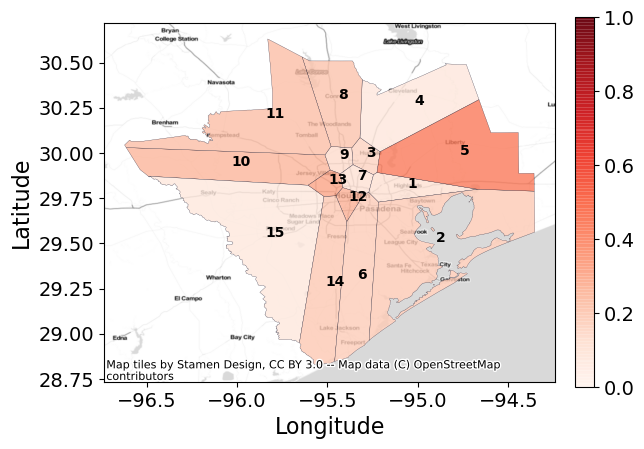} & \hspace{-8pt}
       \includegraphics[width=0.3\textwidth]{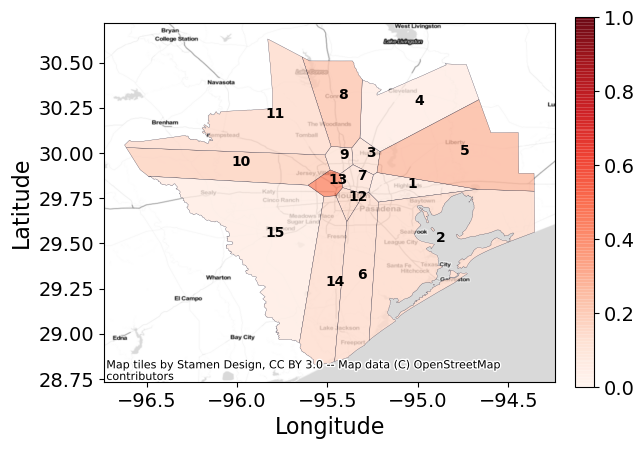} \\
        &  \hspace{0pt} {\footnotesize (c-i)} & \hspace{-5pt} {\footnotesize(c-ii)} & \hspace{-5pt} {\footnotesize(c-iii)}\\
        &\hspace{-17pt} \includegraphics[width=0.29\textwidth]{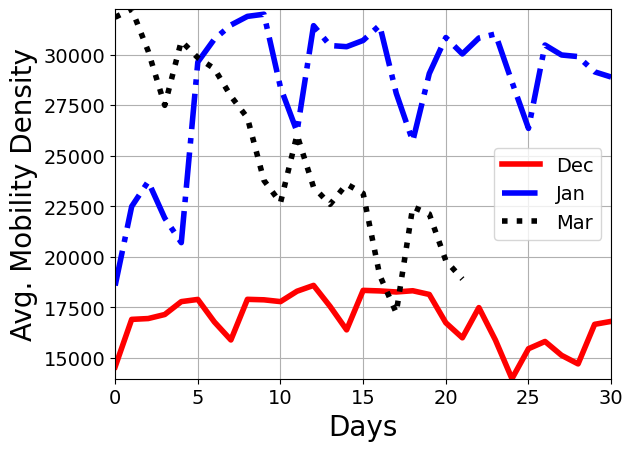}  & \hspace{-17pt}
       \includegraphics[width=0.27\textwidth]{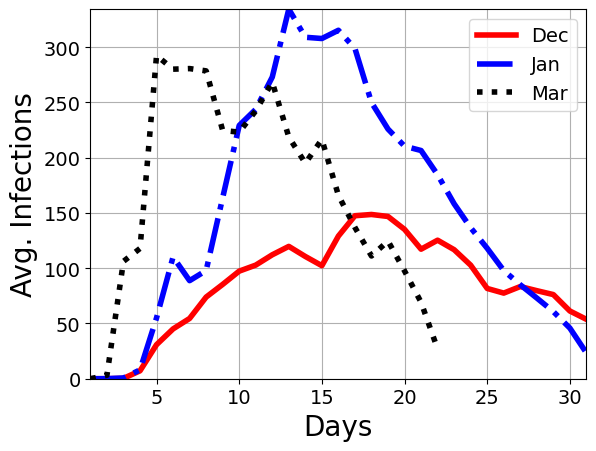} & \hspace{-17pt}
       \includegraphics[width=0.27\textwidth]{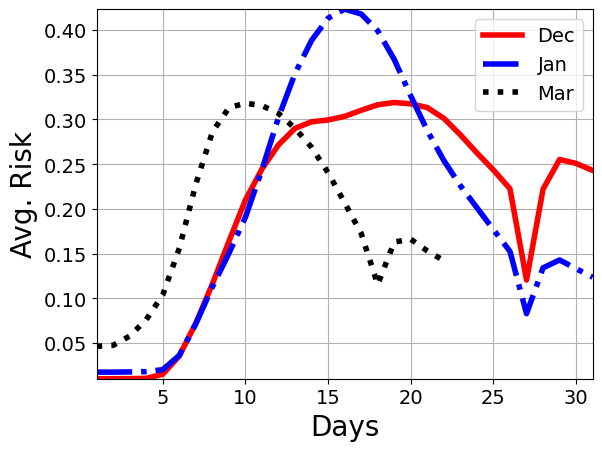} \vspace{-5pt} \\
        & {\footnotesize (d-i)} & \hspace{-5pt} {\footnotesize(d-ii)} & \hspace{-5pt} {\footnotesize(d-iii)}\\

    \end{tabular}
    \vspace{-10pt}
    \caption{Comparing the Risk across months in different regions of Houston (Harris County), TX. Panels (a), (b), and (c) show the evaluated risk ($\rho$) for each cluster (marked via numbers) for the months of Dec `19, Jan `20, and Mar `20, respectively. The panels (i), (ii), and (iii) show the risk varying over day 10, 15, and 20, respectively. In addition, panel (d-i), (d-ii), and (d-iii) show the comparison between the average number of location signals, infections and risk, across clusters over months, respectively. We observe that while the risk for months of Dec and Jan show similar trends for different days, the month of Mar has lower risk later in the month, which can be attributed to the drop in mobility; see cluster $5$ (best viewed digitally). }
    \label{fig:harris}
\end{figure}

In addition, in Figs.~\ref{fig:sf} \ref{fig:miami}, \ref{fig:cook}, and \ref{fig:harris} panels (a-c)(i-iii) we provide visualizations of the predicted spatiotemporal risk by \riskModelName{}$_{Mob^+}$ corresponding to Tables~\ref{tab:SF_res}, \ref{tab:Miami_res}, \ref{tab:Cook_res}, and \ref{tab:Harris_res}, respectively, for day $10$, $15$, and $20$ of the months of Dec. 2019, Jan. 2020, and Mar. 2020. Furthermore, in panel (d) (i-iii) in each of these figures, we compare the corresponding average mobility density, infections and the predicted risk across all clusters (in a city) for these months. Note that here the risk scores are scaled from $[0,1]$ for each month, where a darker color represents a higher risk score relative to the risk in that month. From Figs.~\ref{fig:sf} \ref{fig:miami}, \ref{fig:cook}, and \ref{fig:harris} we note an interesting trend that the similar mobility patterns of the months of Dec. 2019 and Jan. 2020  (except for scale), leads to similar disease spread patterns, and ultimately similar risk scores. The month of Mar. 2020, however, is different for each of these cities from Dec. 2019 and Jan. 2020 across the board. Recall that the stay-at-home order was implemented in the middle of the Mar. 2020 \cite{stayhome2002}, hence our March dataset includes mobility patterns before, during and after the lock-down.  Consequently, we observe higher mobility density during the beginning of the month (which results in higher infections early on). However, as the mobility drops (panel (d-i), there is a corresponding drop in the infections (panel (d-ii), and the risk scores by \riskModelName{}$_{Mob^+}$ (panel (d-iii)). Here, \riskModelName{}$_{Mob^+}$'s risk scores track the infections, emerging as a reliable spatiotemporal risk metric.

Lastly, another important observation is regarding the high-risk areas for each city. Our results using \riskModelName{}$_{Mob^+}$ corroborates our intuition that popular destinations in a city are riskier. For instance, for the metro area of San Francisco, the actual downtown area turns out to be high-risk. This trend can also be observed in other cities as well. \sr{Here, the superior performance of \riskModelName{}$_{Mob^+}$ over Hawkes$_{Den}$ and \riskModelName{}$_{Mob}$ for both infection and risk prediction can be attributed to modeling the infection mobility in addition to the location signal density and mobility. Incorporating the infection mobility as opposed to just relying on popularity of an area, as in case of Hawkes$_{Den}$, allows \riskModelName{}$_{Mob^+}$ to improve infection prediction performance by being cognizant of the past infections in different areas in a city. As a result, \riskModelName{}$_{Mob^+}$ underscores the importance of bringing together mobility patterns and infection spread prediction model to assign high-resolution risk scores.}

\vspace{-5pt}


\section{Related Work}
\label{sec:related}

\subsection{Disease Prediction and Mobility Indicators} Most popular disease prediction models employ compartmental models since they allow explicit modeling of the transmission characteristics. These mainly include variants of the classical Susceptible-Infected-Removed (SIR) model \cite{kermack1927}, such as the S-Exposed-IR (SEIR) model and its variants, which primarily aim to add additional latent states to the SIR model \cite{Miller2020}. Popular in practice due to their simplicity, these models rely on the homogeneous population mixing to learn the model parameters, and have been used to predict reproduction number (R0) at coarser granularity for counties, states and entire countries \cite{Bertozzi16732}. Although useful to communicate the disease characteristics at early stages, coarse-grain risk scores and reproduction number estimates at county or state level are not readily usable for public policy decisions at finer spatial and temporal scales \cite{Chande2020, globalepidemic}. On the other hand, Hawkes process-based point process disease spread models have also emerged as an alternative way to model COVID-19 spread \cite{Chiang2020}. Even though mathematical similarities between compartmental models and Hawkes process-based models exist, Hawkes process-based modeling does not involve complex parameter estimation, model identifiability and mis-specification as in the case of SEIR model \citep{kresincomparison, evans2005structural,roosa2019assessing, hengartner2018quantifying, osthus2017forecasting}. 

Moving away from homogeneous mixing models therefore involves analyzing the specific mobility patterns, and utilizing these covariates in disease spread prediction. To this end, the study in \cite{Chiang2020} leverages the flexibility of the Hawkes process models to incorporate the demographic and mobility indices for COVID-19 prediction. Their work shows that the dynamic reproduction number correlates with the time-delayed mobility density at the county-level across United States, where R0 is viewed as a proxy for the risk associated with a region. Although such coarse-grain risk scores are useful in policy decisions for a country or a state, these may not be informative enough for city-level planning.  To this end, \cite{kiamari2020} use dynamic (time-varying) reproduction number to assign risk scores to communities using a compartmental based model (assume homogeneous mixing), but do not consider mobility features. To this end, \riskModelName{} leverages the Poisson Regression-based reproduction number modelling proposed by \cite{Chiang2020} to incorporate the mobility patterns provided by OD matrices, and infection mobility covariates to develop the spatiotemporal risk scores, as discussed in Sec. \ref{sec:hawkes} and \ref{sec:exps}.

\subsection{Agent-Based Models and Simulations}
Various agent-based simulations are used to model the spread of a disease in a population \cite{kerr2020covasim, chang2020modelling, halloran2008modeling, ferguson2020report, ferguson2006strategies}. They generate synthetic contacts to simulate human contacts using contact matrices, where a pre-defined probability of contact between individuals in different groups of the society is used to decide whether there is contact between individuals at any point in time. Furthermore, their goal is to study different intervention policies. \simlName{} is different from the existing work in two aspects. First, we use real-world location signals to model human mobility. This allows us to create realistic infection patterns in a population that changes over time based on the mobility and is different for different populations. The existing simulations are incapable of capturing this because they generate contacts between individuals synthetically. Second, we use \simlName{} only as a means of generating realistic infection information to allow us to evaluate \riskModelName{}.

\section{Conclusions and Future Work}
\label{sec:discuss}

\sr{In this work, we demonstrated that time-varying location-based risk scores can be a valuable public health tool to facilitate safe reopening of normal activities. The existing risk scores (based on reproduction number learned using compartmental models) either do not provide the information at spatial and time resolutions to be useful, or rely on uniform mixing of population, which is be unrealistic in practice. 

We developed \riskModelName{}, a Hawkes process based model for infection and risk forecasting, where we incorporate actual mobility patterns along with mobility of infected population from different regions of city to assign spatiotemporal risk scores at relatively finer temporal and spatial scales. Subsequently, we demonstrated the applicability of model by, \simlName{}, which simulates the disease spread over actual mobility data from months of Dec. 2019, Jan. 2020, and Mar. 2020 across cities in the United States. Our risk scores emerge as a reliable metric while tracking the infections in a city. One limitation of our approach is that even though we rely on real-world co-locations, the disease spread
mechanism is based on simulation, which is agnostic to any real physical barriers or other factors which may influence disease spread. 

We plan to extend our work in three directions.  First, we intend to develop  individual-level user-specific risk scores by combining user trajectory prediction models with our spatiotemporal risk prediction model. Next, we plan to address the privacy issues associated with assigning user-specific risk scores. Finally, incorporating demographics and electronic medical records (EMR) data, considering spatiotemporal Hawkes process models \cite{yuan2019multivariate} and deep learning based models for long-term forecasting capability, also constitute our future work.}

\begin{table}[t]
    \centering
    \resizebox{0.98\textwidth}{!}{
    \begin{tabular}{c c c c c| c c c c| c c c c}
    \multirow{4}{*}{\textbf{\Large Model}} &  \multicolumn{12}{c}{\textbf{\Large Infection and Risk Prediction for Houston (Harris County), TX} \vspace{2pt}}\\\cline{2-13}
    & \multicolumn{4}{c|}{December `19 } & \multicolumn{4}{c|}{January `20} & \multicolumn{4}{c}{March `20}\\
    & \multicolumn{4}{c|}{$(\alpha, \beta, \Delta) = (2, 2, 5)$ } & \multicolumn{4}{c|}{$(\alpha, \beta, \Delta) = (2, 2, 10)$} & \multicolumn{4}{c}{$(\alpha, \beta, \Delta) = (2, 2, 10)$}\\
    & \texttt{R-MAE(I)} & $\sigma$\texttt{(I)} &
    \texttt{MAE}($\rho_{\mathrm{test}}$) & \texttt{MAE}($\rho_{\mathrm{all}}$) & \texttt{R-MAE(I)} & $\sigma$\texttt{(I)} &
    \texttt{MAE}($\rho_{\mathrm{test}})$ & \texttt{MAE}($\rho_{\mathrm{all}}$) & \texttt{R-MAE(I)} & $\sigma$\texttt{(I)} &
    \texttt{MAE}($\rho_{\mathrm{test}}$) & \texttt{MAE}($\rho_{\mathrm{all}}$)\\ \hline
    Hawkes$_{Den}$ & 0.357 & 0.145 & 0.097 & \textbf{0.055} & 1.022 & 0.206 & 0.085 & 0.074 & 1.285 & 0.166 & 0.200 & 0.115 \\
    \riskModelName{}$_{Mob}$ & 0.208 & 0.133 & 0.112 & 0.063 & 0.276 & 0.135 & \textbf{0.040} & 0.082 & 0.457 & 0.107 & 0.102 & 0.076 \\
    \riskModelName{}$_{Mob^+}$&\textbf{0.159} & \textbf{0.113} & \textbf{0.070} & 0.076 & \textbf{0.204} & \textbf{0.132} & 0.044 & \textbf{0.070} & \textbf{0.256} & \textbf{0.093} & \textbf{0.071} & \textbf{0.066} \\\hline
    \end{tabular}}\vspace{2pt}
    \caption{Predicting (5-day) Infections and Risk for Houston (Harris County), TX. The table shows the error in predicted infections (\texttt{I}), the corresponding standard deviation, risk ($\rho$) for the test set, and over all days for Dec `19, Jan `20, and Mar `20 for the top-$5$ clusters.}
    \label{tab:Harris_res}
    \vspace{-0.5cm}
\end{table}  
\section*{Acknowledgements}
This research has been funded in part by NSF grants IIS-1910950, CNS-2027794, and IIS-1254206, the USC Integrated Media Systems Center (IMSC), and unrestricted cash gifts from Microsoft. We would also like to acknowledge Veraset for providing us with high fidelity location signals.  Any opinions, findings, and conclusions or recommendations expressed in this material are those of the author(s) and do not necessarily reflect the views of the sponsors.
\clearpage
\bibliographystyle{ACM-Reference-Format}
\bibliography{sample-base, high_res_risk}


\begin{thebibliography}{46}


\ifx \showCODEN    \undefined \def \showCODEN     #1{\unskip}     \fi
\ifx \showDOI      \undefined \def \showDOI       #1{#1}\fi
\ifx \showISBNx    \undefined \def \showISBNx     #1{\unskip}     \fi
\ifx \showISBNxiii \undefined \def \showISBNxiii  #1{\unskip}     \fi
\ifx \showISSN     \undefined \def \showISSN      #1{\unskip}     \fi
\ifx \showLCCN     \undefined \def \showLCCN      #1{\unskip}     \fi
\ifx \shownote     \undefined \def \shownote      #1{#1}          \fi
\ifx \showarticletitle \undefined \def \showarticletitle #1{#1}   \fi
\ifx \showURL      \undefined \def \showURL       {\relax}        \fi
\providecommand\bibfield[2]{#2}
\providecommand\bibinfo[2]{#2}
\providecommand\natexlab[1]{#1}
\providecommand\showeprint[2][]{arXiv:#2}

\bibitem[\protect\citeauthoryear{??}{sta}{2020}]%
        {stayhome2002}
 \bibinfo{year}{2020}\natexlab{}.
\newblock \bibinfo{title}{Stay-at-home order}.
\newblock
  \bibinfo{howpublished}{\url{https://covid19.ca.gov/stay-home-except-for-essential-needs/\#:~:text=All\%20individuals\%20living\%20in\%20the,the\%20Questions\%20\%26\%20Answers\%20below)}}.
\newblock
\newblock
\shownote{Accessed: 2020-11-10.}


\bibitem[\protect\citeauthoryear{??}{ver}{2020}]%
        {veraset}
 \bibinfo{year}{2020}\natexlab{}.
\newblock \bibinfo{title}{Veraset Website}.
\newblock \bibinfo{howpublished}{\url{https://www.veraset.com/about-veraset}}.
\newblock
\newblock
\shownote{Accessed: 2020-10-25.}


\bibitem[\protect\citeauthoryear{Abrahamsson}{Abrahamsson}{1998}]%
        {abrahamsson1998estimation}
\bibfield{author}{\bibinfo{person}{Torgil Abrahamsson}.}
  \bibinfo{year}{1998}\natexlab{}.
\newblock \showarticletitle{Estimation of origin-destination matrices using
  traffic counts-a literature survey}.
\newblock  (\bibinfo{year}{1998}).
\newblock


\bibitem[\protect\citeauthoryear{Bertozzi, Franco, Mohler, Short, and
  Sledge}{Bertozzi et~al\mbox{.}}{2020}]%
        {Bertozzi16732}
\bibfield{author}{\bibinfo{person}{Andrea~L. Bertozzi}, \bibinfo{person}{Elisa
  Franco}, \bibinfo{person}{George Mohler}, \bibinfo{person}{Martin~B. Short},
  {and} \bibinfo{person}{Daniel Sledge}.} \bibinfo{year}{2020}\natexlab{}.
\newblock \showarticletitle{The challenges of modeling and forecasting the
  spread of COVID-19}.
\newblock \bibinfo{journal}{\emph{Proceedings of the National Academy of
  Sciences}} \bibinfo{volume}{117}, \bibinfo{number}{29}
  (\bibinfo{year}{2020}), \bibinfo{pages}{16732--16738}.
\newblock
\showISSN{0027-8424}
\urldef\tempurl%
\url{https://doi.org/10.1073/pnas.2006520117}
\showDOI{\tempurl}
\showeprint{https://www.pnas.org/content/117/29/16732.full.pdf}


\bibitem[\protect\citeauthoryear{Bromage}{Bromage}{2020}]%
        {bromage2020risks}
\bibfield{author}{\bibinfo{person}{Erin Bromage}.}
  \bibinfo{year}{2020}\natexlab{}.
\newblock \showarticletitle{The Risks-Know Them-Avoid Them}.
\newblock \bibinfo{journal}{\emph{Erin Bromage: COVID-19 Musings}}
  (\bibinfo{year}{2020}).
\newblock


\bibitem[\protect\citeauthoryear{Chande, Lee, Harris, Nguyen, Beckett, Hilley,
  Andris, and Weitz}{Chande et~al\mbox{.}}{2020}]%
        {Chande2020}
\bibfield{author}{\bibinfo{person}{Aroon Chande}, \bibinfo{person}{Seolha Lee},
  \bibinfo{person}{Mallory Harris}, \bibinfo{person}{Quan Nguyen},
  \bibinfo{person}{Stephen~J. Beckett}, \bibinfo{person}{Troy Hilley},
  \bibinfo{person}{Clio Andris}, {and} \bibinfo{person}{Joshua~S. Weitz}.}
  \bibinfo{year}{2020}\natexlab{}.
\newblock \showarticletitle{Real-time, interactive website for US-county-level
  COVID-19 event risk assessment}.
\newblock \bibinfo{journal}{\emph{Nature Human Behaviour}}
  (\bibinfo{year}{2020}).
\newblock
\urldef\tempurl%
\url{https://doi.org/10.1038/s41562-020-01000-9}
\showDOI{\tempurl}


\bibitem[\protect\citeauthoryear{Chang, Pierson, Koh, Gerardin, Redbird,
  Grusky, and Leskovec}{Chang et~al\mbox{.}}{2020b}]%
        {chang2020mobility}
\bibfield{author}{\bibinfo{person}{Serina Chang}, \bibinfo{person}{Emma
  Pierson}, \bibinfo{person}{Pang~Wei Koh}, \bibinfo{person}{Jaline Gerardin},
  \bibinfo{person}{Beth Redbird}, \bibinfo{person}{David Grusky}, {and}
  \bibinfo{person}{Jure Leskovec}.} \bibinfo{year}{2020}\natexlab{b}.
\newblock \showarticletitle{Mobility network models of COVID-19 explain
  inequities and inform reopening}.
\newblock \bibinfo{journal}{\emph{Nature}} (\bibinfo{year}{2020}),
  \bibinfo{pages}{1--6}.
\newblock


\bibitem[\protect\citeauthoryear{Chang, Harding, Zachreson, Cliff, and
  Prokopenko}{Chang et~al\mbox{.}}{2020a}]%
        {chang2020modelling}
\bibfield{author}{\bibinfo{person}{Sheryl~L Chang}, \bibinfo{person}{Nathan
  Harding}, \bibinfo{person}{Cameron Zachreson}, \bibinfo{person}{Oliver~M
  Cliff}, {and} \bibinfo{person}{Mikhail Prokopenko}.}
  \bibinfo{year}{2020}\natexlab{a}.
\newblock \showarticletitle{Modelling transmission and control of the COVID-19
  pandemic in Australia}.
\newblock \bibinfo{journal}{\emph{arXiv preprint arXiv:2003.10218}}
  (\bibinfo{year}{2020}).
\newblock


\bibitem[\protect\citeauthoryear{Chiang, Liu, and Mohler}{Chiang
  et~al\mbox{.}}{2020}]%
        {Chiang2020}
\bibfield{author}{\bibinfo{person}{Wen-Hao Chiang}, \bibinfo{person}{Xueying
  Liu}, {and} \bibinfo{person}{George Mohler}.}
  \bibinfo{year}{2020}\natexlab{}.
\newblock \showarticletitle{Hawkes process modeling of COVID-19 with mobility
  leading indicators and spatial covariates}.
\newblock \bibinfo{journal}{\emph{medRxiv}} (\bibinfo{year}{2020}).
\newblock
\urldef\tempurl%
\url{https://doi.org/10.1101/2020.06.06.20124149}
\showDOI{\tempurl}
\showeprint{https://www.medrxiv.org/content/early/2020/06/08/2020.06.06.20124149.full.pdf}


\bibitem[\protect\citeauthoryear{Chu, Akl, Duda, Solo, Yaacoub, Sch{\"u}nemann,
  El-harakeh, Bognanni, Lotfi, Loeb, et~al\mbox{.}}{Chu et~al\mbox{.}}{2020}]%
        {chu2020physical}
\bibfield{author}{\bibinfo{person}{Derek~K Chu}, \bibinfo{person}{Elie~A Akl},
  \bibinfo{person}{Stephanie Duda}, \bibinfo{person}{Karla Solo},
  \bibinfo{person}{Sally Yaacoub}, \bibinfo{person}{Holger~J Sch{\"u}nemann},
  \bibinfo{person}{Amena El-harakeh}, \bibinfo{person}{Antonio Bognanni},
  \bibinfo{person}{Tamara Lotfi}, \bibinfo{person}{Mark Loeb}, {et~al\mbox{.}}}
  \bibinfo{year}{2020}\natexlab{}.
\newblock \showarticletitle{Physical distancing, face masks, and eye protection
  to prevent person-to-person transmission of SARS-CoV-2 and COVID-19: a
  systematic review and meta-analysis}.
\newblock \bibinfo{journal}{\emph{The Lancet}} (\bibinfo{year}{2020}).
\newblock


\bibitem[\protect\citeauthoryear{Delamater, Street, Leslie, Yang, and
  Jacobsen}{Delamater et~al\mbox{.}}{2019}]%
        {delamater2019}
\bibfield{author}{\bibinfo{person}{Paul~L Delamater}, \bibinfo{person}{Erica~J
  Street}, \bibinfo{person}{Timothy~F Leslie}, \bibinfo{person}{Y~Tony Yang},
  {and} \bibinfo{person}{Kathryn~H Jacobsen}.} \bibinfo{year}{2019}\natexlab{}.
\newblock \showarticletitle{Complexity of the basic reproduction number (R0)}.
\newblock \bibinfo{journal}{\emph{Emerging infectious diseases}}
  \bibinfo{volume}{25}, \bibinfo{number}{1} (\bibinfo{year}{2019}),
  \bibinfo{pages}{1}.
\newblock


\bibitem[\protect\citeauthoryear{Eichler, Dahlhaus, and Dueck}{Eichler
  et~al\mbox{.}}{2017}]%
        {eichler2017graphical}
\bibfield{author}{\bibinfo{person}{Michael Eichler}, \bibinfo{person}{Rainer
  Dahlhaus}, {and} \bibinfo{person}{Johannes Dueck}.}
  \bibinfo{year}{2017}\natexlab{}.
\newblock \showarticletitle{Graphical modeling for multivariate hawkes
  processes with nonparametric link functions}.
\newblock \bibinfo{journal}{\emph{Journal of Time Series Analysis}}
  \bibinfo{volume}{38}, \bibinfo{number}{2} (\bibinfo{year}{2017}),
  \bibinfo{pages}{225--242}.
\newblock


\bibitem[\protect\citeauthoryear{Evans, White, Chapman, Godfrey, and
  Chappell}{Evans et~al\mbox{.}}{2005}]%
        {evans2005structural}
\bibfield{author}{\bibinfo{person}{Neil~D Evans}, \bibinfo{person}{Lisa~J
  White}, \bibinfo{person}{Michael~J Chapman}, \bibinfo{person}{Keith~R
  Godfrey}, {and} \bibinfo{person}{Michael~J Chappell}.}
  \bibinfo{year}{2005}\natexlab{}.
\newblock \showarticletitle{The structural identifiability of the susceptible
  infected recovered model with seasonal forcing}.
\newblock \bibinfo{journal}{\emph{Mathematical biosciences}}
  \bibinfo{volume}{194}, \bibinfo{number}{2} (\bibinfo{year}{2005}),
  \bibinfo{pages}{175--197}.
\newblock


\bibitem[\protect\citeauthoryear{Farrington, Kanaan, and Gay}{Farrington
  et~al\mbox{.}}{2003}]%
        {farrington2003branching}
\bibfield{author}{\bibinfo{person}{CP Farrington}, \bibinfo{person}{MN Kanaan},
  {and} \bibinfo{person}{NJ Gay}.} \bibinfo{year}{2003}\natexlab{}.
\newblock \showarticletitle{Branching process models for surveillance of
  infectious diseases controlled by mass vaccination}.
\newblock \bibinfo{journal}{\emph{Biostatistics}} \bibinfo{volume}{4},
  \bibinfo{number}{2} (\bibinfo{year}{2003}), \bibinfo{pages}{279--295}.
\newblock


\bibitem[\protect\citeauthoryear{Ferguson, Laydon, Nedjati~Gilani, Imai,
  Ainslie, Baguelin, Bhatia, Boonyasiri, Cucunuba~Perez, Cuomo-Dannenburg,
  et~al\mbox{.}}{Ferguson et~al\mbox{.}}{2020}]%
        {ferguson2020report}
\bibfield{author}{\bibinfo{person}{Neil Ferguson}, \bibinfo{person}{Daniel
  Laydon}, \bibinfo{person}{Gemma Nedjati~Gilani}, \bibinfo{person}{Natsuko
  Imai}, \bibinfo{person}{Kylie Ainslie}, \bibinfo{person}{Marc Baguelin},
  \bibinfo{person}{Sangeeta Bhatia}, \bibinfo{person}{Adhiratha Boonyasiri},
  \bibinfo{person}{ZULMA Cucunuba~Perez}, \bibinfo{person}{Gina
  Cuomo-Dannenburg}, {et~al\mbox{.}}} \bibinfo{year}{2020}\natexlab{}.
\newblock \showarticletitle{Report 9: Impact of non-pharmaceutical
  interventions (NPIs) to reduce COVID19 mortality and healthcare demand}.
\newblock  (\bibinfo{year}{2020}).
\newblock


\bibitem[\protect\citeauthoryear{Ferguson, Cummings, Fraser, Cajka, Cooley, and
  Burke}{Ferguson et~al\mbox{.}}{2006}]%
        {ferguson2006strategies}
\bibfield{author}{\bibinfo{person}{Neil~M Ferguson}, \bibinfo{person}{Derek~AT
  Cummings}, \bibinfo{person}{Christophe Fraser}, \bibinfo{person}{James~C
  Cajka}, \bibinfo{person}{Philip~C Cooley}, {and} \bibinfo{person}{Donald~S
  Burke}.} \bibinfo{year}{2006}\natexlab{}.
\newblock \showarticletitle{Strategies for mitigating an influenza pandemic}.
\newblock \bibinfo{journal}{\emph{Nature}} \bibinfo{volume}{442},
  \bibinfo{number}{7101} (\bibinfo{year}{2006}), \bibinfo{pages}{448--452}.
\newblock


\bibitem[\protect\citeauthoryear{Ferretti, Wymant, Kendall, Zhao, Nurtay,
  Abeler-D{\"o}rner, Parker, Bonsall, and Fraser}{Ferretti
  et~al\mbox{.}}{2020}]%
        {ferretti2020quantifying}
\bibfield{author}{\bibinfo{person}{Luca Ferretti}, \bibinfo{person}{Chris
  Wymant}, \bibinfo{person}{Michelle Kendall}, \bibinfo{person}{Lele Zhao},
  \bibinfo{person}{Anel Nurtay}, \bibinfo{person}{Lucie Abeler-D{\"o}rner},
  \bibinfo{person}{Michael Parker}, \bibinfo{person}{David Bonsall}, {and}
  \bibinfo{person}{Christophe Fraser}.} \bibinfo{year}{2020}\natexlab{}.
\newblock \showarticletitle{Quantifying SARS-CoV-2 transmission suggests
  epidemic control with digital contact tracing}.
\newblock \bibinfo{journal}{\emph{Science}} \bibinfo{volume}{368},
  \bibinfo{number}{6491} (\bibinfo{year}{2020}).
\newblock


\bibitem[\protect\citeauthoryear{for Disease Control~(CDC)}{for Disease
  Control~(CDC)}{2020}]%
        {cdccolocation}
\bibfield{author}{\bibinfo{person}{Center for Disease Control~(CDC)}.}
  \bibinfo{year}{2020}\natexlab{}.
\newblock \bibinfo{title}{Coronavirus Disease 2019 (COVID-19): Daily Activities
  and Going Out}.
\newblock
  \bibinfo{howpublished}{\url{https://www.cdc.gov/coronavirus/2019-ncov/daily-life-coping/going-out.html}}.
\newblock
\newblock
\shownote{Accessed: 2020-11-12.}


\bibitem[\protect\citeauthoryear{Granger}{Granger}{1969}]%
        {Granger1969}
\bibfield{author}{\bibinfo{person}{Clive~WJ Granger}.}
  \bibinfo{year}{1969}\natexlab{}.
\newblock \showarticletitle{Investigating causal relations by econometric
  models and cross-spectral methods}.
\newblock \bibinfo{journal}{\emph{Econometrica: journal of the Econometric
  Society}} (\bibinfo{year}{1969}), \bibinfo{pages}{424--438}.
\newblock


\bibitem[\protect\citeauthoryear{Halloran, Ferguson, Eubank, Longini, Cummings,
  Lewis, Xu, Fraser, Vullikanti, Germann, et~al\mbox{.}}{Halloran
  et~al\mbox{.}}{2008}]%
        {halloran2008modeling}
\bibfield{author}{\bibinfo{person}{M~Elizabeth Halloran},
  \bibinfo{person}{Neil~M Ferguson}, \bibinfo{person}{Stephen Eubank},
  \bibinfo{person}{Ira~M Longini}, \bibinfo{person}{Derek~AT Cummings},
  \bibinfo{person}{Bryan Lewis}, \bibinfo{person}{Shufu Xu},
  \bibinfo{person}{Christophe Fraser}, \bibinfo{person}{Anil Vullikanti},
  \bibinfo{person}{Timothy~C Germann}, {et~al\mbox{.}}}
  \bibinfo{year}{2008}\natexlab{}.
\newblock \showarticletitle{Modeling targeted layered containment of an
  influenza pandemic in the United States}.
\newblock \bibinfo{journal}{\emph{Proceedings of the National Academy of
  Sciences}} \bibinfo{volume}{105}, \bibinfo{number}{12}
  (\bibinfo{year}{2008}), \bibinfo{pages}{4639--4644}.
\newblock


\bibitem[\protect\citeauthoryear{He, Lau, Wu, Deng, Wang, Hao, Lau, Wong, Guan,
  Tan, et~al\mbox{.}}{He et~al\mbox{.}}{2020}]%
        {he2020temporal}
\bibfield{author}{\bibinfo{person}{Xi He}, \bibinfo{person}{Eric~HY Lau},
  \bibinfo{person}{Peng Wu}, \bibinfo{person}{Xilong Deng},
  \bibinfo{person}{Jian Wang}, \bibinfo{person}{Xinxin Hao},
  \bibinfo{person}{Yiu~Chung Lau}, \bibinfo{person}{Jessica~Y Wong},
  \bibinfo{person}{Yujuan Guan}, \bibinfo{person}{Xinghua Tan},
  {et~al\mbox{.}}} \bibinfo{year}{2020}\natexlab{}.
\newblock \showarticletitle{Temporal dynamics in viral shedding and
  transmissibility of COVID-19}.
\newblock \bibinfo{journal}{\emph{Nature medicine}} \bibinfo{volume}{26},
  \bibinfo{number}{5} (\bibinfo{year}{2020}), \bibinfo{pages}{672--675}.
\newblock


\bibitem[\protect\citeauthoryear{Hengartner and Fenimore}{Hengartner and
  Fenimore}{2018}]%
        {hengartner2018quantifying}
\bibfield{author}{\bibinfo{person}{Nicolas Hengartner} {and}
  \bibinfo{person}{Paul Fenimore}.} \bibinfo{year}{2018}\natexlab{}.
\newblock \showarticletitle{Quantifying Model Form Uncertainty of Epidemic
  Forecasting Models from Incidence Data}.
\newblock \bibinfo{journal}{\emph{Online Journal of Public Health Informatics}}
  \bibinfo{volume}{10}, \bibinfo{number}{1} (\bibinfo{year}{2018}).
\newblock


\bibitem[\protect\citeauthoryear{Institute}{Institute}{2020}]%
        {globalepidemic}
\bibfield{author}{\bibinfo{person}{Harvard Global~Health Institute}.}
  \bibinfo{year}{2020}\natexlab{}.
\newblock \bibinfo{title}{Key Metrics for COVID Suppression}.
\newblock
  \bibinfo{howpublished}{\url{https://globalepidemics.org/key-metrics-for-covid-suppression/}}.
\newblock
\newblock
\shownote{Accessed: 2020-11-12.}


\bibitem[\protect\citeauthoryear{Kamra, Zhang, Rambhatla, Meng, and Liu}{Kamra
  et~al\mbox{.}}{2020}]%
        {kamra2020polsird}
\bibfield{author}{\bibinfo{person}{Nitin Kamra}, \bibinfo{person}{Yizhou
  Zhang}, \bibinfo{person}{Sirisha Rambhatla}, \bibinfo{person}{Chuizheng
  Meng}, {and} \bibinfo{person}{Yan Liu}.} \bibinfo{year}{2020}\natexlab{}.
\newblock \bibinfo{title}{PolSIRD: Modeling Epidemic Spread under Intervention
  Policies and an Application to the Spread of COVID-19}.
\newblock
\newblock
\showeprint[arxiv]{q-bio.PE/2009.01894}


\bibitem[\protect\citeauthoryear{Keeling, Hollingsworth, and Read}{Keeling
  et~al\mbox{.}}{2020}]%
        {Keeling861}
\bibfield{author}{\bibinfo{person}{Matt~J Keeling}, \bibinfo{person}{T~Deirdre
  Hollingsworth}, {and} \bibinfo{person}{Jonathan~M Read}.}
  \bibinfo{year}{2020}\natexlab{}.
\newblock \showarticletitle{Efficacy of contact tracing for the containment of
  the 2019 novel coronavirus (COVID-19)}.
\newblock \bibinfo{journal}{\emph{Journal of Epidemiology \& Community Health}}
  \bibinfo{volume}{74}, \bibinfo{number}{10} (\bibinfo{year}{2020}),
  \bibinfo{pages}{861--866}.
\newblock
\showISSN{0143-005X}
\urldef\tempurl%
\url{https://doi.org/10.1136/jech-2020-214051}
\showDOI{\tempurl}
\showeprint{https://jech.bmj.com/content/74/10/861.full.pdf}


\bibitem[\protect\citeauthoryear{Kermack and McKendrick}{Kermack and
  McKendrick}{1927}]%
        {kermack1927}
\bibfield{author}{\bibinfo{person}{William~Ogilvy Kermack} {and}
  \bibinfo{person}{Anderson~G McKendrick}.} \bibinfo{year}{1927}\natexlab{}.
\newblock \showarticletitle{A contribution to the mathematical theory of
  epidemics}.
\newblock \bibinfo{journal}{\emph{Proceedings of the royal society of london.
  Series A, Containing papers of a mathematical and physical character}}
  \bibinfo{volume}{115}, \bibinfo{number}{772} (\bibinfo{year}{1927}),
  \bibinfo{pages}{700--721}.
\newblock


\bibitem[\protect\citeauthoryear{Kerr, Stuart, Mistry, Abeysuriya, Hart,
  Rosenfeld, Selvaraj, Nunez, Hagedorn, George, et~al\mbox{.}}{Kerr
  et~al\mbox{.}}{2020}]%
        {kerr2020covasim}
\bibfield{author}{\bibinfo{person}{Cliff~C Kerr}, \bibinfo{person}{Robyn~M
  Stuart}, \bibinfo{person}{Dina Mistry}, \bibinfo{person}{Romesh~G
  Abeysuriya}, \bibinfo{person}{Gregory Hart}, \bibinfo{person}{Katherine
  Rosenfeld}, \bibinfo{person}{Prashanth Selvaraj}, \bibinfo{person}{Rafael~C
  Nunez}, \bibinfo{person}{Brittany Hagedorn}, \bibinfo{person}{Lauren George},
  {et~al\mbox{.}}} \bibinfo{year}{2020}\natexlab{}.
\newblock \showarticletitle{Covasim: an agent-based model of COVID-19 dynamics
  and interventions}.
\newblock \bibinfo{journal}{\emph{medRxiv}} (\bibinfo{year}{2020}).
\newblock


\bibitem[\protect\citeauthoryear{Kiamari, Ramachandran, Nguyen, Pereira, Holm,
  and Krishnamachari}{Kiamari et~al\mbox{.}}{2020}]%
        {kiamari2020}
\bibfield{author}{\bibinfo{person}{Mehrdad Kiamari}, \bibinfo{person}{Gowri
  Ramachandran}, \bibinfo{person}{Quynh Nguyen}, \bibinfo{person}{Eva Pereira},
  \bibinfo{person}{Jeanne Holm}, {and} \bibinfo{person}{Bhaskar
  Krishnamachari}.} \bibinfo{year}{2020}\natexlab{}.
\newblock \showarticletitle{COVID-19 Risk Estimation using a Time-varying
  SIR-model}.
\newblock \bibinfo{journal}{\emph{1st ACM SIGSPATIAL International Workshop on
  Modeling and Understanding the Spread of COVID-19}} (\bibinfo{year}{2020}).
\newblock


\bibitem[\protect\citeauthoryear{Kresin, Schoenberg, and Mohler}{Kresin
  et~al\mbox{.}}{2020}]%
        {kresincomparison}
\bibfield{author}{\bibinfo{person}{Conor Kresin},
  \bibinfo{person}{Frederic~Paik Schoenberg}, {and} \bibinfo{person}{George
  Mohler}.} \bibinfo{year}{2020}\natexlab{}.
\newblock \showarticletitle{Comparison of the Hawkes and SEIR models for the
  spread of Covid-19}.
\newblock  (\bibinfo{year}{2020}).
\newblock


\bibitem[\protect\citeauthoryear{Li, Blakeley, et~al\mbox{.}}{Li
  et~al\mbox{.}}{2011}]%
        {li2011failure}
\bibfield{author}{\bibinfo{person}{Jing Li}, \bibinfo{person}{Daniel Blakeley},
  {et~al\mbox{.}}} \bibinfo{year}{2011}\natexlab{}.
\newblock \showarticletitle{The failure of R 0}.
\newblock \bibinfo{journal}{\emph{Computational and mathematical methods in
  medicine}}  \bibinfo{volume}{2011} (\bibinfo{year}{2011}).
\newblock


\bibitem[\protect\citeauthoryear{Li, Guan, Wu, Wang, Zhou, Tong, Ren, Leung,
  Lau, Wong, et~al\mbox{.}}{Li et~al\mbox{.}}{2020}]%
        {li2020early}
\bibfield{author}{\bibinfo{person}{Qun Li}, \bibinfo{person}{Xuhua Guan},
  \bibinfo{person}{Peng Wu}, \bibinfo{person}{Xiaoye Wang},
  \bibinfo{person}{Lei Zhou}, \bibinfo{person}{Yeqing Tong},
  \bibinfo{person}{Ruiqi Ren}, \bibinfo{person}{Kathy~SM Leung},
  \bibinfo{person}{Eric~HY Lau}, \bibinfo{person}{Jessica~Y Wong},
  {et~al\mbox{.}}} \bibinfo{year}{2020}\natexlab{}.
\newblock \showarticletitle{Early transmission dynamics in Wuhan, China, of
  novel coronavirus--infected pneumonia}.
\newblock \bibinfo{journal}{\emph{New England Journal of Medicine}}
  (\bibinfo{year}{2020}).
\newblock


\bibitem[\protect\citeauthoryear{Li, Yu, Shahabi, and Liu}{Li
  et~al\mbox{.}}{2018}]%
        {li2018diffusion}
\bibfield{author}{\bibinfo{person}{Yaguang Li}, \bibinfo{person}{Rose Yu},
  \bibinfo{person}{Cyrus Shahabi}, {and} \bibinfo{person}{Yan Liu}.}
  \bibinfo{year}{2018}\natexlab{}.
\newblock \showarticletitle{Diffusion Convolutional Recurrent Neural Network:
  Data-Driven Traffic Forecasting}. In \bibinfo{booktitle}{\emph{International
  Conference on Learning Representations}}.
\newblock
\urldef\tempurl%
\url{https://openreview.net/forum?id=SJiHXGWAZ}
\showURL{%
\tempurl}


\bibitem[\protect\citeauthoryear{Lotfi, Hamblin, and Rezaei}{Lotfi
  et~al\mbox{.}}{2020}]%
        {lotfi2020covid}
\bibfield{author}{\bibinfo{person}{Melika Lotfi}, \bibinfo{person}{Michael~R
  Hamblin}, {and} \bibinfo{person}{Nima Rezaei}.}
  \bibinfo{year}{2020}\natexlab{}.
\newblock \showarticletitle{COVID-19: Transmission, prevention, and potential
  therapeutic opportunities}.
\newblock \bibinfo{journal}{\emph{Clinica Chimica Acta}}
  (\bibinfo{year}{2020}).
\newblock


\bibitem[\protect\citeauthoryear{Meyer, Held, and H{\"o}hle}{Meyer
  et~al\mbox{.}}{2015}]%
        {meyer2015spatio}
\bibfield{author}{\bibinfo{person}{Sebastian Meyer}, \bibinfo{person}{Leonhard
  Held}, {and} \bibinfo{person}{Michael H{\"o}hle}.}
  \bibinfo{year}{2015}\natexlab{}.
\newblock \showarticletitle{Spatiotemporal analysis of epidemic phenomena using
  the R package surveillance}.
\newblock \bibinfo{journal}{\emph{Statistics-Computation}}
  \bibinfo{number}{1411.0416} (\bibinfo{year}{2015}).
\newblock


\bibitem[\protect\citeauthoryear{Miller, Foti, Lewnard, Jewell, Guestrin, and
  Fox}{Miller et~al\mbox{.}}{2020}]%
        {Miller2020}
\bibfield{author}{\bibinfo{person}{Andrew~C Miller},
  \bibinfo{person}{Nicholas~J Foti}, \bibinfo{person}{Joseph~A Lewnard},
  \bibinfo{person}{Nicholas~P Jewell}, \bibinfo{person}{Carlos Guestrin}, {and}
  \bibinfo{person}{Emily~B Fox}.} \bibinfo{year}{2020}\natexlab{}.
\newblock \showarticletitle{Mobility trends provide a leading indicator of
  changes in SARS-CoV-2 transmission}.
\newblock \bibinfo{journal}{\emph{medRxiv}} (\bibinfo{year}{2020}).
\newblock
\urldef\tempurl%
\url{https://doi.org/10.1101/2020.05.07.20094441}
\showDOI{\tempurl}
\showeprint{https://www.medrxiv.org/content/early/2020/05/11/2020.05.07.20094441.full.pdf}


\bibitem[\protect\citeauthoryear{Mohler, Schoenberg, Short, and Sledge}{Mohler
  et~al\mbox{.}}{2020}]%
        {mohler2020analyzing}
\bibfield{author}{\bibinfo{person}{George Mohler}, \bibinfo{person}{Frederic
  Schoenberg}, \bibinfo{person}{Martin~B Short}, {and} \bibinfo{person}{Daniel
  Sledge}.} \bibinfo{year}{2020}\natexlab{}.
\newblock \showarticletitle{Analyzing the World-Wide Impact of Public Health
  Interventions on the Transmission Dynamics of COVID-19}.
\newblock \bibinfo{journal}{\emph{arXiv preprint arXiv:2004.01714}}
  (\bibinfo{year}{2020}).
\newblock


\bibitem[\protect\citeauthoryear{Osthus, Hickmann, Caragea, Higdon, and
  Del~Valle}{Osthus et~al\mbox{.}}{2017}]%
        {osthus2017forecasting}
\bibfield{author}{\bibinfo{person}{Dave Osthus}, \bibinfo{person}{Kyle~S
  Hickmann}, \bibinfo{person}{Petru{\c{t}}a~C Caragea}, \bibinfo{person}{Dave
  Higdon}, {and} \bibinfo{person}{Sara~Y Del~Valle}.}
  \bibinfo{year}{2017}\natexlab{}.
\newblock \showarticletitle{Forecasting seasonal influenza with a state-space
  SIR model}.
\newblock \bibinfo{journal}{\emph{The annals of applied statistics}}
  \bibinfo{volume}{11}, \bibinfo{number}{1} (\bibinfo{year}{2017}),
  \bibinfo{pages}{202}.
\newblock


\bibitem[\protect\citeauthoryear{Pei and Shaman}{Pei and Shaman}{2020}]%
        {pei2020initial}
\bibfield{author}{\bibinfo{person}{Sen Pei} {and} \bibinfo{person}{Jeffrey
  Shaman}.} \bibinfo{year}{2020}\natexlab{}.
\newblock \showarticletitle{Initial Simulation of SARS-CoV2 Spread and
  Intervention Effects in the Continental US}.
\newblock \bibinfo{journal}{\emph{medRxiv}} (\bibinfo{year}{2020}).
\newblock


\bibitem[\protect\citeauthoryear{Rizoiu, Mishra, Kong, Carman, and Xie}{Rizoiu
  et~al\mbox{.}}{2018}]%
        {rizoiu2018sir}
\bibfield{author}{\bibinfo{person}{Marian-Andrei Rizoiu},
  \bibinfo{person}{Swapnil Mishra}, \bibinfo{person}{Quyu Kong},
  \bibinfo{person}{Mark Carman}, {and} \bibinfo{person}{Lexing Xie}.}
  \bibinfo{year}{2018}\natexlab{}.
\newblock \showarticletitle{SIR-Hawkes: linking epidemic models and Hawkes
  processes to model diffusions in finite populations}. In
  \bibinfo{booktitle}{\emph{Proceedings of the 2018 World Wide Web
  Conference}}. \bibinfo{pages}{419--428}.
\newblock


\bibitem[\protect\citeauthoryear{Roosa and Chowell}{Roosa and Chowell}{2019}]%
        {roosa2019assessing}
\bibfield{author}{\bibinfo{person}{Kimberlyn Roosa} {and}
  \bibinfo{person}{Gerardo Chowell}.} \bibinfo{year}{2019}\natexlab{}.
\newblock \showarticletitle{Assessing parameter identifiability in
  compartmental dynamic models using a computational approach: application to
  infectious disease transmission models}.
\newblock \bibinfo{journal}{\emph{Theoretical Biology and Medical Modelling}}
  \bibinfo{volume}{16}, \bibinfo{number}{1} (\bibinfo{year}{2019}),
  \bibinfo{pages}{1}.
\newblock


\bibitem[\protect\citeauthoryear{Times}{Times}{2020a}]%
        {infection_count}
\bibfield{author}{\bibinfo{person}{New~York Times}.}
  \bibinfo{year}{2020}\natexlab{a}.
\newblock \bibinfo{title}{Actual Coronavirus Infections Vastly Under counted,
  C.D.C. Data Shows}.
\newblock
  \bibinfo{howpublished}{\url{https://www.nytimes.com/2020/06/27/health/coronavirus-antibodies-asymptomatic.html}}.
\newblock
\newblock
\shownote{Accessed: 2020-10-27.}


\bibitem[\protect\citeauthoryear{Times}{Times}{2020b}]%
        {pandemicfatigue1}
\bibfield{author}{\bibinfo{person}{New~York Times}.}
  \bibinfo{year}{2020}\natexlab{b}.
\newblock \bibinfo{title}{As the Coronavirus Surges, a New Culprit Emerges:
  Pandemic Fatigue}.
\newblock
  \bibinfo{howpublished}{\url{https://www.nytimes.com/2020/10/17/us/coronavirus-pandemic-fatigue.html}}.
\newblock
\newblock
\shownote{Accessed: 2020-10-29.}


\bibitem[\protect\citeauthoryear{Times}{Times}{2020c}]%
        {pandemicfatigue2}
\bibfield{author}{\bibinfo{person}{New~York Times}.}
  \bibinfo{year}{2020}\natexlab{c}.
\newblock \bibinfo{title}{As Virus Surges in Europe, Resistance to New
  Restrictions Also Grows}.
\newblock
  \bibinfo{howpublished}{\url{https://www.nytimes.com/2020/10/09/world/europe/coronavirus-europe-fatigue.html}}.
\newblock
\newblock
\shownote{Accessed: 2020-10-29.}


\bibitem[\protect\citeauthoryear{Wiersinga, Rhodes, Cheng, Peacock, and
  Prescott}{Wiersinga et~al\mbox{.}}{2020}]%
        {wiersinga2020pathophysiology}
\bibfield{author}{\bibinfo{person}{W~Joost Wiersinga}, \bibinfo{person}{Andrew
  Rhodes}, \bibinfo{person}{Allen~C Cheng}, \bibinfo{person}{Sharon~J Peacock},
  {and} \bibinfo{person}{Hallie~C Prescott}.} \bibinfo{year}{2020}\natexlab{}.
\newblock \showarticletitle{Pathophysiology, transmission, diagnosis, and
  treatment of coronavirus disease 2019 (COVID-19): a review}.
\newblock \bibinfo{journal}{\emph{Jama}} \bibinfo{volume}{324},
  \bibinfo{number}{8} (\bibinfo{year}{2020}), \bibinfo{pages}{782--793}.
\newblock


\bibitem[\protect\citeauthoryear{Yuan, Li, Bertozzi, Brantingham, and
  Porter}{Yuan et~al\mbox{.}}{2019}]%
        {yuan2019multivariate}
\bibfield{author}{\bibinfo{person}{Baichuan Yuan}, \bibinfo{person}{Hao Li},
  \bibinfo{person}{Andrea~L Bertozzi}, \bibinfo{person}{P~Jeffrey Brantingham},
  {and} \bibinfo{person}{Mason~A Porter}.} \bibinfo{year}{2019}\natexlab{}.
\newblock \showarticletitle{Multivariate spatiotemporal hawkes processes and
  network reconstruction}.
\newblock \bibinfo{journal}{\emph{SIAM Journal on Mathematics of Data Science}}
  \bibinfo{volume}{1}, \bibinfo{number}{2} (\bibinfo{year}{2019}),
  \bibinfo{pages}{356--382}.
\newblock


\bibitem[\protect\citeauthoryear{Zou, Wang, Xu, Chen, Zhang, and Gu}{Zou
  et~al\mbox{.}}{2020}]%
        {zou2020epidemic}
\bibfield{author}{\bibinfo{person}{Difan Zou}, \bibinfo{person}{Lingxiao Wang},
  \bibinfo{person}{Pan Xu}, \bibinfo{person}{Jinghui Chen},
  \bibinfo{person}{Weitong Zhang}, {and} \bibinfo{person}{Quanquan Gu}.}
  \bibinfo{year}{2020}\natexlab{}.
\newblock \showarticletitle{Epidemic Model Guided Machine Learning for COVID-19
  Forecasts in the United States}.
\newblock \bibinfo{journal}{\emph{medRxiv}} (\bibinfo{year}{2020}).
\newblock


\end{thebibliography}

\clearpage
\appendix

\section{Additional Experimental Results}\label{sec:exp_additional}
\begin{figure}
    \centering
    \begin{tabular}{cccc}
    & \hspace{-15pt} Day $10$ & \hspace{-30pt} Day $15$ & \hspace{-30pt} Day $20$ \\ 
      \rotatebox{90}{\hspace{1.3cm} Dec. `19} & \includegraphics[width=0.33\textwidth]{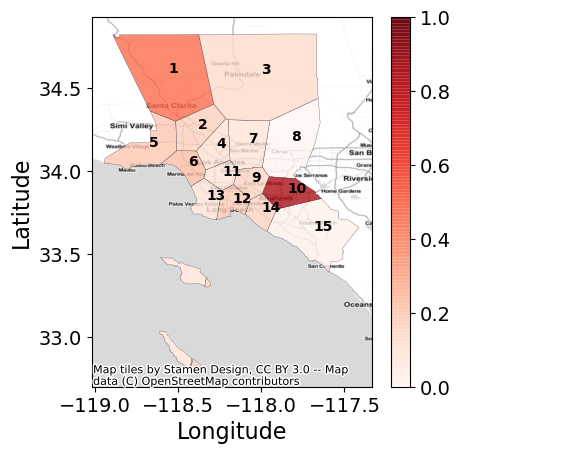} & \hspace{-15pt}
       \includegraphics[width=0.33\textwidth]{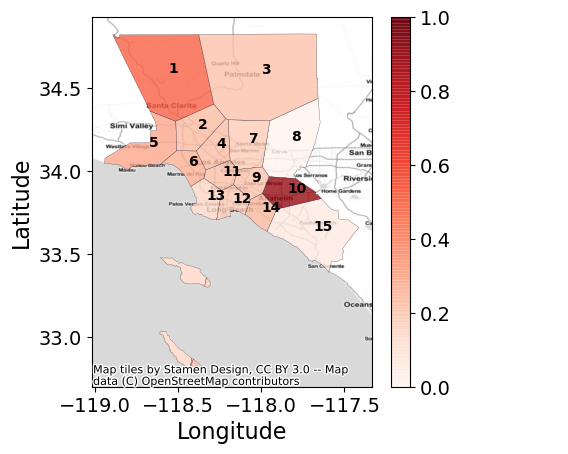}& \hspace{-15pt}
       \includegraphics[width=0.33\textwidth]{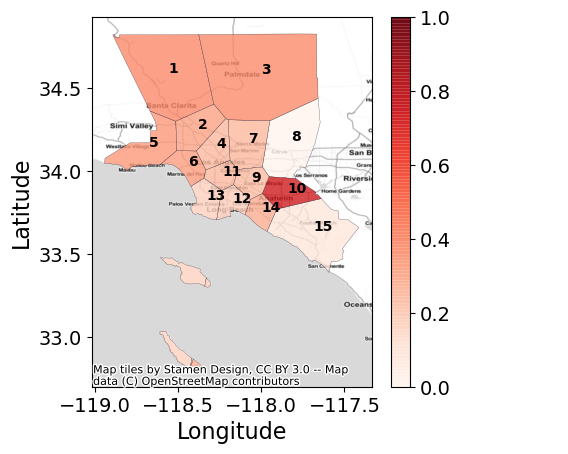} \vspace{-5pt} \\
       & \hspace{-20pt} {\footnotesize (a-i)} & \hspace{-30pt} {\footnotesize (a-ii)} & \hspace{-30pt} {\footnotesize (a-iii)}\\
        \rotatebox{90}{ \hspace{1.3cm} Jan. `20}& \includegraphics[width=0.33\textwidth]{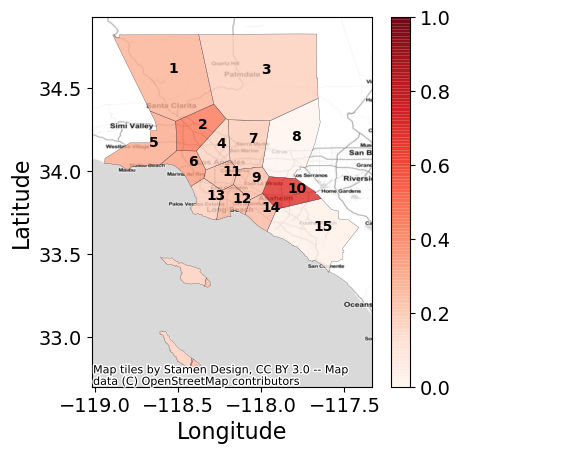} & \hspace{-15pt}
       \includegraphics[width=0.33\textwidth]{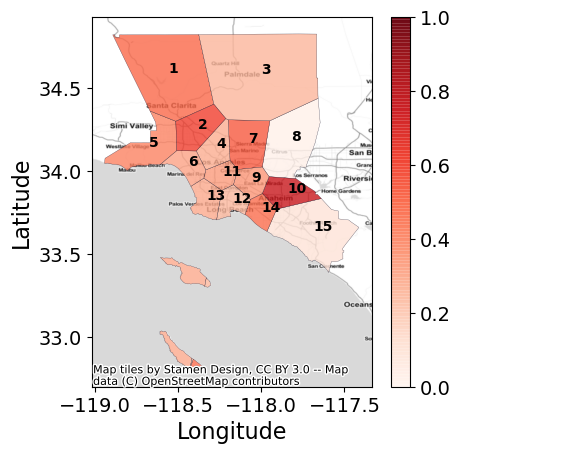} & \hspace{-15pt}
       \includegraphics[width=0.33\textwidth]{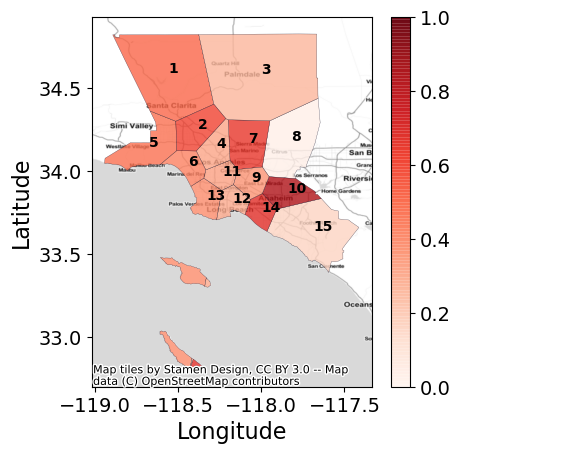} \vspace{-5pt} \\
        & \hspace{-20pt} {\footnotesize(b-i)} & \hspace{-30pt} {\footnotesize (b-ii)} & \hspace{-30pt} {\footnotesize (b-iii)}\\
     \rotatebox{90}{\hspace{1.3cm} Mar. `20}& \includegraphics[width=0.33\textwidth]{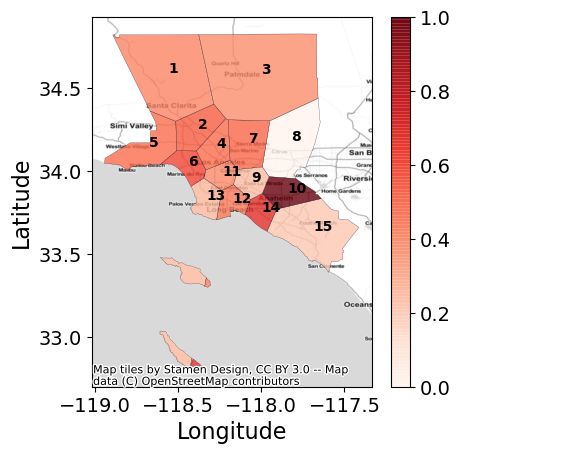}  & \hspace{-15pt}
       \includegraphics[width=0.33\textwidth]{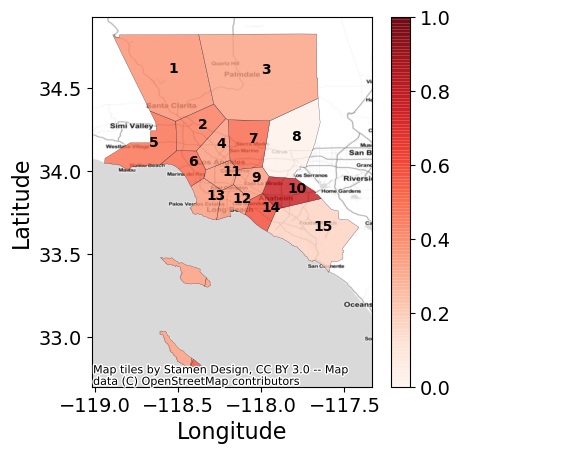} & \hspace{-15pt}
       \includegraphics[width=0.33\textwidth]{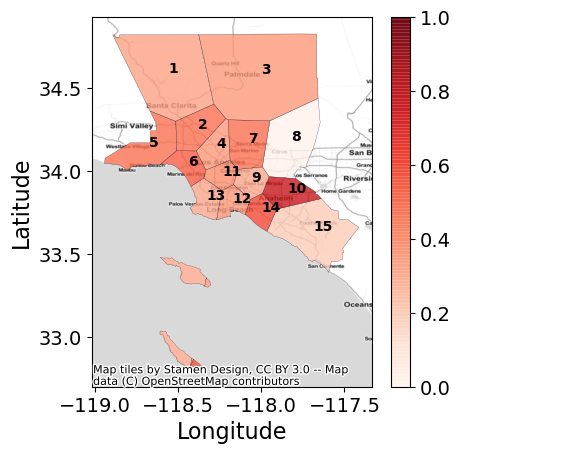} \vspace{-5pt} \\
        &  \hspace{-20pt} {\footnotesize (c-i)} & \hspace{-30pt} {\footnotesize(c-ii)} & \hspace{-30pt} {\footnotesize(c-iii)}\\
        &\hspace{-35pt} \includegraphics[width=0.28\textwidth]{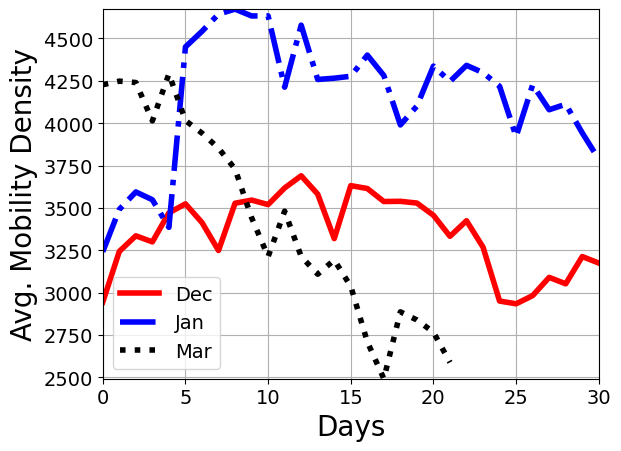}  & \hspace{-45pt}
       \includegraphics[width=0.27\textwidth]{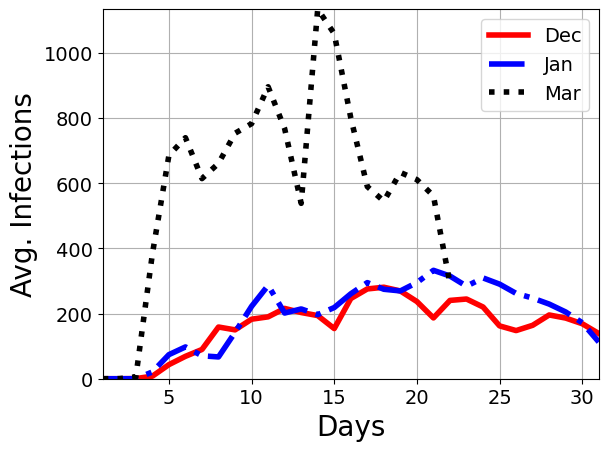} & \hspace{-45pt}
       \includegraphics[width=0.26\textwidth]{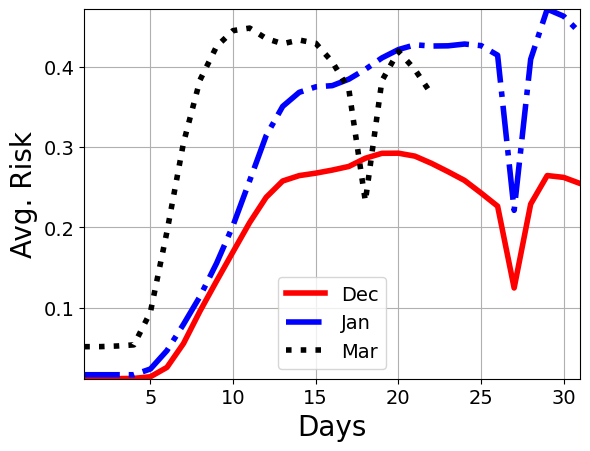} \vspace{-5pt} \\
        &  \hspace{-20pt} {\footnotesize (d-i)} & \hspace{-30pt} {\footnotesize(d-ii)} & \hspace{-30pt} {\footnotesize(d-iii)}\\
    \end{tabular}
    \vspace{-10pt}
    \caption{Comparing the Risk across months in different regions of Los Angeles, CA. Panels (a), (b), and (c) show the evaluated risk ($\rho$) for each cluster (marked via numbers) for the months of Dec `19, Jan `20, and Mar `20, respectively. The panels (i), (ii), and (iii) show the risk varying over day 10, 15, and 20, respectively. In addition, panel (d-i), (d-ii), and (d-iii) show the comparison between the average number of location signals, infections and risk, across clusters over months, respectively. We observe that while the risk for months of Dec and Jan show similar trends for different days, the month of Mar has lower risk. This can be attributed to the drop in mobility.}
    \label{fig:LA}
\end{figure}

\begin{figure}
    \centering
    \begin{tabular}{cccc}
      & \hspace{0pt} Day $10$ & \hspace{-5pt} Day $15$ & \hspace{-5pt} Day $20$ \\ 
      \rotatebox{90}{\hspace{0.9cm} Dec. `19}& \includegraphics[width=0.3\textwidth]{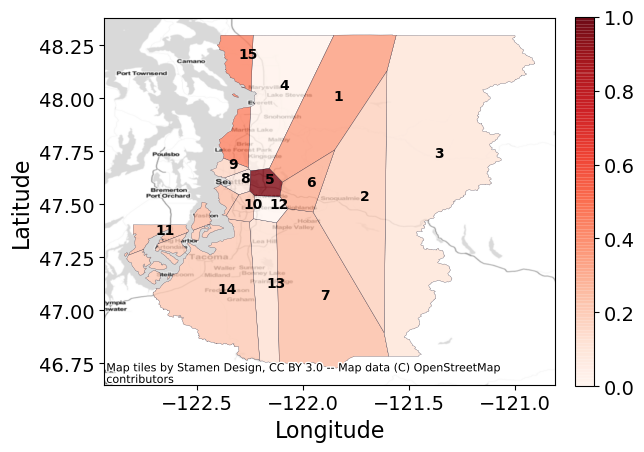}  & \hspace{-8pt}
       \includegraphics[width=0.3\textwidth]{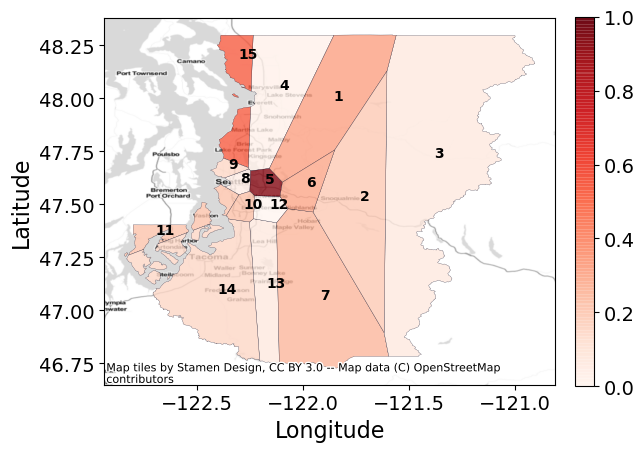} & \hspace{-8pt}
       \includegraphics[width=0.3\textwidth]{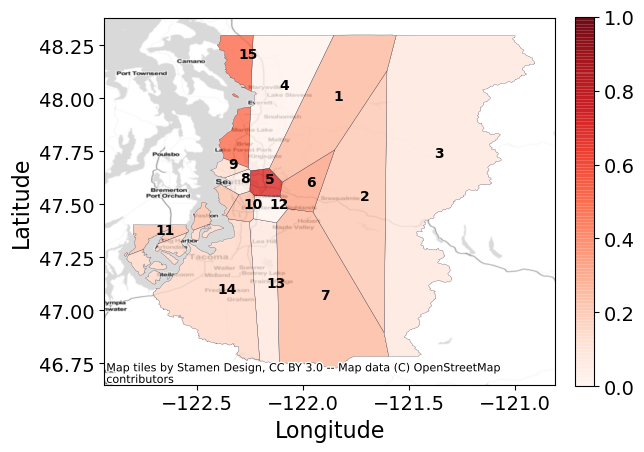} \vspace{-5pt}\\
       & {\footnotesize (a-i)} & \hspace{-5pt} {\footnotesize (a-ii)} & \hspace{-5pt} {\footnotesize (a-iii)}\\
        \rotatebox{90}{\hspace{0.9cm} Jan. `20}& \includegraphics[width=0.3\textwidth]{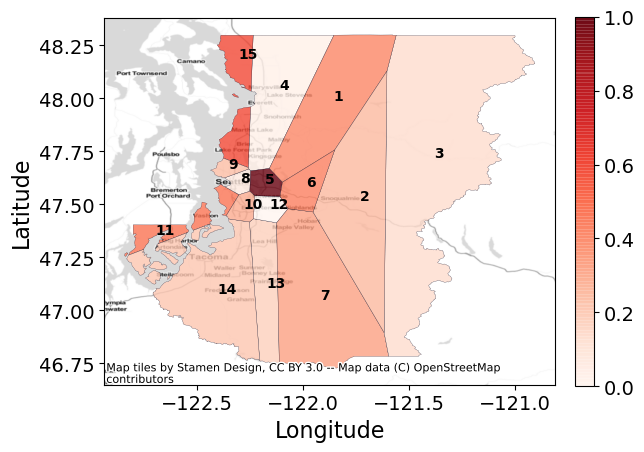} & \hspace{-8pt}
       \includegraphics[width=0.3\textwidth]{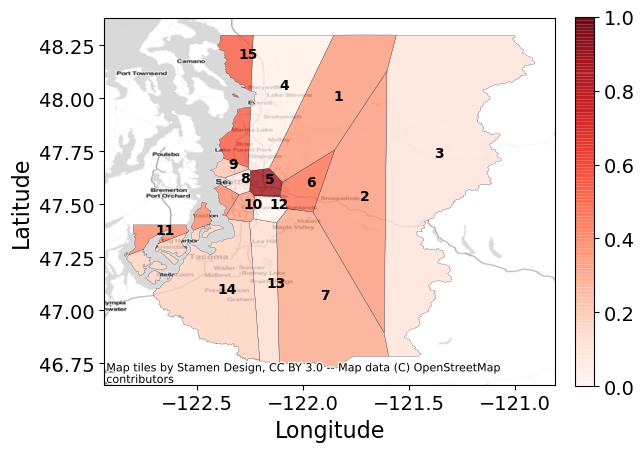} & \hspace{-8pt}
       \includegraphics[width=0.3\textwidth]{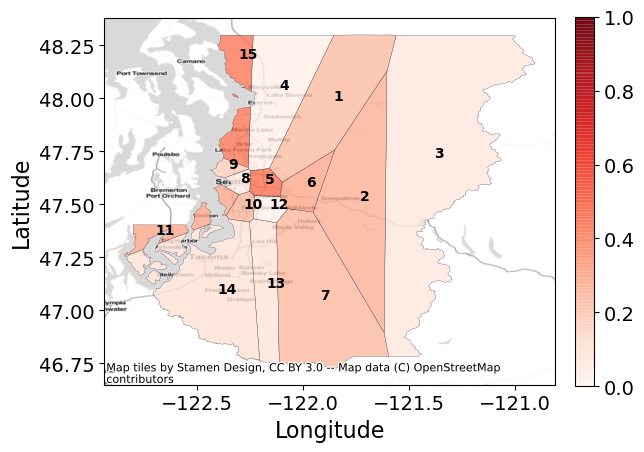} \vspace{-5pt} \\
           &  {\footnotesize(b-i)} & \hspace{-5pt} {\footnotesize (b-ii)} & \hspace{-5pt} {\footnotesize (b-iii)}\\
     \rotatebox{90}{ \hspace{0.9cm} Mar. `20}&\includegraphics[width=0.3\textwidth]{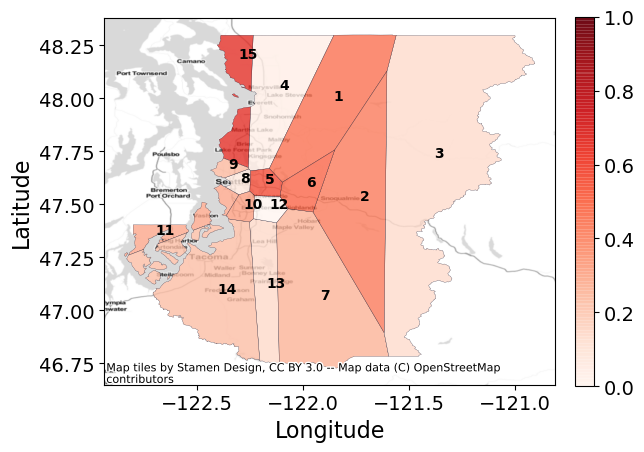} & \hspace{-8pt}
       \includegraphics[width=0.3\textwidth]{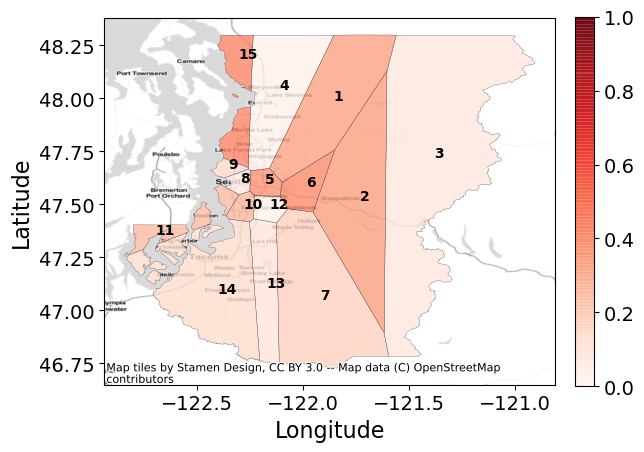} & \hspace{-8pt}
       \includegraphics[width=0.3\textwidth]{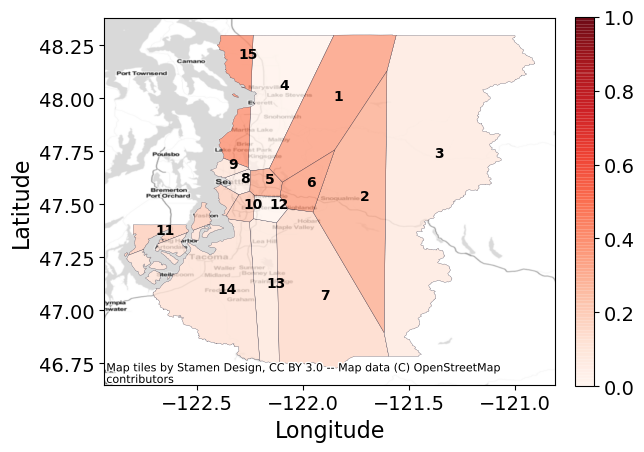} \vspace{-5pt} \\
        & {\footnotesize (c-i)} & \hspace{-5pt} {\footnotesize(c-ii)} & \hspace{-5pt} {\footnotesize(c-iii)}\\
         &\hspace{-17pt} \includegraphics[width=0.29\textwidth]{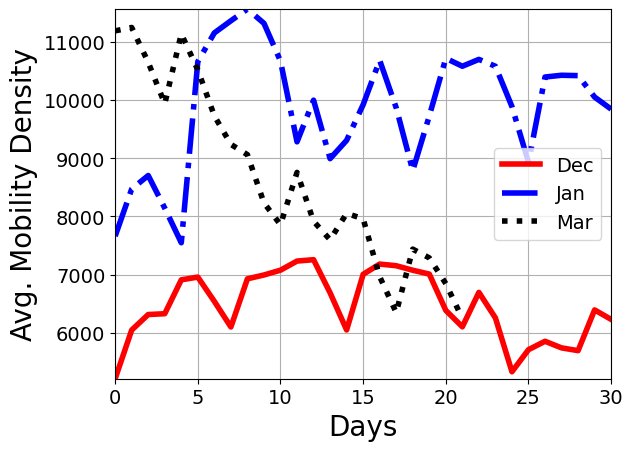}  & \hspace{-17pt}
       \includegraphics[width=0.27\textwidth]{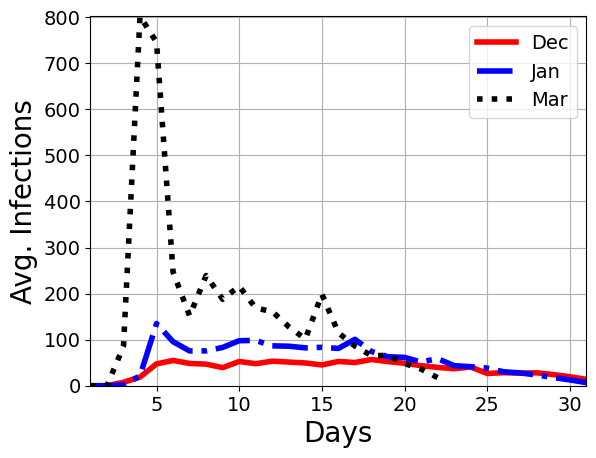} & \hspace{-17pt}
       \includegraphics[width=0.27\textwidth]{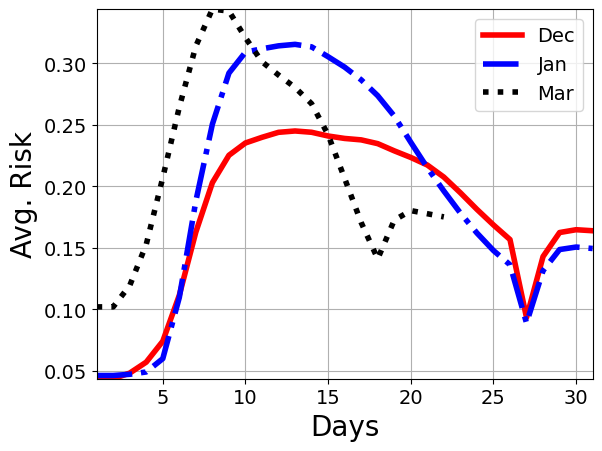} \vspace{-5pt} \\
        & {\footnotesize (d-i)} & \hspace{-5pt} {\footnotesize(d-ii)} & \hspace{-5pt} {\footnotesize(d-iii)}\\
    \end{tabular}
    \vspace{-10pt}
    \caption{Comparing the Risk across months in different regions of Seattle, WA. Panels (a), (b), and (c) show the evaluated risk ($\rho$) for each cluster (marked via numbers) for the months of Dec `19, Jan `20, and Mar `20, respectively. The panels (i), (ii), and (iii) show the risk varying over day 10, 15, and 20, respectively. In addition, panel (d-i), (d-ii), and (d-iii) show the comparison between the average number of location signals, infections and risk, across clusters over months, respectively. We observe that while the risk for months of Dec and Jan show similar trends for different days, the month of Mar has lower risk. This can be attributed to the drop in mobility. }
    \label{fig:sea}
\end{figure}

\begin{figure}
    \centering
    \resizebox{1\textwidth}{!}{
    \begin{tabular}{cccc}
      & \hspace{0pt} Day $10$ & \hspace{-5pt} Day $15$ & \hspace{-5pt} Day $20$ \\ 
      \rotatebox{90}{\hspace{1.1cm} Dec. `19}& \hspace{15pt} \includegraphics[width=0.35\textwidth]{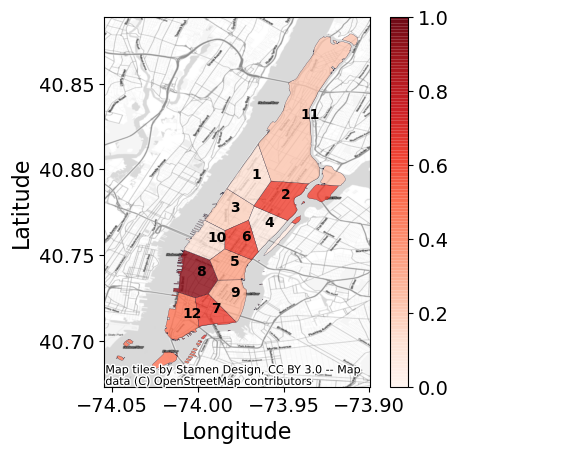}  & \hspace{8pt}
       \includegraphics[width=0.35\textwidth]{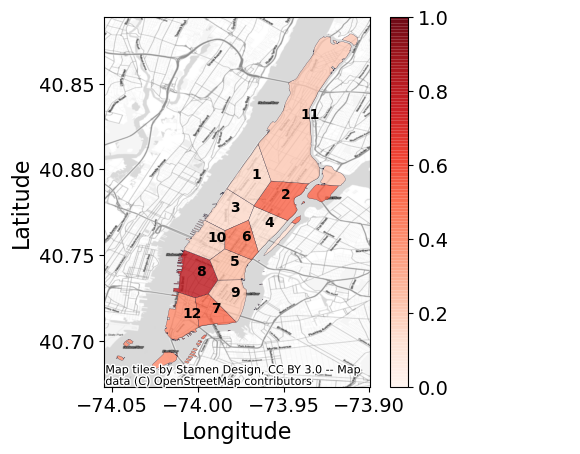} & \hspace{8pt}
       \includegraphics[width=0.35\textwidth]{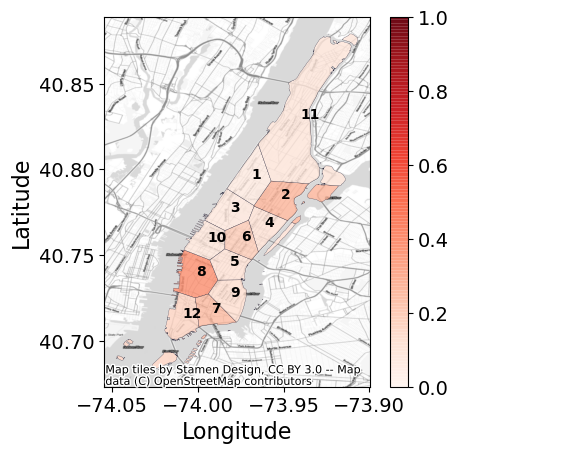} \vspace{-5pt}\\
       & {\footnotesize (a-i)} & \hspace{-5pt} {\footnotesize (a-ii)} & \hspace{-5pt} {\footnotesize (a-iii)}\\
        \rotatebox{90}{\hspace{1.1cm} Jan. `20}& \hspace{15pt}\includegraphics[width=0.35\textwidth]{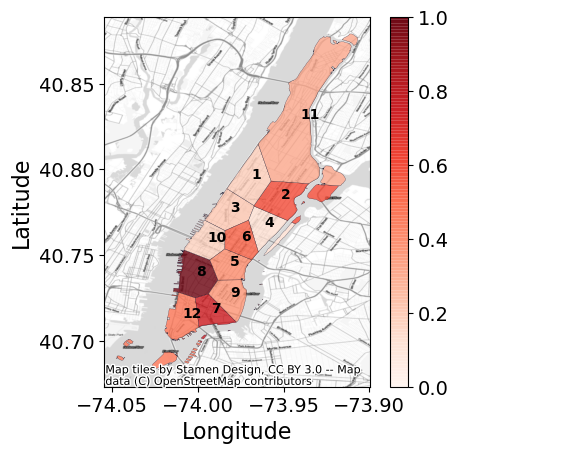} & \hspace{8pt}
       \includegraphics[width=0.35\textwidth]{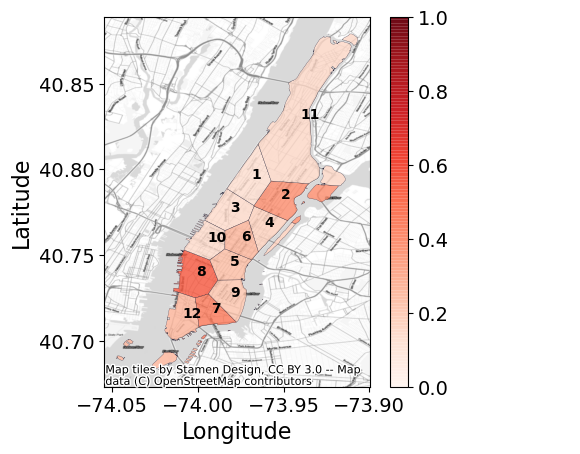} & \hspace{8pt}
       \includegraphics[width=0.35\textwidth]{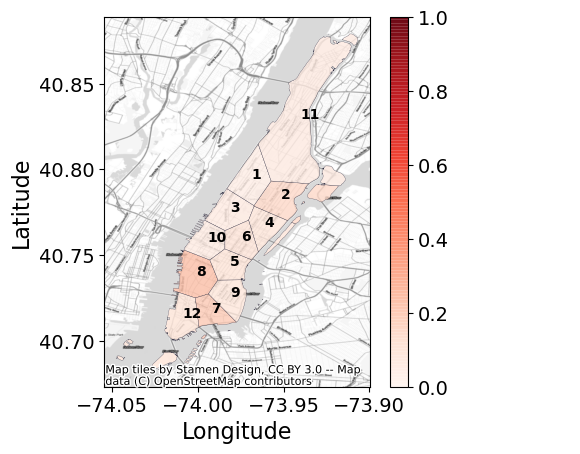} \vspace{-5pt} \\
           &  {\footnotesize(b-i)} & \hspace{-5pt} {\footnotesize (b-ii)} & \hspace{-5pt} {\footnotesize (b-iii)}\\
     \rotatebox{90}{ \hspace{1.1cm} Mar. `20}& \hspace{15pt} \includegraphics[width=0.35\textwidth]{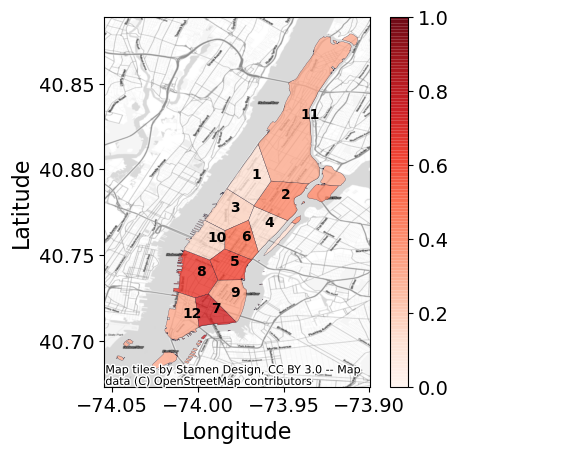} & \hspace{8pt}
       \includegraphics[width=0.35\textwidth]{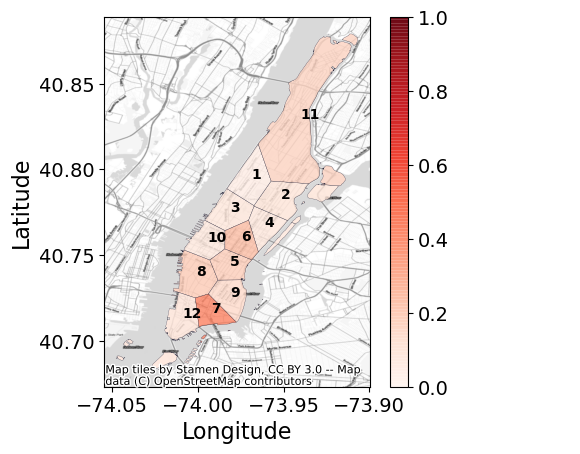} & \hspace{8pt}
       \includegraphics[width=0.35\textwidth]{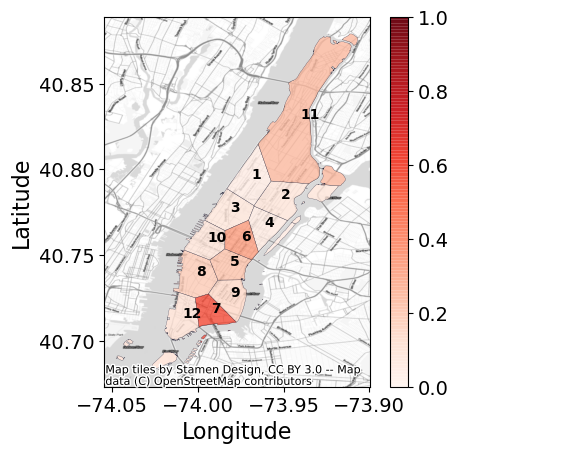} \vspace{-5pt} \\
        & {\footnotesize (c-i)} & \hspace{-5pt} {\footnotesize(c-ii)} & \hspace{-5pt} {\footnotesize(c-iii)}\\
         &\hspace{-17pt} \includegraphics[width=0.29\textwidth]{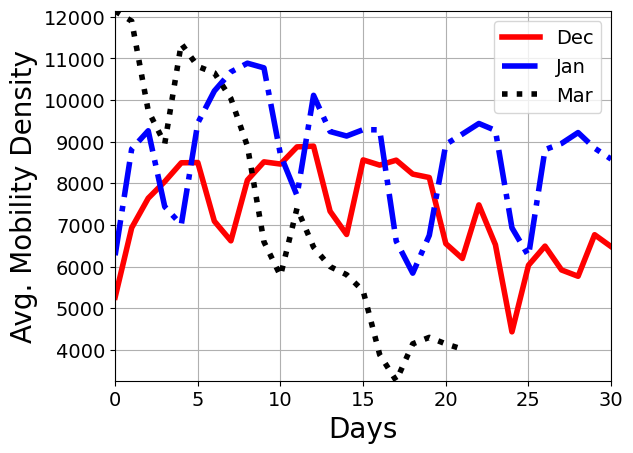}  & \hspace{-17pt}
       \includegraphics[width=0.27\textwidth]{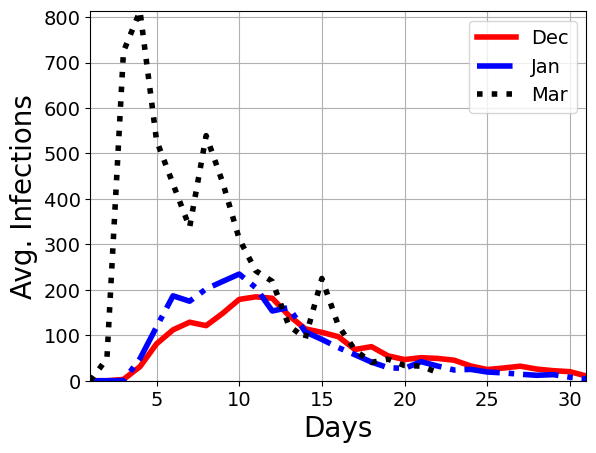} & \hspace{-17pt}
       \includegraphics[width=0.27\textwidth]{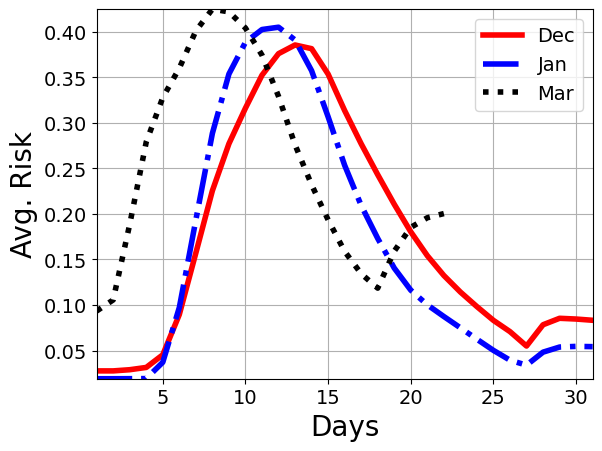} \vspace{-5pt} \\
        & {\footnotesize (d-i)} & \hspace{-5pt} {\footnotesize(d-ii)} & \hspace{-5pt} {\footnotesize(d-iii)}\\
    \end{tabular}}
    \vspace{-10pt}
    \caption{Comparing the Risk across months in different regions of Manhattan (New York), NY. Panels (a), (b), and (c) show the evaluated risk ($\rho$) for each cluster (marked via numbers) for the months of Dec `19, Jan `20, and Mar `20, respectively. The panels (i), (ii), and (iii) show the risk varying over day 10, 15, and 20, respectively. In addition, panel (d-i), (d-ii), and (d-iii) show the comparison between the average number of location signals, infections and risk, across clusters over months, respectively. We observe that while the risk for months of Dec and Jan show similar trends for different days, the month of Mar has lower risk. This can be attributed to the drop in mobility. }
    \label{fig:ny}
\end{figure}

\begin{figure}
    \centering
    \resizebox{1\textwidth}{!}{
    \begin{tabular}{cccc}
    & \hspace{0pt} Day $10$ & \hspace{-5pt} Day $15$ & \hspace{-5pt} Day $20$ \\ 
      \rotatebox{90}{\hspace{0.9cm} Dec. `19} \hspace{-8pt}&  \includegraphics[width=0.3\textwidth]{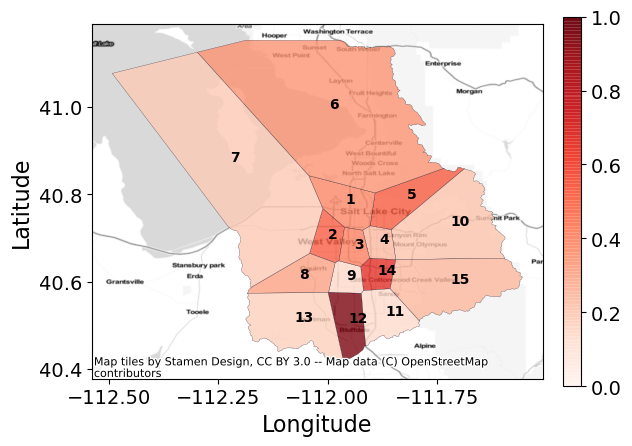}  & \hspace{-15pt}
       \includegraphics[width=0.3\textwidth]{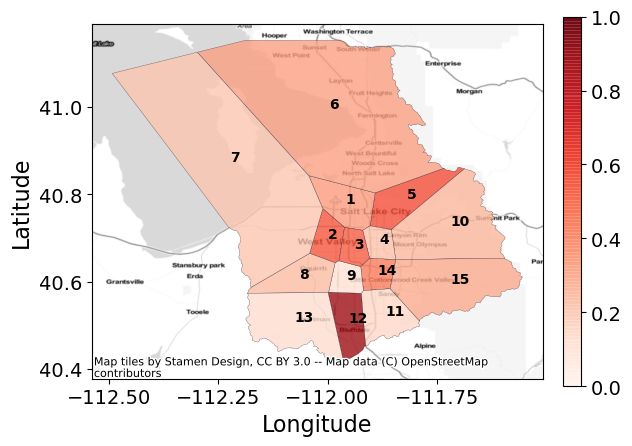} & \hspace{-15pt}
       \includegraphics[width=0.3\textwidth]{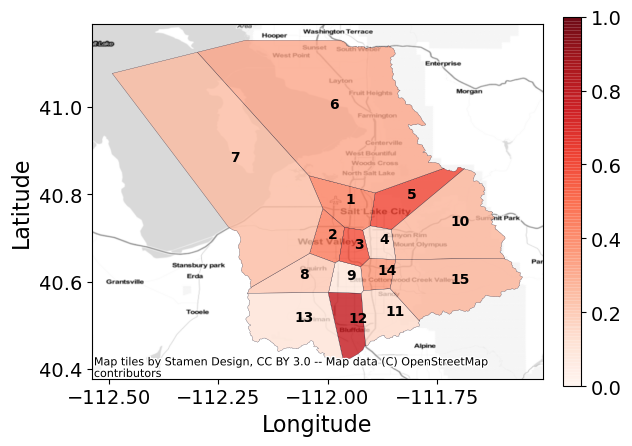} \hspace{-5pt}\\
       & \hspace{0pt} {\footnotesize (a-i)} & \hspace{-5pt} {\footnotesize (a-ii)} & \hspace{-5pt} {\footnotesize (a-iii)}\\
       \rotatebox{90}{\hspace{0.9cm} Jan. `20} \hspace{-8pt}& \includegraphics[width=0.3\textwidth]{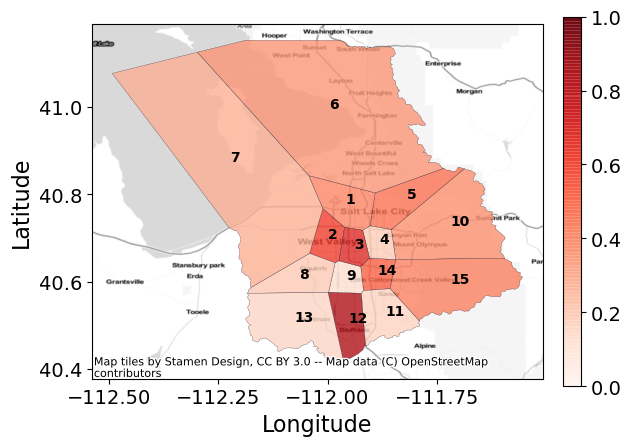} & \hspace{-15pt}
       \includegraphics[width=0.3\textwidth]{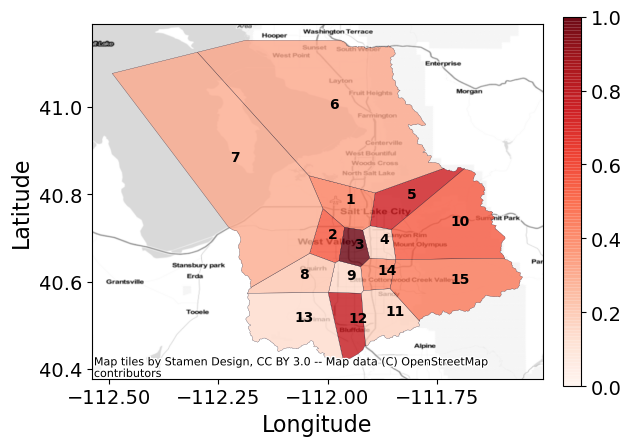} & \hspace{-15pt}
       \includegraphics[width=0.3\textwidth]{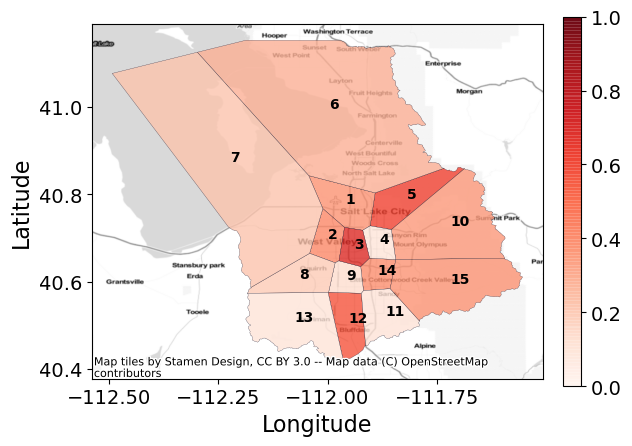}\hspace{-5pt} \\
          & \hspace{0pt} {\footnotesize(b-i)} & \hspace{-5pt} {\footnotesize (b-ii)} & \hspace{-5pt} {\footnotesize (b-iii)}\\
     \rotatebox{90}{ \hspace{0.9cm} Mar. `20} \hspace{-8pt}&\includegraphics[width=0.3\textwidth]{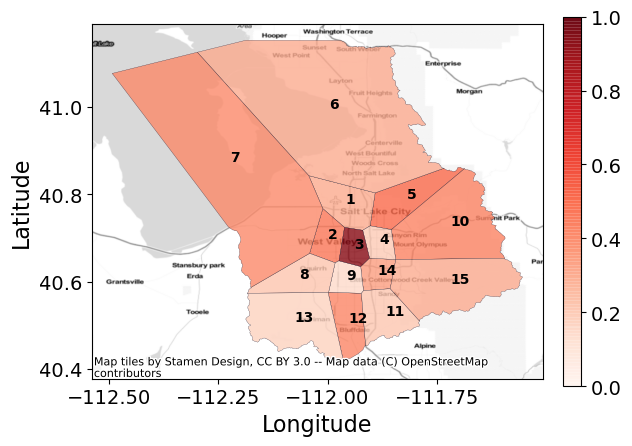} &
       \includegraphics[width=0.3\textwidth]{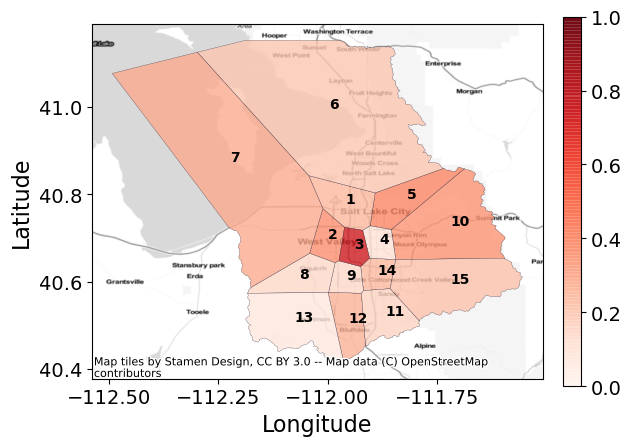} & 
       \includegraphics[width=0.3\textwidth]{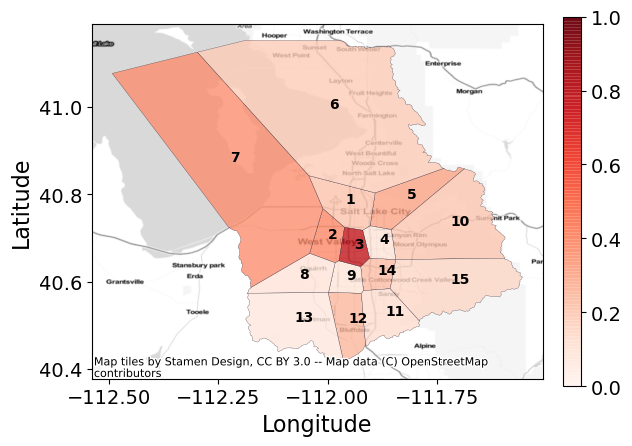} \\
        &  \hspace{0pt} {\footnotesize (c-i)} & \hspace{-5pt} {\footnotesize(c-ii)} & \hspace{-5pt} {\footnotesize(c-iii)}\\
        &\hspace{-17pt} \includegraphics[width=0.29\textwidth]{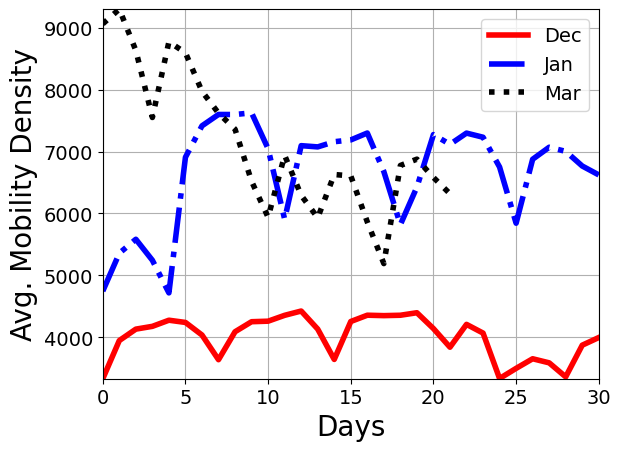}  & \hspace{-17pt}
       \includegraphics[width=0.27\textwidth]{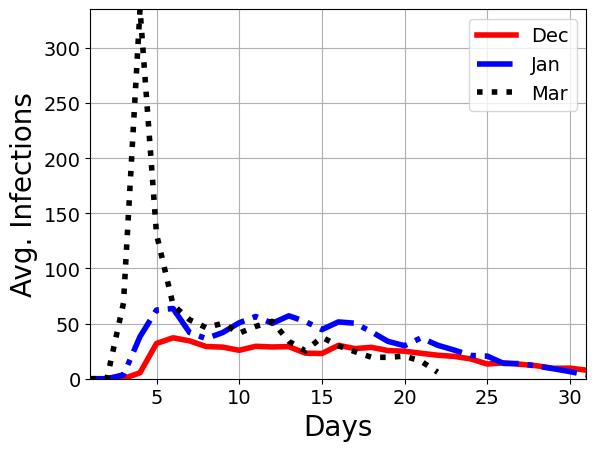} & \hspace{-17pt}
       \includegraphics[width=0.27\textwidth]{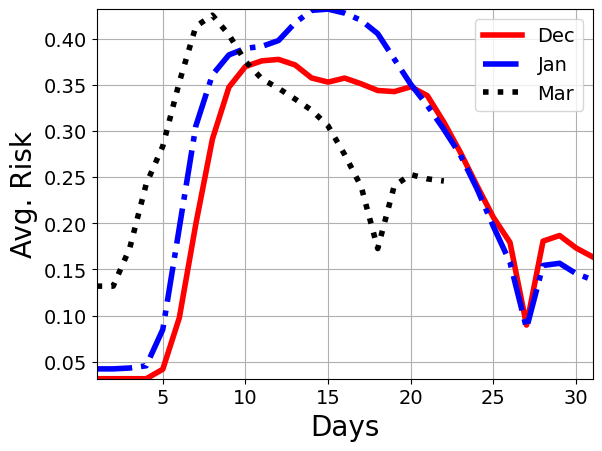} \vspace{-5pt} \\
        & {\footnotesize (d-i)} & \hspace{-5pt} {\footnotesize(d-ii)} & \hspace{-5pt} {\footnotesize(d-iii)}\\

    \end{tabular}}
    \vspace{-10pt}
    \caption{Comparing the Risk across months in different regions of Salt Lake County, UT. Panels (a), (b), and (c) show the evaluated risk ($\rho$) for each cluster (marked via numbers) for the months of Dec `19, Jan `20, and Mar `20, respectively. The panels (i), (ii), and (iii) show the risk varying over day 10, 15, and 20, respectively.In addition, panel (d-i), (d-ii), and (d-iii) show the comparison between the average number of location signals, infections and risk, across clusters over months, respectively. We observe that while the risk for months of Dec and Jan show similar trends for different days, the month of Mar has lower risk. This can be attributed to the drop in mobility. }
    \label{fig:slc}
\end{figure}

We ran our experiments, with the same setup described in Sec. \ref{sec:exp}, but for the areas around four more cities, namely Los Angeles, Seattle (King county), New York (Manhattan) and Salt Lake county. The statistics of these datasets can be found in Tables \ref{tab:checkin_data_extra} and \ref{tab:agent_extra}. The results are presented in Tables \ref{tab:LA_res}, \ref{tab:SEA_res}, \ref{tab:NY_res} and \ref{tab:SLC_res} and Figs. \ref{fig:LA}, \ref{fig:sea}, \ref{fig:ny} and \ref{fig:slc}. 

An interesting observation, in cities of New York (Manhattan) and Salt lake County shown in Fig~\ref{fig:ny} and Table~\ref{tab:NY_res}, and Fig~\ref{fig:slc} and Table~\ref{tab:SLC_res}, respectively, is that an increased mobility in Mar. 2020 leading to a large number of infections decreases model's capacity to predict later on when the infections drop. This is because the Hawkes process model tends to attribute the infections to the background rate $\mu_c$ rather than the mobility-dependent $R_c^t$. This leads to poor infection prediction performance (for all methods) since the model anticipates infections to happen at the relatively large background rate. In practice, such a modality can be avoided by using longer traces like \cite{Chiang2020}. Nevertheless, our risk scores \eqref{eq:risk} based on the intensity function $\lambda_c(t)$ \eqref{eq:intensity} are still faithful to the infections since they take into account both the background rate and the mobility-dependence. This in fact highlights the advantages and reliability of the proposed risk score metric. In addition, our risk score for Manhattan (New York) also illustrates its applicability for fine-grained spatiotemporal risk assignment.
\begin{table}[h]
    \centering
    \begin{minipage}{0.48\textwidth}
    \centering
    \resizebox{0.87\textwidth}{!}{
    \begin{tabular}{c|c|c|c}
        \textbf{City} & \textbf{December} & \textbf{January} & \textbf{March}  \\\hline
        Los Angeles & $331\times 10^6$ &282$\times 10^6$ & 171$\times 10^6$\\\hline
        Seattle & 53$\times 10^6$ & 66$\times 10^6$& 38$\times 10^6$\\\hline
        New York &41$\times 10^6$ & 36$\times 10^6$ & 21$\times 10^6$\\\hline
        Salt Lake & 31$\times 10^6$ &39$\times 10^6$ & 29$\times 10^6$
    \end{tabular}}
    \caption{Total No. location signals per Month}
    \label{tab:checkin_data_extra}
    \end{minipage}
    \hfill
    \begin{minipage}{0.48\textwidth}
    \centering
    \resizebox{0.87\textwidth}{!}{
    \begin{tabular}{c|c|c|c}
        \textbf{City} & \textbf{December} & \textbf{January} & \textbf{March}  \\\hline
        Los Angeles & 159$\times10^3$ &  171$\times10^3$&  131$\times10^3$\\\hline
        Seattle & 29$\times10^3$ & 38$\times10^3$ & 32$\times10^3$\\\hline
        New York & 40$\times10^3$ & 41$\times10^3$ & 32$\times10^3$\\\hline
        Salt Lake & 16$\times10^3$ & 22$\times10^3$& 23$\times10^3$
    \end{tabular}}
    \caption{No. Agents per Month}
    \label{tab:agent_extra}
    \end{minipage}
    \vspace{-20pt}
\end{table}

 \begin{table}[h]
    \centering
    \resizebox{0.98\textwidth}{!}{
    \begin{tabular}{c c c c c |c c c c |c c c c}
    \multirow{4}{*}{\textbf{\Large Model}} &  \multicolumn{12}{c}{\textbf{\Large Infection and Risk Prediction for Los Angeles, CA} \vspace{2pt}}\\\cline{2-13}
    & \multicolumn{4}{c|}{December `19 } & \multicolumn{4}{c|}{January `20} & \multicolumn{4}{c}{March `20}\\
    & \multicolumn{4}{c|}{$(\alpha, \beta, \Delta) = (2, 2, 5)$ } & \multicolumn{4}{c|}{$(\alpha, \beta, \Delta) = (2, 2, 5)$} & \multicolumn{4}{c}{$(\alpha, \beta, \Delta) = (2, 2, 5)$}\\
    & \texttt{R-MAE(I)} & $\sigma$\texttt{(I)} &
    \texttt{MAE}($\rho_{\mathrm{test}}$) & \texttt{MAE}($\rho_{\mathrm{all}}$) & \texttt{R-MAE(I)} & $\sigma$\texttt{(I)} &
    \texttt{MAE}($\rho_{\mathrm{test}})$ & \texttt{MAE}($\rho_{\mathrm{all}}$) & \texttt{R-MAE(I)} & $\sigma$\texttt{(I)} &
    \texttt{MAE}($\rho_{\mathrm{test}}$) & \texttt{MAE}($\rho_{\mathrm{all}}$) \\ \hline
    Hawkes$_{Den}$  & 0.249 & 0.099 & 0.105 & 0.047 & 0.526 & 0.091 & 0.148 & 0.101 & 0.517 & 0.054 & 0.162 & 0.133\\
    \riskModelName{}$_{Mob}$ &  0.235 & 0.088 & 0.164 & 0.195 & 0.117 & 0.077 & 0.193 & 0.166 & 0.184 & \textbf{0.037} & 0.180 & \textbf{0.096}\\
    \riskModelName{}$_{Mob^+}$ & \textbf{0.095} & \textbf{0.083} & \textbf{0.068} & \textbf{0.062} & \textbf{0.106} & \textbf{0.074} & \textbf{0.103} & \textbf{0.077} & \textbf{0.163} & \textbf{0.037} & \textbf{0.130} & 0.136
 \\\hline
    \end{tabular}}\vspace{2pt}
    \caption{Predicting (5-day) Infections and Risk for Los Angeles, CA. The table shows the error in predicted infections (\texttt{I}), the corresponding standard deviation, risk ($\rho$) for the test set, and over all days for Dec `19, Jan `20, and Mar `20 for the top-$5$ clusters.}
    \label{tab:LA_res}
    \vspace{-25pt}
\end{table}  







\begin{table}[h]
    \centering
    \resizebox{0.98\textwidth}{!}{
    \begin{tabular}{c c c c c| c c c c| c c c c}
    \multirow{4}{*}{\textbf{\Large Model}} &  \multicolumn{12}{c}{\textbf{\Large Infection and Risk Prediction for Seattle, WA} \vspace{2pt}}\\\cline{2-13}
    & \multicolumn{4}{c|}{December `19 } & \multicolumn{4}{c|}{January `20} & \multicolumn{4}{c}{March `20}\\
    & \multicolumn{4}{c|}{$(\alpha, \beta, \Delta) = (2, 2, 12)$ } & \multicolumn{4}{c|}{$(\alpha, \beta, \Delta) = (2, 2, 12)$} & \multicolumn{4}{c}{$(\alpha, \beta, \Delta) = (2, 2, 12)$}\\
    & \texttt{R-MAE(I)} & $\sigma$\texttt{(I)} &
    \texttt{MAE}($\rho_{\mathrm{test}}$) & \texttt{MAE}($\rho_{\mathrm{all}}$) & \texttt{R-MAE(I)} & $\sigma$\texttt{(I)} &
    \texttt{MAE}($\rho_{\mathrm{test}})$ & \texttt{MAE}($\rho_{\mathrm{all}}$) & \texttt{R-MAE(I)} & $\sigma$\texttt{(I)} &
    \texttt{MAE}($\rho_{\mathrm{test}}$) & \texttt{MAE}($\rho_{\mathrm{all}}$) \\ \hline
    Hawkes$_{Den}$ & 0.424 & 0.213 & 0.097 & 0.039 & 1.190 & 0.312 & 0.110 & 0.067 & 1.518 & 0.173 & 0.179 & 0.109\\
    \riskModelName{}$_{Mob}$ & 0.259 & 0.198 & \textbf{0.075} & 0.038 & 0.756 & 0.267 & 0.100 & \textbf{0.052} & 1.329 & 0.168 & 0.206 & 0.115 \\
    \riskModelName{}$_{Mob^+}$&\textbf{0.227} & \textbf{0.195} & 0.077 & \textbf{0.032} & \textbf{0.614} & \textbf{0.254} & \textbf{0.085} & 0.055 & \textbf{0.504}&\textbf{0.131}&\textbf{0.155}&\textbf{0.104} \\\hline
    \end{tabular}}\vspace{2pt}
    \caption{Predicting (5-day) Infections and Risk for Seattle, WA. The table shows the error in predicted infections (\texttt{I}), the corresponding standard deviation, risk ($\rho$) for the test set, and over all days for Dec `19, Jan `20, and Mar `20 for the top-$5$ clusters. }
    \label{tab:SEA_res}
    \vspace{-25pt}
\end{table}  

    



\clearpage
\begin{table}[th]
    \centering
    \resizebox{0.98\textwidth}{!}{
    \begin{tabular}{c c c c c |c c c c |c c c c}
    \multirow{4}{*}{\textbf{\Large Model}} &  \multicolumn{12}{c}{\textbf{\Large Infection and Risk Prediction for New York (Manhattan), NY} \vspace{2pt}}\\\cline{2-13}
    & \multicolumn{4}{c|}{December `19 } & \multicolumn{4}{c|}{January `20} & \multicolumn{4}{c}{March `20}\\
    & \multicolumn{4}{c|}{$(\alpha, \beta, \Delta) = (2, 2, 16)$ } & \multicolumn{4}{c|}{$(\alpha, \beta, \Delta) = (2, 2, 18)$} & \multicolumn{4}{c}{$(\alpha, \beta, \Delta) = (2, 2, 11)$}\\
    & \texttt{R-MAE(I)} & $\sigma$\texttt{(I)} &
    \texttt{MAE}($\rho_{\mathrm{test}}$) & \texttt{MAE}($\rho_{\mathrm{all}}$) & \texttt{R-MAE(I)} & $\sigma$\texttt{(I)} &
    \texttt{MAE}($\rho_{\mathrm{test}})$ & \texttt{MAE}($\rho_{\mathrm{all}}$) & \texttt{R-MAE(I)} & $\sigma$\texttt{(I)} &
    \texttt{MAE}($\rho_{\mathrm{test}}$) & \texttt{MAE}($\rho_{\mathrm{all}}$) \\ \hline
    Hawkes$_{Den}$ & 1.045 & 0.270 & 0.076 & 0.067 &2.529$^*$ & 0.508$^*$ & 0.080$^*$ & 0.092$^*$ & 7.173 & 0.482 & 0.311 & 0.199 \\
    \riskModelName{}$_{Mob}$ &  0.210 & 0.170 & 0.035 & \textbf{0.058} & 1.068 & 0.356 & 0.060 & 0.089 &  7.446 & 0.492 & 0.344 & 0.205\\
    \riskModelName{}$_{Mob^+}$& \textbf{0.145} & \textbf{0.162} & \textbf{0.021} & 0.061 & \textbf{0.448} & \textbf{0.294} & \textbf{0.037} & \textbf{0.080} & \textbf{4.440} & \textbf{0.370} & \textbf{0.248} & \textbf{0.174} \\\hline
    \end{tabular}}\vspace{2pt}
    \caption{Predicting (5-day) Infections and Risk for New York (Manhattan), NY. The table shows the error in predicted infections (\texttt{I}), the corresponding standard deviation, risk ($\rho$) for the test set, and over all days for Dec `19, Jan `20, and Mar `20 for the top-$5$ clusters. $^*$ The model is essentially a constant model based on the $\chi^2$ statistic.}
    \label{tab:NY_res}
    \vspace{-25pt}
\end{table}  




    

\vspace{20pt}

\begin{table}[!h]
    \centering
    \resizebox{0.98\textwidth}{!}{
    \begin{tabular}{c c c c c| c c c c| c c c c}
    \multirow{4}{*}{\textbf{\Large Model}} &  \multicolumn{12}{c}{\textbf{\Large Infection and Risk Prediction for Salt Lake County, UT}\vspace{2pt}}\\\cline{2-13}
    & \multicolumn{4}{c|}{December `19 } & \multicolumn{4}{c|}{January `20} & \multicolumn{4}{c}{March `20}\\
    & \multicolumn{4}{c|}{$(\alpha, \beta, \Delta) = (2, 3, 17)$ } & \multicolumn{4}{c|}{$(\alpha, \beta, \Delta) = (2, 3, 19)$} & \multicolumn{4}{c}{$(\alpha, \beta, \Delta) = (2, 3, 13)$}\\
    & \texttt{R-MAE(I)} & $\sigma$\texttt{(I)} &
    \texttt{MAE}($\rho_{\mathrm{test}}$) & \texttt{MAE}($\rho_{\mathrm{all}}$) & \texttt{R-MAE(I)} & $\sigma$\texttt{(I)} &
    \texttt{MAE}($\rho_{\mathrm{test}})$ & \texttt{MAE}($\rho_{\mathrm{all}}$) & \texttt{R-MAE(I)} & $\sigma$\texttt{(I)} &
    \texttt{MAE}($\rho_{\mathrm{test}}$) & \texttt{MAE}($\rho_{\mathrm{all}}$) \\ \hline
    Hawkes$_{Den}$ & 0.813$^*$ & 0.409$^*$ & 0.137$^*$ & 0.069$^*$ & 1.390 & 0.430 & 0.162 & 0.087 & 1.694 & 0.363 & 0.238 & 0.162 \\
    \riskModelName{}$_{Mob}$ & 0.197 & 0.298 & 0.070 & 0.062 & 0.624 & 0.335 & 0.110 & 0.072 & 1.133 & 0.308 & \textbf{0.198} & \textbf{0.141} \\
    \riskModelName{}$_{Mob^+}$&\textbf{0.189} & \textbf{0.287} & \textbf{0.069} & \textbf{0.059} & \textbf{0.429} & \textbf{0.308} & \textbf{0.088} & \textbf{0.066} & \textbf{1.064} & \textbf{0.307} & 0.234 & 0.167 \\\hline
    \end{tabular}}\vspace{2pt}
    \caption{Predicting (5-day) Infections and Risk for Salt Lake County, UT. The table shows the error in predicted infections (\texttt{I}), the corresponding standard deviation, risk ($\rho$) for the test set, and over all days for Dec `19, Jan `20, and Mar `20 for the top-$5$ clusters. $^*$ The model is essentially a constant model based on the $\chi^2$ statistic.}
    \label{tab:SLC_res}
    \vspace{-25pt}
\end{table}  


\end{document}